\pdfoutput=1

\documentclass[11pt]{article}

\usepackage[final]{acl}
\usepackage{authblk}

\usepackage{times}
\usepackage{latexsym}

\usepackage[T1]{fontenc}
\usepackage[utf8]{inputenc}

\usepackage{microtype}

\usepackage{inconsolata}

\usepackage{graphicx}
\usepackage{booktabs}
\usepackage{tabularx}
\usepackage{fontawesome}
\usepackage[inline]{enumitem}
\usepackage{multirow}
\usepackage[inkscapearea=page]{svg}
\usepackage[most]{tcolorbox}
\usepackage{makecell}
\usepackage{subcaption}
\usepackage{dashrule}
\usepackage{rotating}
\usepackage{subcaption} 
\title{Table Understanding and (Multimodal) LLMs:\\ A Cross-Domain Case Study on Scientific vs.\ Non-Scientific Data}

\author{
  \textbf{Ekaterina Borisova}\textsuperscript{1,2},
  \textbf{Fabio Barth}\textsuperscript{1}, 
  \textbf{Nils Feldhus}\textsuperscript{1,2,3}, \\
  \textbf{Raia Abu Ahmad}\textsuperscript{1,2},
  \textbf{Malte Ostendorff}\textsuperscript{4},
  \textbf{Pedro Ortiz Suarez}\textsuperscript{5}, \\
  \textbf{Georg Rehm}\textsuperscript{1,6},
  \textbf{Sebastian Möller}\textsuperscript{1,2}
}

\affil{
    \textsuperscript{1}Deutsches Forschungszentrum für Künstliche Intelligenz GmbH (DFKI), \\ 
    \textsuperscript{2}Technische Universität Berlin, 
    \textsuperscript{3}BIFOLD, 
    \textsuperscript{4}Deutsche Telekom, \\ 
    \textsuperscript{5}Common Crawl Foundation,
    \textsuperscript{6}Humboldt-Universität zu Berlin \\ 
  \small{
   Corresponding author: \href{mailto:ekaterina.borisova@dfki.de}{ekaterina.borisova@dfki.de}
  }
}

\begin{document}
\maketitle
\begin{abstract}
Tables are among the most widely used tools for representing structured data in research, business, medicine, and education. Although LLMs demonstrate strong performance in downstream tasks, their efficiency in processing tabular data remains underexplored. In this paper, we investigate the effectiveness of both text-based and multimodal LLMs on table understanding tasks through a cross-domain and cross-modality evaluation. Specifically, we compare their performance on tables from scientific vs.\ non-scientific contexts and examine their robustness on tables represented as images vs.\ text. Additionally, we conduct an interpretability analysis to measure context usage and input relevance. We also introduce the \textbf{TableEval} benchmark, comprising 3017 tables from scholarly publications, Wikipedia, and financial reports, where each table is provided in five different formats: Image, Dictionary, HTML, XML, and \LaTeX{}. Our findings indicate that while LLMs maintain robustness across table modalities, they face significant challenges when processing scientific tables.
\end{abstract}

\section{Introduction}
\label{intro}

Tables are one of the most ubiquitous tools for presenting data in a structured or semi-structured manner. They are commonly represented in a variety of textual (e.\,g., HTML, \LaTeX{}, XML) or image formats (e.\,g., PNG, JPEG) and used across domains such as finance, medicine, and business, as well as in research and education.

\begin{figure*}[ht] 
\center{\includegraphics[width=0.99\textwidth]{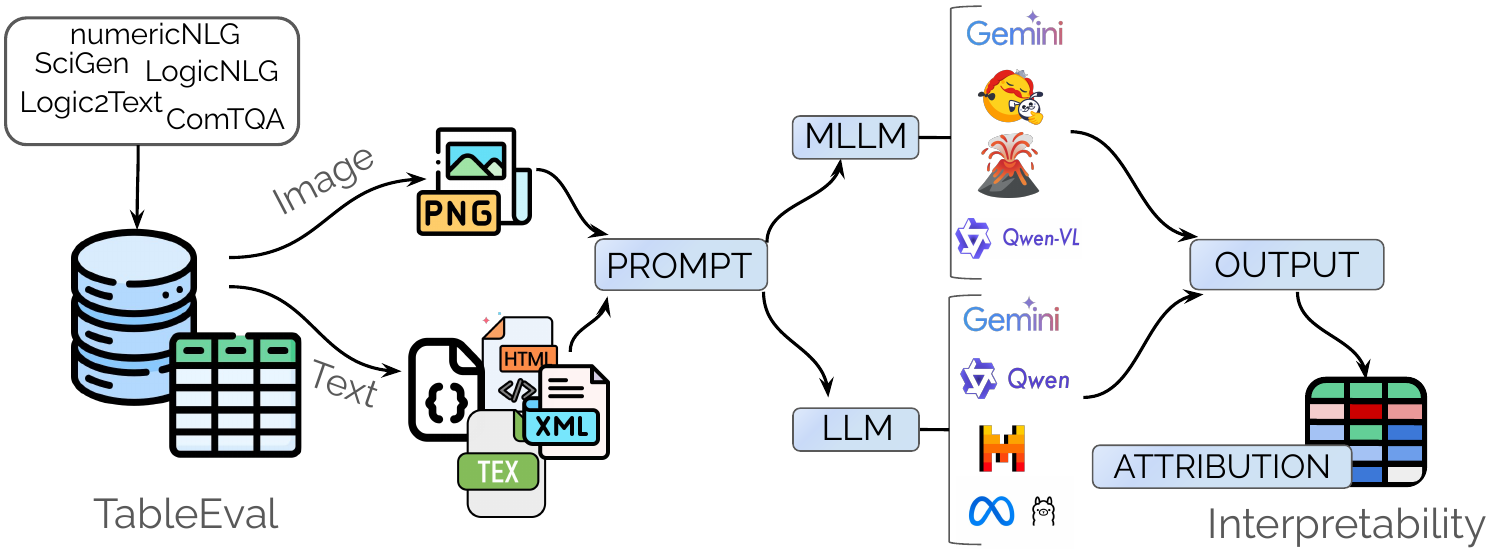}}
    \caption{Schematic representation of the main phases in our experiments: 1. Develop TableEval dataset, 2. Evaluate each (M)LLM on individual data subsets from TableEval using various table representations (Image, \LaTeX{}, XML, HTML, Dict), 3. Apply interpretability tools to the output yielding post-hoc feature attributions (e.\,g., using gradient-based saliency) which signify the importance of each token with respect to the model's output.}
    \label{fig: concept_diagrm}
\end{figure*}

In recent years, there has been a growing interest in table understanding (TU) techniques \cite{zhang2020, gorishniy2021, sahakyan2021, Borisov_2022, sui-2024-table-meets-llm, deng-2024-tables-as-texts-or-images}, aiming to extract and interpret information and knowledge contained in tables for tasks such as question answering (QA) and table-to-text generation (T2T) \cite{nan-etal-2022, cheng-etal-2022, oses-grijalba-etal-2024-question, zheng-etal-2024-multimodal}. While large language models (LLMs) demonstrate strong performance in a wide range of applications \cite{chang2024evaluationofllms, raiaan2024reviewonlargelanguagemodels, caffagni-etal-2024-revolution, zhang-etal-2024-mm, geminiteam2024geminifamilyhighlycapable, openai2024gpt4technicalreport}, their ability to understand (semi-)structured data remains under-researched \cite{sui-2024-table-meets-llm,fang2024largelanguagemodelsllmstabular} -- especially for tables from \emph{scientific} sources such as peer-reviewed articles, conference proceedings, and pre-prints.\footnote{Throughout this paper, we refer to such tables as \emph{scientific} and to tables from other sources as \emph{non-scientific}.} There is also limited research on the impact of the representation modality of structured data (i.\,e., image vs.\ text) on model performance \cite{deng-2024-tables-as-texts-or-images, zhang-2024-flextaf}, and to the best of our knowledge, there are no approaches yet that specifically address scientific tables. In particular, most TU studies primarily focus on tables from \emph{non-scientific} contexts such as Wikipedia \cite{parikh-etal-2020-totto, chen2021openquestionansweringtables, mammotab2022, wu2024tablebenchcomprehensivecomplexbenchmark, pang-2024-tabis}. However, compared to these domains, scientific tables often include technical terminology, complex concepts, abbreviations, and dense numerical values, requiring domain-specific knowledge and strong arithmetic reasoning skills \cite{ho2024surveypretrainedlanguagemodels,moosavi2021learningreasontextgeneration}. Recent works \cite{yang-2025-does-table-source-matter,wu-2024-scimmir} indicate that scientific tables present challenges to multimodal LLMs (MLLMs) and incorporating such (semi-)structured data into pretraining improves performance. As the number of published articles continues to increase rapidly \cite{fortunato2018science, bornmann2021growth, Hong2021}, TU for scientific contexts, e.\,g., for scholarly document processing including information extraction and research knowledge graph construction, is becoming even more relevant. Finally, we notice that interpretability analysis \cite{ferrando-2024-primer-inner-workings} for TU has received little attention and remains underexplored \cite{fang2024largelanguagemodelsllmstabular}.

In this paper, we address the aforementioned gaps by examining the efficiency of both LLMs and MLLMs on a set of TU tasks. Specifically, we compare their ability to handle (semi-)structured data from scientific and non-scientific sources and explore the effects of image vs. diverse text-based table representations on model performance. We also conduct feature importance analyses to interpret the use of context information in LLMs. Figure~\ref{fig: concept_diagrm} illustrates the main phases of our experiments. 

Our contributions can be summarised as follows:
\begin{itemize}
    \item We introduce TableEval, a cross-domain benchmark containing 3017 tables from scholarly publications, Wikipedia, and financial reports, available in image and four text formats (Dictionary, HTML, XML, and \LaTeX{}). The dataset is publicly available on Hugging Face: \url{https://huggingface.co/datasets/katebor/TableEval}
    \item We conduct an extensive evaluation revealing that, although current (M)LLMs remain robust across table modalities, their performance significantly declines on scientific tables compared to non-scientific ones.
    \item We examine the applicability of gradient-based explanations for LLMs \cite{sarti-2023-inseq} to TU to learn about the relevance of table content in prompts.
\end{itemize}

\section{TableEval benchmark} 
\label{data}

Since no existing dataset covers both scientific and non-scientific tables across text and image modalities, we construct a benchmark tailored to our evaluation. This section outlines the collection processes of data (\S\ref{source_data}) and diverse table formats (\S\ref{table_formats}).

\begin{table*}[ht!]  
\centering
\resizebox{\textwidth}{!}{
\begin{tabular}{lccccccc}
\toprule
\textbf{Dataset} &  \textbf{Task} & \textbf{Source} & \textbf{Image} & \textbf{Dict} & \textbf{\LaTeX{}} & \textbf{HTML} & \textbf{XML}  \\ 
\midrule
\multicolumn{8}{c}{\emph{Scientific tables}} \\
\midrule
ComTQA (PubTables-1M) &   VQA  & PubMed Central & \faDownload & \faGears    &\faGears   & \faGears    & \faFilesO  \\ 
numericNLG     &    T2T & ACL Anthology & \faFilesO  &\faDownload   & \faGears  &  \faDownload  & \faGears\\ 
SciGen     & T2T &	arXiv and ACL Anthology & \faFilesO  & \faDownload  & \faFilesO & \faGears    & \faGears \\
\midrule
\multicolumn{8}{c}{\emph{Non-scientific tables}} \\
\midrule
ComTQA (FinTabNet)& VQA & 	Earnings reports of S\&P 500 companies  & \faFilesO  & \faGears	 & \faGears & \faGears   & \faGears \\
LogicNLG    & T2T & Wikipedia        & \faGears   &\faDownload   & \faGears & \faFilesO  & \faGears  \\ 
Logic2Text   & T2T & Wikipedia       & \faGears   &\faDownload   & \faGears & \faFilesO  & \faGears \\ 
\bottomrule
\end{tabular}}
\caption{Overview on the formats and collection methods for each dataset. Symbol \faDownload \space indicates formats already available in the given corpus, while \faFilesO \space and \faGears \space denote formats extracted from the table source files (e.\,g., article PDF, Wikipedia page) and generated from other formats in this study, respectively.}
\label{table: data_formats}
\end{table*}

\subsection{Source data}
\label{source_data}

To study the cross-domain performance of (M)LLMs, we developed the TableEval benchmark by leveraging pre-existing datasets of scientific and non-scientific tables. We collected relevant datasets based on the following criteria: \begin{enumerate*}
    \item data is open-access;
    \item test set with the gold labels is available; 
    \item metadata includes references to the sources of tables, such as DOIs for scholarly papers or URLs for Wikipedia pages;
    \item target tasks (e.\,g., QA, T2T) are identical or very similar across datasets to maintain consistency and ensure comparability;
    \item tables can be converted to the pre-defined formats (see~\S\ref{table_formats}).
\end{enumerate*}
The following five datasets were selected (see Table~\ref{table: data_formats}): \begin{enumerate*}[label=(\alph*)] \item \textbf{ComTQA} \cite{zhao2024tabpedia}, a visual QA (VQA) benchmark containing tables from PubTables-1M \cite{pubtables1m} and FinTabNet \cite{zheng2020globaltableextractorgte}, originating from PubMed Central\footnote{\url{https://pubmed.ncbi.nlm.nih.gov}} (PMC) papers and annual earnings reports, respectively. The annotations are generated using Gemini Pro \cite{geminiteam2024geminifamilyhighlycapable} and include questions requiring multiple answers, calculations, and logical reasoning. \item \textbf{numericNLG} \cite{suadaa-etal-2021-towards}, a dataset focusing on the T2T generation task with numerical reasoning based on tables and their textual descriptions extracted from ACL Anthology\footnote{\url{https://aclanthology.org}} articles and annotated by experts in the Computer Science field. \item \textbf{SciGen} \cite{moosavi2021learningreasontextgeneration}, a corpus designed for reasoning-aware T2T generation, comprising tables from arXiv\footnote{\url{https://arxiv.org}} papers across fields such as Computation and Language, Machine Learning, Computer Science, Computational Geometry, etc. Its test set contains expert-annotated data. \item \textbf{LogicNLG} \cite{chen-etal-2020-logical}, a T2T dataset of open-domain tables from Wikipedia and associated with manually annotated natural language statements that can be logically entailed by the given data. \item \textbf{Logic2Text} \cite{chen-etal-2020-logic2text}, features open-domain Wikipedia tables manually annotated with descriptions of common logic types and their underlying logical forms for the T2T task. \end{enumerate*} As shown in Table~\ref{table: data_formats}, the final TableEval corpus contains six data subsets, covering two downstream tasks (QA and T2T), and comprising 3017 tables and 11312 instances in total (for the detailed statistics see Table~\ref{table: instances_per_format} in Appendix~\ref{appendix:dataset_stat}). All annotations are taken from the source datasets. Examples from each dataset are provided in Appendix~\ref{appendix: task_ex_per_dataset}.

\subsection{Table formats}
\label{table_formats}

We represent tables from each TableEval subset as PNG images and in structured or semi-structured textual formats including HTML, XML, \LaTeX{}, and Python Dictionary (Dict) to analyse LLMs' performance across different modalities. HTML is chosen as it is the original format of Wikipedia tables, XML for its use in encoding tables from PMC articles, \LaTeX{} as it is the primary format for scientific tables, and Dict since it is readily available in most source datasets. Instances of tables in various representation formats were obtained using one of the following methods (see Table~\ref{table: data_formats}): \begin{enumerate*} 
\item extraction from the original dataset; 
\item extraction from the table source (e.\,g., article PDF);
\item generation from other formats (e.\,g., HTML $\Leftrightarrow$ XML). 
\end{enumerate*}
Note that for the latter two, we manually validate the final results for each format and data subset by checking a random sample of about 100 instances. In what follows, the way we assembled each table format in the TableEval corpus is described in detail. Additional information is provided in Appendix~\ref{appendix: additiona_info_formats}.

\textbf{Image}. Since the PubTables-1M subset of ComTQA already includes JPGs of tables, we simply convert them to PNGs. In contrast, other datasets provide only textual representations of tables. Thus, for numericNLG and SciGen, we first collect PDF files of the arXiv and ACL papers, and then use the PDFFigure2.0 \cite{PDFFigures} tool to extract images of tables.\footnote{In SciGen, some PDFs are taken from the ACL Anthology as they are no longer available on arXiv.} Whenever PDFFigure2.0 fails to produce an image, we utilise the MinerU tool \cite{wang2024mineruopensourcesolutionprecise} as an alternative. Note that SciGen instances associated with papers that are no longer open-access or do not contain tables are excluded. In case of FinTabNet, images of tables are extracted from the corresponding PDF pages of financial reports using the gold annotations of the bounding boxes. Finally, images of the Wikipedia tables in LogicNLG and Logic2Text are generated by converting their HTML representations into PNG files with the imgkit Python wrapper\footnote{\url{https://pypi.org/project/imgkit/}}. Distribution of image aspect rations across data subsets is provided in Figure~\ref{fig: aspect_ratio} in Appendix~\ref{appendix: aspect_ratios}.

\textbf{XML and HTML}. PubTables-1M is the only dataset where the original XML sources of tables can be obtained. To achieve this, we retrieve the source papers based on their PMC ID using the E-utilities API\footnote{\url{https://www.ncbi.nlm.nih.gov/home/develop/api/}} and extract the tables with the ElementTree parser\footnote{\url{https://docs.python.org/3/library/xml.etree.elementtree.html\#}}. When it comes to HTML, we are unable to retrieve the original format since systematic downloading of article batches from the PMC website is prohibited\footnote{\url{https://pmc.ncbi.nlm.nih.gov/about/copyright/}}. This is why we generate HTML from XML using a custom Python script instead. Similarly, for numericNLG, we convert already available HTML into XML with a Python script. 
For SciGen, we download the source \LaTeX{} code of each paper from arXiv, use the \LaTeX{ML} tool\footnote{\url{https://math.nist.gov/~BMiller/LaTeXML/}} to produce both XML and HTML, and extract tables from the resulting files. In contrast, we construct HTML for FinTabNet tables by leveraging gold annotations of HTML structure which provide tags and associated cell values. Afterwards, the HTML code is converted to XML in the same way as described for numericNLG. Finally, HTML in LogicNLG and Logic2Text are collected from the respective Wikipedia pages, while the XML format is obtained using the same approach applied to numericNLG and FinTabNet. 
 
\textbf{\LaTeX{}}. For SciGen, we obtain the \LaTeX{} code directly from the source files of the papers. In contrast to arXiv data, no \LaTeX{} code is available for PMC and ACL papers. Thus, we generate \LaTeX{} for numericNLG and PubTables-1M tables from their HTML representations. To ensure the validity of the output, we compile the code and resolve any errors encountered. The same approach is used to obtain \LaTeX{} for Wikipedia and financial tables. 

\textbf{Dictionary}. All datasets except ComTQA already include linearised tables represented as lists of column headers and cell values, although the encoding conventions slightly vary across them (see Appendix~\ref{appendix: additiona_info_formats}). To align with these datasets, we collect column headers, subheaders, and cell values for the PMC subset in ComTQA by parsing the table XML code with ElementTree. In case of FinTabNet, we extract these elements from a dataframe representation of each table obtained during the HTML collection phase. For the experiments, the linearised tables are represented as a Dict containing lists of column headers, lists of subheaders (if extracted), lists of rows, as well as title, caption, and footnote (if available). 

\section{Experiments}

We benchmark various (M)LLMs using individual data subsets and representations of tables from TableEval. This is followed by an interpretability analysis applied to the output yielding attributions from a gradient-based method. In the following, we first describe the experimental set up (\S\ref{exp_setup}), then report and analyse the results (\S\ref{results}).

\subsection{Experimental setup}
\label{exp_setup}

\paragraph{Models.} We evaluate both smaller and larger models in terms of parameter size (3-14 billion), see Table~\ref{table:list_of_llms}.\footnote{Due to limited computational resources, we restricted the evaluation to (M)LLMs with up to 14 billion parameters.} We primarily focus on open-source instruction-tuned (M)LLMs published on Hugging Face\footnote{\url{https://huggingface.co}} (HF). The only closed-source model we use is Gemini-2.0-Flash \cite{geminiteam2024geminifamilyhighlycapable}, which serves as our baseline, since Gemini is currently considered among the state-of-the-art. For MLLMs, we select 
LLaVa-NeXT \cite{li2024llavanext-strong}, Qwen2.5-VL \cite{bai2025qwen25vltechnicalreport}, and Idefics3 \cite{laurençon2024buildingbetterunderstandingvisionlanguage}. As for text-based LLMs, we evaluate Llama-3 \cite{grattafiori2024llama3herdmodels}, Qwen2.5 \cite{qwen2025qwen25technicalreport}, and Mistral-Nemo\footnote{\url{https://mistral.ai/news/mistral-nemo}}.

\begin{table}[ht]  
\centering
\resizebox{\columnwidth}{!}{
\begin{tabular}{llcc}
\toprule
\textbf{Model} & \textbf{HF checkpoint} & \textbf{Size (B)} & \textbf{Vision}\\ 
\midrule
\textbf{Gemini-2.0-Flash}              & -- & -- &\faCheck \\
\textbf{LLaVa-NeXT}                   & llama3-llava-next-8b-hf & 8 & \faCheck \\ 
\multirow{2}{*}{\textbf{Qwen2.5-VL}}  & Qwen2.5-VL-3B-Instruct  &3  & \faCheck\\  
                                      & Qwen2.5-VL-7B-Instruct  & 7 & \faCheck\\  
\textbf{Idefics3}                     & Idefics3-8B-Llama3      & 8& \faCheck \\  
\textbf{Llama-3}                      & Llama-3.2-3B-Instruct   & 3 & \faTimes\\  
\multirow{2}{*}{\textbf{Qwen2.5}}     & Qwen2.5-3B-Instruct     &  3& \faTimes \\
                                      & Qwen2.5-14B-Instruct    &14 &\faTimes \\
\textbf{Mistral-Nemo}                 & Mistral-Nemo-Instruct-2407& 12 &\faTimes\\ 
\bottomrule
\end{tabular}}
\caption{(M)LLMs used in the experiments (``Size'' indicates the number of parameters in billions).}
\label{table:list_of_llms}
\end{table}

\begin{figure*}[ht] 
    \center{\includegraphics[width=0.8\textwidth]{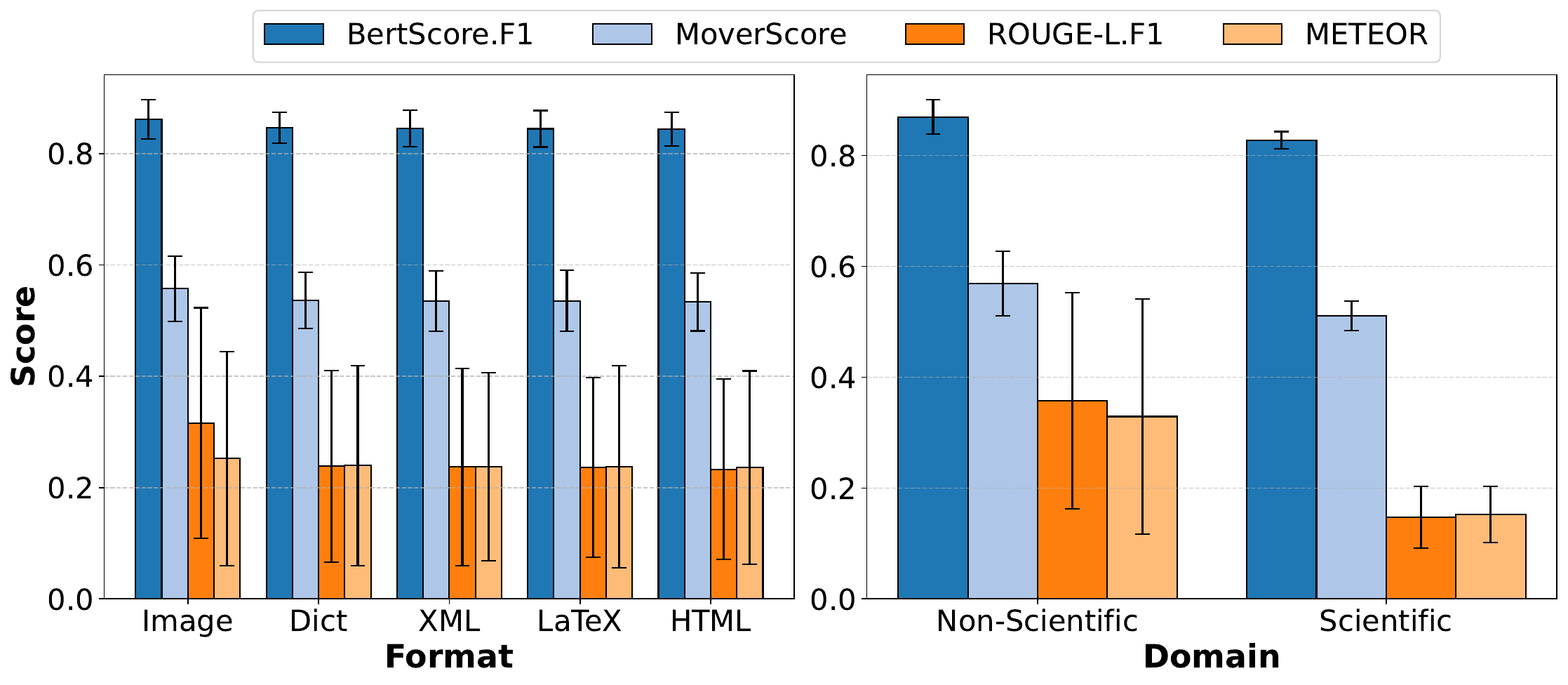}}
    \caption{BertScore.F1, MoverScore, ROUGE-L.F1, and METEOR for the table formats averaged over data subsets and models (left), and for scientific vs.\ non-scientific domain averaged over data subsets, models, and formats (right). Error bars indicate standard deviation.}
    \label{fig: exp_format_domain}
\end{figure*}

\paragraph{Prompts and data.} We run experiments on every data subset from the TableEval corpus and develop prompt templates that are customised to each task, applying them uniformly across all models to ensure consistency during the evaluation. To study the models' true capability to understand various table representations, we exclude explicit document type indicators (e.\,g., HTML/XML headers) and do not specify the format in the prompt. Additionally, given the diversity of the (M)LLMs and the fact that they may not always adhere to a specific output structure (which can hinder proper parsing of the answer), we do not enforce a particular response format. The prompt templates are provided in Appendix~\ref{appendix: prompts}. 
\paragraph{Evaluation metrics.} We follow the scores reported in the original papers for each data subset. Thus, we compute BLEU-N \cite{papineni-etal-2002-bleu}, SacreBLEU \cite{post-2018-call}, METEOR \cite{banerjee-lavie-2005-meteor}, ROUGE-N, ROUGE-L \cite{lin-2004-rouge}, MoverScore \cite{zhao-etal-2019-moverscore}, BertScore \cite{zhang2020BERTScore}, and BLEURT \cite{sellam-etal-2020-bleurt}. Given the extensive set of metrics, we report only BertScore.F1, MoverScore, ROUGE-L.F1, and METEOR in the main text, while providing all raw score values in Appendix~\ref{appendix: tables_results}. 

\paragraph{Interpretability analysis.} Inseq \cite{sarti-2023-inseq} applies feature attribution methods to generative LLMs to highlight how important each token in the input is for generating the next token with the help of a heatmap.
In our experimental setup, we perform post-hoc analyses using the model outputs as custom attribution targets on an instance level. Input x Gradient \cite{simonyan-2014-deep-inside}, provided by Inseq, is selected as it is both computationally efficient and more faithful than, e.\,g., attention weights. The saliency is averaged to produce a one-dimensional vector of token attributions,
which we visualise as a heatmap.

\paragraph{Implementation details.} All experiments are conducted in a zero-shot setting using the (M)LLMs' default hyperparameters with the seed value set to 42. We choose the batch size equal to 1 for all open-source (M)LLMs and to the size of the given subset for Gemini-2.0-Flash. We use Nvidia A100 (40GB, 80GB), H100 (80GB), H200 (141GB), and L40S (48GB) GPUs for the open-source models depending on the given LLM and TableEval subset size. The Gemini-2.0-Flash results are evaluated using the Batch API through the LiteLLM framework\footnote{\url{https://www.litellm.ai}}. We developed an end-to-end evaluation pipeline\footnote{\url{https://github.com/esborisova/TableEval-Study}} for the experiments and use HF transformers or LiteLLM and the datasets library to load the models and datasets, respectively.

\begin{figure*}[ht] 
    \center{\includegraphics[width=0.99\textwidth]{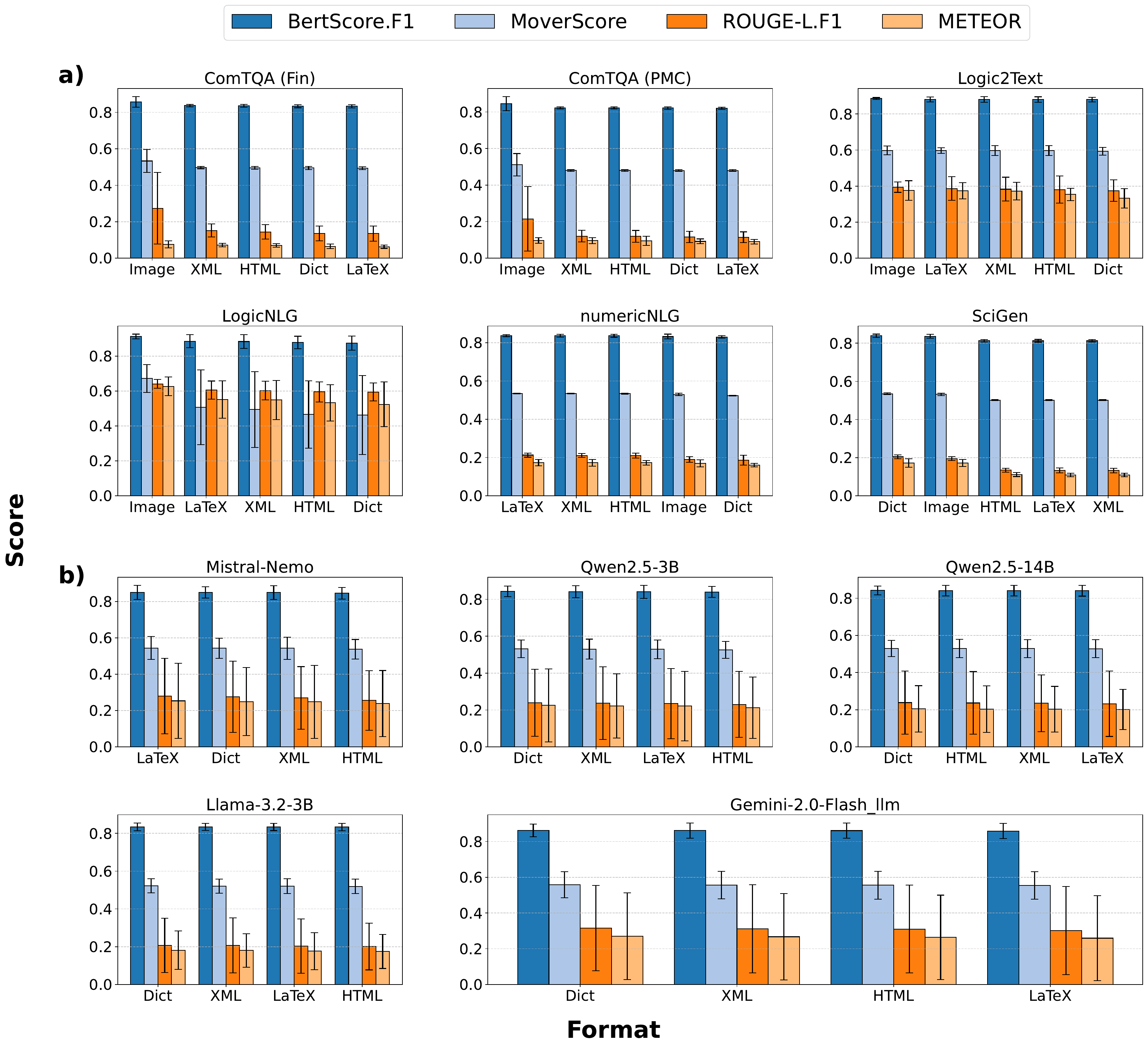}}
    \caption{Values of BertScore.F1, MoverScore, ROUGE-L.F1, and METEOR \textbf{a)} for individual data subsets and all formats averaged over models, and \textbf{b)} for individual models and text formats averaged over data subsets. Error bars indicate standard deviation. Here ``Fin'' stands for FinTabNet, ``PMC'' denotes PubTables-1M, while ``\_llm'' indicates text input for Gemini-2.0-Flash.}
    \label{fig: results_per_llm_format_split}
\end{figure*}

\subsection{Results and analysis}
\label{results}
\paragraph{Image vs.\ text.} Averaged score values across models and data subsets for each table format are given in Figure~\ref{fig: exp_format_domain} (left), whereas raw results are shown in Table~\ref{table: scores_per_format} in Appendix~\ref{appendix: tables_results}. The use of images outperforms the use of text across all metrics by approximately 1-13\%. In particular, for ComTQA and LogicNLG, image achieves the best results, while for other data subsets the outcomes are either similar or the text modality prevails (by about 1-10\%), as shown in Figure~\ref{fig: results_per_llm_format_split} a) and Tables~\ref{table: raw_scores_per_format_comtqa_fin}--\ref{table: raw_scores_per_format_scigen} in Appendix~\ref{appendix: tables_results}. This aligns with previous studies \cite{deng-2024-tables-as-texts-or-images} reporting comparable or significantly better performance of models on the vision modality. Unlike prior works \cite{sui-2024-table-meets-llm,singha2023tabularrepresentationnoisyoperators,deng-2024-tables-as-texts-or-images}, we do not observe a large variation in results across LLMs and the four text formats, with the maximum gap equal to about 4\%. Further analysis of the metrics for individual models and formats also indicates similar accuracy across the LLMs, see Figure~\ref{fig: results_per_llm_format_split} b) and Tables~\ref{table: raw_scores_per_llm_format_llama}--\ref{table: raw_scores_per_llm_format_gemini} in Appendix~\ref{appendix: tables_results}. Hence, our findings suggest that current models are less sensitive to diverse text representations of tables. Such outcomes may be attributed to LLMs' exposure to data encoded in the given formats during pretraining. 

\paragraph{Scientific vs.\ non-scientific.} The results for each domain are shown in Figure~\ref{fig: exp_format_domain} (right) and Table~\ref{table: scores_per_domain} in Appendix~\ref{appendix: tables_results}. The findings indicate that LLMs are more efficient on TU tasks from the non-scientific split, achieving a score boost of up to 34\%. The best score values are obtained for LogicNLG followed by Logic2Text, see Figure~\ref{fig: exp_data_model} (left) and Table~\ref{table: scores_per_subset} in Appendix~\ref{appendix: tables_results}. 

\begin{figure*}[ht] 
    \center{\includegraphics[width=0.9\textwidth]{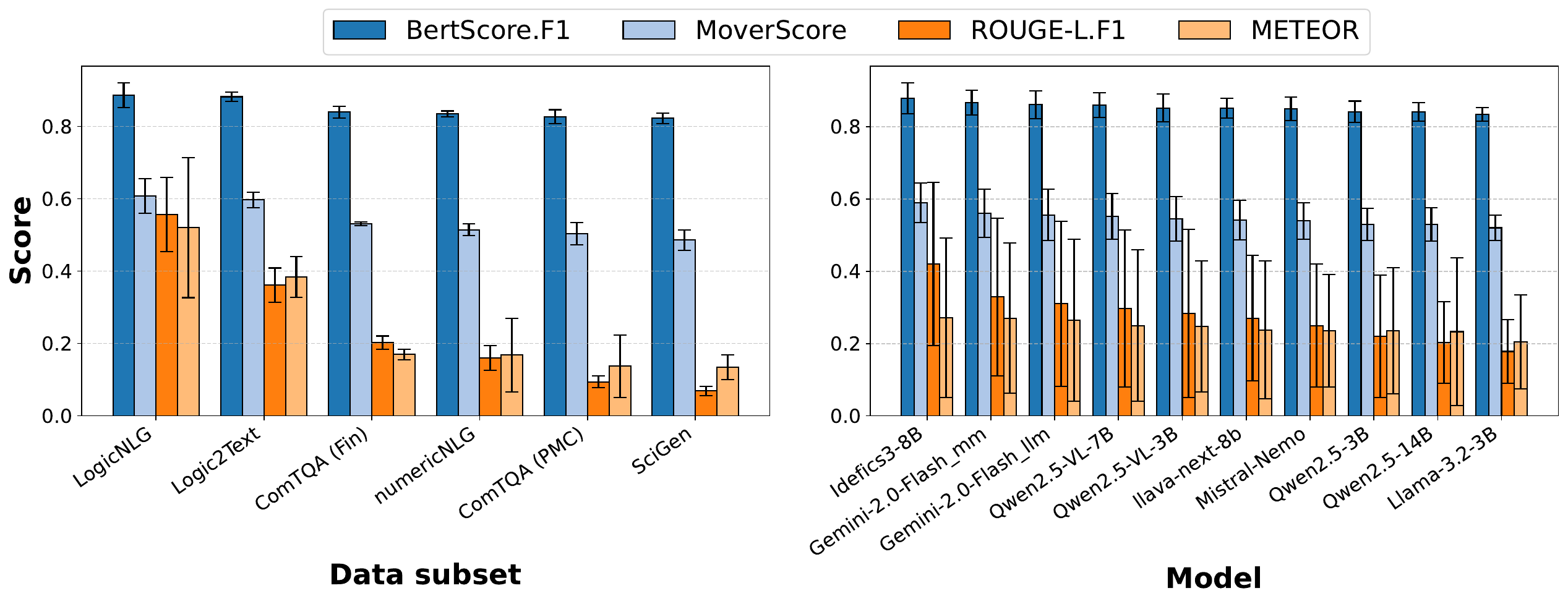}}
    \caption{BertScore.F1, MoverScore, ROUGE-L.F1, and METEOR for each data subset averaged over table formats and models (left), and for individual models averaged over data subsets and formats (right). Error bars indicate standard deviation. Here ``Fin'' stands for FinTabNet, ``PMC'' denotes PubTables-1M, while ``\_llm'' and ``\_mm'' are used to distinguish between text and image input for Gemini-2.0-Flash, respectively.}
    \label{fig: exp_data_model}
\end{figure*}

We hypothesise that this difference could arise from \begin{enumerate*}[label=(\alph*)]
    \item the complexity level of the given data and the target task;
    \item lack or sparsity of the data from scientific contexts in the pre-training corpus of (M)LLMs. 
\end{enumerate*} In numericNLG and SciGen, the goal is to generate a coherent paragraph or a collection of paragraphs summarising the table's content. In contrast, both LogicNLG and Logic2Text involve producing a single statement, filling in masked entities in a sentence and generating text based on a logical form, respectively. Furthermore, according to \citeauthor{moosavi2021learningreasontextgeneration} (\citeyear{moosavi2021learningreasontextgeneration}), SciGen is characterised by a higher level of complexity than LogicNLG. This is because each gold description in SciGen summarises the entire table content and involves multiple types of reasoning, whereas, in LogicNLG each statement often focuses on a subset of table rows and is associated with a single type of reasoning. Similar to LogicNLG, Logic2Text descriptions involve only one type of logic. Notably, comparable performance is achieved across models for both subsets in ComTQA, with the gap in scores equal to about 1-3\% (except for a 17\% higher BLEURT score for PubTables-1M). Given that ComTQA was also proposed as a more challenging benchmark compared to existing datasets, comprising questions with multiple answers, numerical, and logical reasoning, the lower performance of (M)LLMs could lie in the complexity of the data as well. Finally, reasoning over scientific tables requires in-domain knowledge, the absence of which likely contributes to a decline in accuracy for the respective TableEval subsets.  

\paragraph{Comparison of (M)LLMs.} Figure~\ref{fig: exp_data_model} (right) and Table~\ref{table: scores_per_model} in Appendix~\ref{appendix: tables_results} outline results for individual models. Among MLLMs, Gemini-2.0-Flash and Idefics3 perform best, with the former outperforming the latter on BLEU-N, BLEURT, METEOR, ROUGE-3, and ROUGE-4 (by 1-4\%). Next in the ranking are Qwen2.5-VL models and LLaVa-NeXT. For LLMs, Gemini-2.0-Flash obtains the highest score values, followed by Mistral-Nemo. Qwen2.5 models rank next with the 3B version achieving either similar or slightly better results than its 14B counterpart. On the contrary, Llama-3 consistently shows the weakest performance. We observe that on average, Idefics3 tends to generate concise responses with the shortest outputs produced for QA task (e.\,g., just a numeric value), whereas other models provide longer outputs. A similar trend is observed for LLMs, with Gemini-2.0-Flash providing shorter predictions compared to other models. Table~\ref{table: pred_length_stat} outlines the statistics on prediction lengths for each (M)LLM. Additionally, Figure~\ref{fig: pred_lenth_for_each_dataset} (Appendix~\ref{appendix: tables_results}) illustrates the mean lengths for each model and data subset, while Figure~\ref{fig: pred_lenght_ex} (Appendix~\ref{appendix: case_study}) demonstrates prediction examples. Since we do not postprocess the models' outputs, such difference in response length can contribute to the discrepancy across (M)LLMs in BLEU-N and ROUGE-N, which rely on n-gram overlap. Overall, our evaluation indicates that open-source models still remain behind the closed-source Gemini-2.0-Flash. On another note, we could not observe any correlation between model size and accuracy.  

\begin{table}[ht]  
\centering
\resizebox{\columnwidth}{!}{
\begin{tabular}{lccc}
\toprule
\textbf{Model} & \textbf{Mean} & \textbf{Min} & \textbf{Max} \\
\midrule
Idefics3-8B-Llama3        & \cellcolor{blue!30}139   & \cellcolor{blue!30}0 & 4416 \\
Qwen2.5-VL-3B-Instruct    & 360                     & 2                      & 4170 \\
Qwen2.5-VL-7B-Instruct    & 292                      & 4                     & 3464 \\
llama3-llava-next-8b-hf   & 311                      & 24                    & 6336 \\
Gemini-2.0-Flash\_mm       & 207                      & 2                   & 3097 \\
Gemini-2.0-Flash\_llm      & 259                      & \cellcolor{blue!30}0                  & \cellcolor{pink!90}10282 \\
Llama-3.2-3B-Instruct     & 464                      & 22                    &5626 \\
Mistral-Nemo-Instruct-2407 & 303                     & 21                    & \cellcolor{blue!30}2941 \\
Qwen2.5-14B-Instruct       & \cellcolor{pink!90}481  & \cellcolor{pink!90}29 & 4154 \\
Qwen2.5-3B-Instruct        & 465                     & 26                    & 4535 \\
\bottomrule
\end{tabular}}
\caption{Statistics on the mean, minimum, and maximum prediction lengths (in characters) for each model across TableEval subsets. \colorbox{blue!30}{Blue} and \colorbox{pink!90}{pink} colours highlight the lowest and highest values in each column, respectively. Here ``\_llm'' and ``\_mm'' are used to distinguish between text and image input for Gemini-2.0-Flash, respectively.}
\label{table: pred_length_stat}
\end{table}

\begin{figure*}[ht]
    \centering
    \begin{minipage}[t]{0.03\textwidth}
        \vspace{.5cm}
        \rotatebox{90}{\small \textbf{Input/Prompt Attribution}}
    \end{minipage}
    \begin{subfigure}[t]{0.48\textwidth}
        \centering
        \textbf{\small Mistral-Nemo-Instruct-2407} \\
        \includegraphics[width=.9\columnwidth, trim=0cm 14.75cm 0cm 0cm, clip]{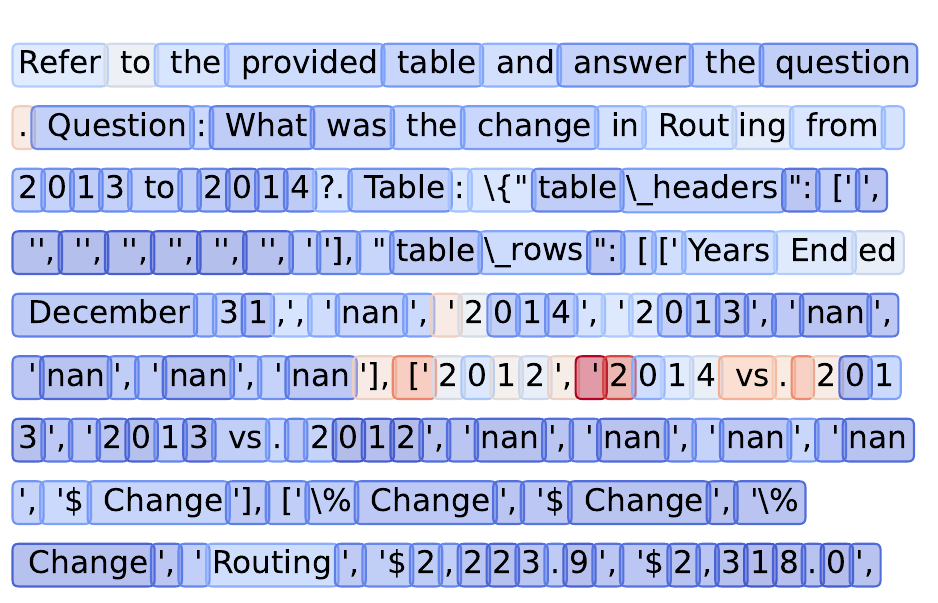}
    \end{subfigure}\hfill
    \begin{subfigure}[t]{0.48\textwidth}
        \centering
        \textbf{\small Llama-3.2-3B-Instruct} \\
        \includegraphics[width=.9\columnwidth, trim=0cm 13.75cm 0cm 0cm, clip]{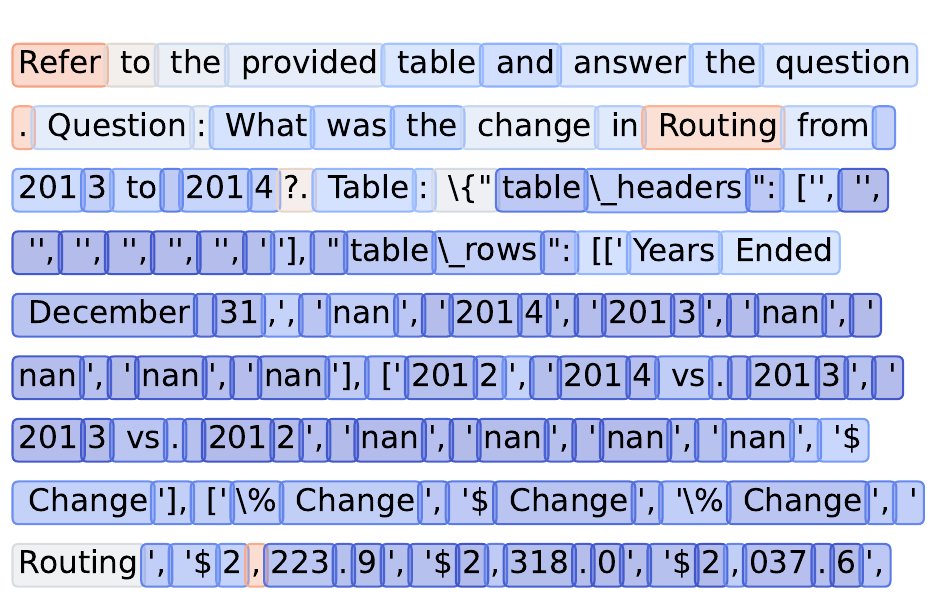}
    \end{subfigure}

    \vspace{0.05cm}
    \hdashrule{\textwidth}{0.5pt}{3pt}

    \begin{minipage}[t]{0.03\textwidth}
        \hspace{5pt}
    \end{minipage}
    \begin{subfigure}[t]{0.48\textwidth}
        \centering
        \includegraphics[width=.9\columnwidth, trim=0cm 0cm 0cm 21.75cm, clip]{figs/mistral_fin_110758-id-11-context-saliency.pdf}
    \end{subfigure}\hfill
    \begin{subfigure}[t]{0.48\textwidth}
        \centering
        \includegraphics[width=.9\columnwidth, trim=0cm 0cm 0cm 20.75cm, clip]{figs/llama_fin_110758-id-11-context-saliency.pdf}
    \end{subfigure}

    \vspace{0.1cm}
    \hrule

    \begin{minipage}[t]{0.015\textwidth}
        \vspace{-2.5cm}
        \rotatebox{90}{\footnotesize \textbf{Generation}}
    \end{minipage}
    \begin{minipage}[t]{0.015\textwidth}
        \vspace{-2.5cm}
        \rotatebox{90}{\footnotesize \textbf{Log-probabilities}}
    \end{minipage}
    \begin{subfigure}[t]{0.425\textwidth}
        \centering
        \vspace{-2.5cm}
        \includegraphics[width=.75\columnwidth, trim=0cm 0cm 0cm 0.675cm, clip]{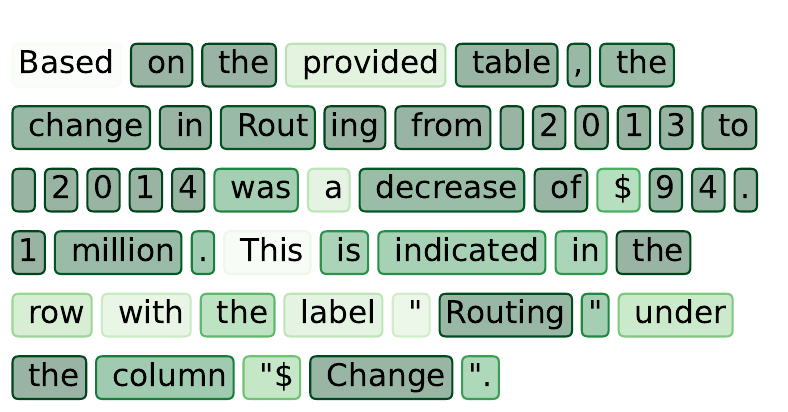}
    \end{subfigure}\hfill
    \begin{subfigure}[t]{0.475\textwidth}
        \centering
        \includegraphics[width=.875\columnwidth, trim=0cm 0cm 0cm 0.5cm, clip]{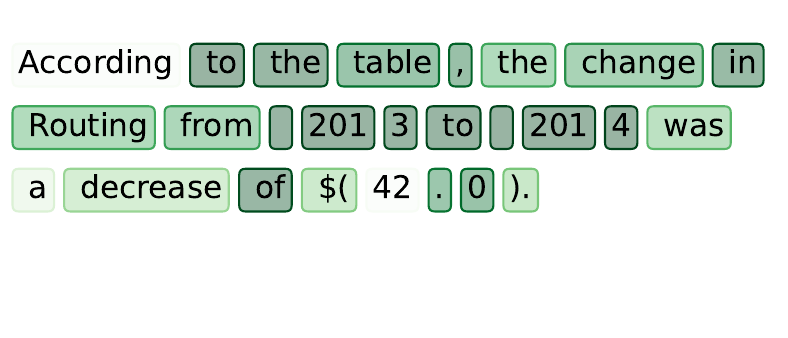}
    \end{subfigure}
    \caption{Interpretability analysis using Input x Gradient on Mistral-Nemo (correct prediction) and Llama3 (incorrect prediction) for a ComTQA (FinTabNet) instance with the Dict format. The gold answer to the given question is \emph{``decrease of \$94.1''}. Redder highlights correspond to higher importance. The prompts are abbreviated in the middle, indicated with the dashed line. In addition, for the output, we visualise the log-probabilities representing the model's confidence (dark green = very confident).}
    \label{fig:interp_results}
\end{figure*}

\paragraph{Interpretability.} We choose instance-level analysis because dataset-level statistics tend to flatten important nuances, especially in generative settings without a finite number of classes \cite{ronnqvist-2022-explaining-classes}. Due to computational and visualisation constraints, we selected four ComTQA and two LogicNLG instances. The former was chosen for its shorter reference and prediction lengths compared to other subsets, while the latter was selected for achieving the highest scores across LLMs. We compare the best (Mistral-Nemo) and worst (Llama3) performing open-source LLMs.\footnote{Saliency maps for these examples, along with additional instances, are available also in our GitHub repository.} 

Figure~\ref{fig:interp_results} shows saliency maps as determined by the Input x Gradient explainer and log-probabilities for the generation (see~\S \ref{exp_setup}). 
In this ComTQA (FinTabNet) example, with the table represented as a Dict in the input, we first notice that positive attributions are generally sparse due to the saturation problem \cite{shrikumar-2017-deeplift} and potentially the long context. Llama3 puts most attribution towards start and end of the prompt and the row value mentioned in the question (\emph{``Routing''}). Mistral-Nemo, on the other hand, focuses much more on the year columns that are relevant to answering the question correctly. A key difference also lies in the tokenisation: While Mistral-Nemo splits all numbers into single digits, Llama3 often uses three-digit tokens where the fourth digit of a year is cut off. We assume that this makes it harder for Llama3 to process the marginal differences correctly.

The log-probabilities for the generated tokens are a proxy for the model's confidence. Here, we observe high uncertainty in Llama3 generating the core of the answer, the number token \emph{``42''}, which is incorrect. Mistral-Nemo, on the contrary, correctly answers the question and we can see that it is certain about it from the high log-probabilities. Additionally, the model shows high confidence in the row \emph{``Routing''} and column \emph{``Change''} as the location of the answer, which indeed corresponds to the true position of the value (see also Figure~\ref{fig: interp_fin-id-11-image.png} in Appendix~\ref{appendix: interp}). At the same time, it is uncertain about optional, meaning-preserving generations such as the token \emph{``provided''} as a qualifier for \emph{``table''} and the beginning of the second sentence following the answer which serves as a rationale for the model's decision-making \cite{lu-2024-tart}.

Appendix~\ref{appendix: interp} shows five more examples for ComTQA and LogicNLG instances. We also observe a repeating pattern of the start and end of a prompt being attributed the most. While these observations are based on a small set of instances, our pipeline enables computing saliency maps for any combination of prompt, input format, model, and dataset in future experiments.

\section{Related work}
\label{related_work}

Earlier TU studies leverage LLMs by representing tables as sequential text, either through naïve linearisation or by incorporating delimiters and special tokens \cite{fang2024largelanguagemodelsllmstabular}. Some works focus on fine-tuning LLMs to enhance TU \cite{zhang2024tablellmenablingtabulardata,zhang-etal-2024-tablellama,herzig-etal-2020-tapas,yin-etal-2020-tabert,gong-etal-2020-tablegpt,iida-etal-2021-tabbie}, while others explore LLMs' table reasoning abilities through prompt engineering \cite{zhao-etal-2023-large, chen-2023-large, sui-2024-table-meets-llm}. However, compared to natural language, tables present unique challenges to LLMs due to their varying layout structures, feature heterogeneity, and a large number of components leading to excessively long sequences \cite{Borisov_2022}. The latter is particularly problematic, as most LLMs become inefficient due to the quadratic complexity of self-attention \cite{vaswani2017}. With recent advances in vision and multimodality research, using MLLMs for TU has gained increasing attention with models like GPT-4 \cite{openai2024gpt4technicalreport} and Gemini \cite{geminiteam2024geminifamilyhighlycapable}, being widely adopted. Although, similar to LLMs, MLLMs also struggle with understanding structured data \cite{zheng-etal-2024-multimodal}.  

Several studies examine the impact of the table representation on models' efficiency, indicating that different table formats suit specific TU tasks and LLMs at hand \cite{deng-2024-tables-as-texts-or-images, sui-2024-table-meets-llm, zhang-2024-flextaf, singha2023tabularrepresentationnoisyoperators}. For instance, \citet{sui-2024-table-meets-llm} find HTML and XML being better understood by GPT models than Markdown, JSON, and natural language with separators encoding. 
In contrast, \citet{singha2023tabularrepresentationnoisyoperators} observe that using HTML leads to lower performance for the fact-finding and transformation tasks compared to dataframe-based and JSON formats. Meanwhile, \citet{deng-2024-tables-as-texts-or-images} analyse how models' reasoning abilities vary when tables are represented as text vs.\ images showing that Gemini Pro and GPT-4 perform similarly across both modalities. 

While these studies offer insights into the effectiveness of (M)LLMs in interpreting structured data across formats, they focus primarily on non-scientific contexts like Wikipedia and finance. This is likely due to the abundance of established, large-scale datasets based on tables from these sources, including WikiTables \cite{bhagavatulawikitables2015}, ToTTo \cite{parikh-etal-2020-totto}, and TabFact, \cite{chentabfact2020}, to name a few. Furthermore, interpretability for TU tasks remains under-researched, as related works mainly consider unstructured text and are disconnected from downstream applications \cite{ferrando-2024-primer-inner-workings,tenney-2024-sequence-salience}, rarely focusing on other long-form tasks like retrieval-augmented generation \cite{qi-2024-mirage} or QA \cite{enouen-2024-textgenshap}. \citet{nguyen-2025-interpretable-llm-based-table-qa} use attributions to make tabular QA explainable but they are constrained to the text-to-SQL setup. Unlike prior studies, this paper focuses on cross-domain and cross-modality evaluation, comparing the performance and explanations of (M)LLMs on both scientific and non-scientific tables, covering image and diverse text representations of tables. 

\section{Conclusion}
We conducted an evaluation study to explore the robustness of diverse (M)LLMs on scientific vs.\ non-scientific tables across image and four text formats. The findings reveal that current models obtain decent performance across both vision and text modalities but significantly struggle with scientific tabular data. Additionally, we explored the applicability of interpretability methods to TU tasks to get insights into the decision-making of LLMs. We found feature attributions to be a useful tool for revealing model uncertainty, its attention to table structure and relevant content, and tokenisation differences which might potentially affect predictions. 

\section*{Limitations}
Although this study provides insights into the strengths and limitations of (M)LLMs in understanding tables, it has several limitations. First, we use the same prompts across (M)LLMs and do not postprocess the predictions which may contribute to lower score values. Experimenting with model-specific prompts and structured outputs using tools such as Jsonformer\footnote{\url{https://github.com/1rgs/jsonformer}} could lead to better results. Second, we rely on automatic metrics, the drawbacks of which have been well-documented previously \cite{schmidtova-etal-2024-automatic-metrics,gehrmann2023}. Third, we focus only on interpretability for the text input, while methods like CC-SHAP \cite{parcalabescu-frank-2025-cc-shap-vlm} remain the next step to measure the importance of each modality in MLLM decision-making. Fourth, annotating all subsets in TableEval for a common task and evaluating (M)LLMs on the entire corpus could be beneficial and we leave it for future work. Finally, the dataset is limited to the English language and thus does not allow for the assessment of multilingual TU.  

\section*{Ethics statement}

The data used in this study is based on publicly available datasets. We adhere to their respective licenses and conditions of use in our experiments. Additional table formats are generated with Python scripts and open-access tools or collected from the original table sources which are under permissive licenses. All (M)LLMs, except Gemini-2.0-Flash, employed for the experiments are open-access. Those models might potentially possess biases, as outlined by their developers, which researchers should be aware of. 

\section*{Acknowledgments}

This work was supported by the consortium NFDI for Data Science and Artificial Intelligence (NFDI4DS)\footnote{\url{https://www.nfdi4datascience.de}} as part of the non-profit association National Research Data Infrastructure (NFDI e.\,V.). The consortium is funded by the Federal Republic of Germany and its states through the German Research Foundation (DFG) project NFDI4DS (no.~460234259). We would like to thank Melina Plakidis, Maximilian Dustin Nasert, and Shuiai Xu for their help in manually reviewing certain subsets of the data. 


\appendix
\newpage
\clearpage

\onecolumn

\section{Dataset statistics}
\label{appendix:dataset_stat}

\begin{table}[ht]  
\centering
\resizebox{\textwidth}{!}{
\begin{tabular}{lcccccccccccc}
\toprule
\multirow{3}{*}{\textbf{Dataset}} & \multicolumn{2}{c}{\textbf{Image}} & \multicolumn{2}{c}{\textbf{Dict}} & \multicolumn{2}{c}{\textbf{\LaTeX{}}} & \multicolumn{2}{c}{\textbf{HTML}} & \multicolumn{2}{c}{\textbf{XML}}  \\ 
\cmidrule(lr){2-3} 
\cmidrule(lr){4-5} 
\cmidrule(lr){6-7} 
\cmidrule(lr){8-9} 
\cmidrule(lr){10-11}
 & \textbf{Instances} & \textbf{Tables} & \textbf{Instances} & \textbf{Tables} & \textbf{Instances} & \textbf{Tables} & \textbf{Instances} & \textbf{Tables} & \textbf{Instances} & \textbf{Tables} \\
\midrule
\multicolumn{11}{c}{\emph{Scientific tables}} \\
\midrule
ComTQA (PubTables-1M) & 6232 & 932    &  6232 & 932    & 6232 & 932        & 6232 & 932  &  6232 & 932\\ 
numericNLG            & 135 & 135     & 135 & 135      & 135 & 135         & 135 & 135   & 135 & 135 \\ 
SciGen                & 1035 & 1035   &  1035 & 1035   &  928 & 928        &  985 & 985 &  961 & 961   \\ 
\textbf{Total}        & \textbf{7402} & \textbf{2102} & \textbf{7402} & \textbf{2102}  &   \textbf{7295} & \textbf{1995}  &\textbf{7352} & \textbf{2052}&  \textbf{7328} & \textbf{2028}  \\ 
\midrule
\multicolumn{11}{c}{\emph{Non-scientific tables}} \\
\midrule
ComTQA (FinTabNet) &  2838 & 659           &  2838 & 659         & 2838 & 659   &  2838 & 659  & 2838 & 659    \\ 
LogicNLG           & 917 & 184              & 917 & 184             & 917 & 184  & 917 & 184  & 917 & 184\\ 
Logic2Text         & 155 & 72                & 155 & 72               &155 & 72     & 155 & 72    &155 & 72 \\ 
\textbf{Total}     & \textbf{3910} & \textbf{915}     &  \textbf{3910} & \textbf{915}   &  \textbf{3910} & \textbf{915}  &   \textbf{3910} & \textbf{915}       & \textbf{3910} & \textbf{915}\\ 
\bottomrule
\end{tabular}}
\caption{Data distribution in the TableEval corpus for each format and subset.}
\label{table: instances_per_format}
\end{table}

\section{Dataset examples}
\label{appendix: task_ex_per_dataset}

\begin{figure*}[ht]
\begin{tcolorbox}[colback=white, colframe=purple!50, title = QA task: ComTQA (PubTables-1M), width=0.99\textwidth]

\begin{center}
\includegraphics[width=0.7\linewidth]{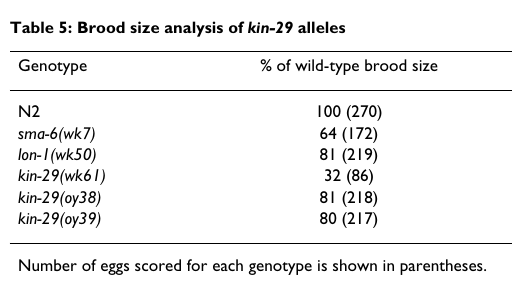}
\end{center}

\texttt{\textbf{Question:} What is the title of the table?}

\vspace{0.5em}

\texttt{\textbf{Answer:} Brood size analysis of kin-29 alleles}

\end{tcolorbox}
 \caption{An example from ComTQA (PubTables-1M), illustrating a table, a corresponding question, and a gold answer.}
  \label{fig: data_ex_pmc}
\end{figure*}

\begin{figure*}[ht]
\begin{tcolorbox}[colback=white, colframe=purple!50, title = QA task: ComTQA (FinTabNet), width=0.99\textwidth]

\begin{center}
\includegraphics[width=0.7\linewidth]{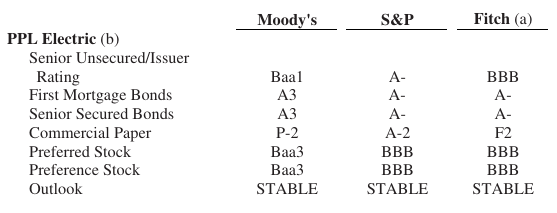}
\end{center}

\texttt{\textbf{Question:} What is the rating of commercial paper?}

\vspace{0.5em}

\texttt{\textbf{Answer:} P-2  A-2  F2}

\end{tcolorbox}
 \caption{An example from ComTQA (FinTabNet), illustrating a table, a corresponding question, and a gold answer.}
  \label{fig: data_ex_fin}
\end{figure*}

\begin{figure*}[ht]
\begin{tcolorbox}[colback=white, colframe=purple!50, title = T2T task: numericNLG, width=0.99\textwidth]

\begin{center}
\includegraphics[width=0.7\linewidth]{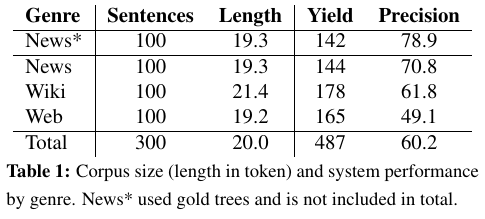}
\end{center}

\texttt{\textbf{Description:} Results. From the whole corpus of 300 sentences, PropsDE extracted 487 tuples, yielding on average 1.6 per sentence with 2.9 arguments. 60\% of them were labeled as correct. Table 1 shows that most extractions are made from Wikipedia articles, whereas the highest precision can be observed for newswire text. According to our expectations, web pages are most challenging, presumably due to noisier language. These differences between the genres can also be seen in the precision-yield curve (Figure 2).}

\end{tcolorbox}
 \caption{An example from numericNLG, illustrating a table and its corresponding gold description.}
  \label{fig: data_ex_numericnlg}
\end{figure*}

\begin{figure*}[ht]
\begin{tcolorbox}[colback=white, colframe=purple!50, title = T2T task: SciGen, width=0.99\textwidth]

\begin{center}
\includegraphics[width=0.7\linewidth]{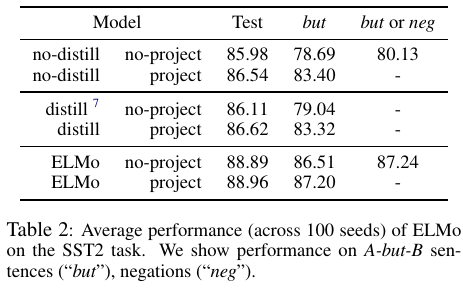}
\end{center}

\texttt{\textbf{Description:} Switching to ELMo word embeddings improves performance by 2.9 percentage points on an average, corresponding to about 53 test sentences. Of these, about 32 sentences (60\% of the improvement) correspond to A-but-B and negation style sentences, [CONTINUE] As further evidence that ELMo helps on these specific constructions, the non-ELMo baseline model (no-project, no-distill) gets 255 sentences wrong in the test corpus on average, only 89 (34.8\%) of which are A-but-B style or negations.}

\end{tcolorbox}
 \caption{An example from SciGen, illustrating a table and its corresponding gold description.}
  \label{fig: data_ex_scigen}
\end{figure*}

\begin{figure*}[ht]
\begin{tcolorbox}[colback=white, colframe=purple!50, title = T2T task: LogicNLG, width=0.99\textwidth]

\begin{center}
\includegraphics[width=0.7\linewidth]{figs/logicnlg_ex.pdf}
\end{center}

\texttt{\textbf{Title:} black ice (album)}

\vspace{0.5em}

\texttt{\textbf{Template:} the album [ENT] was first released in [ENT]}

\vspace{0.5em}

\texttt{\textbf{Statement:} the album Black Ice was first released in Europe. }

\end{tcolorbox}
 \caption{An example from LogicNLG, illustrating a table, a statement with masked entities, and a corresponding gold statement.}
  \label{fig: data_ex_logicnlg}
\end{figure*}

\newpage
\clearpage
\begin{figure*}[ht]
\begin{tcolorbox}[colback=white, colframe=purple!50, title = T2T task: Logic2Text, width=0.99\textwidth]

\begin{center}
\includegraphics[width=0.7\linewidth]{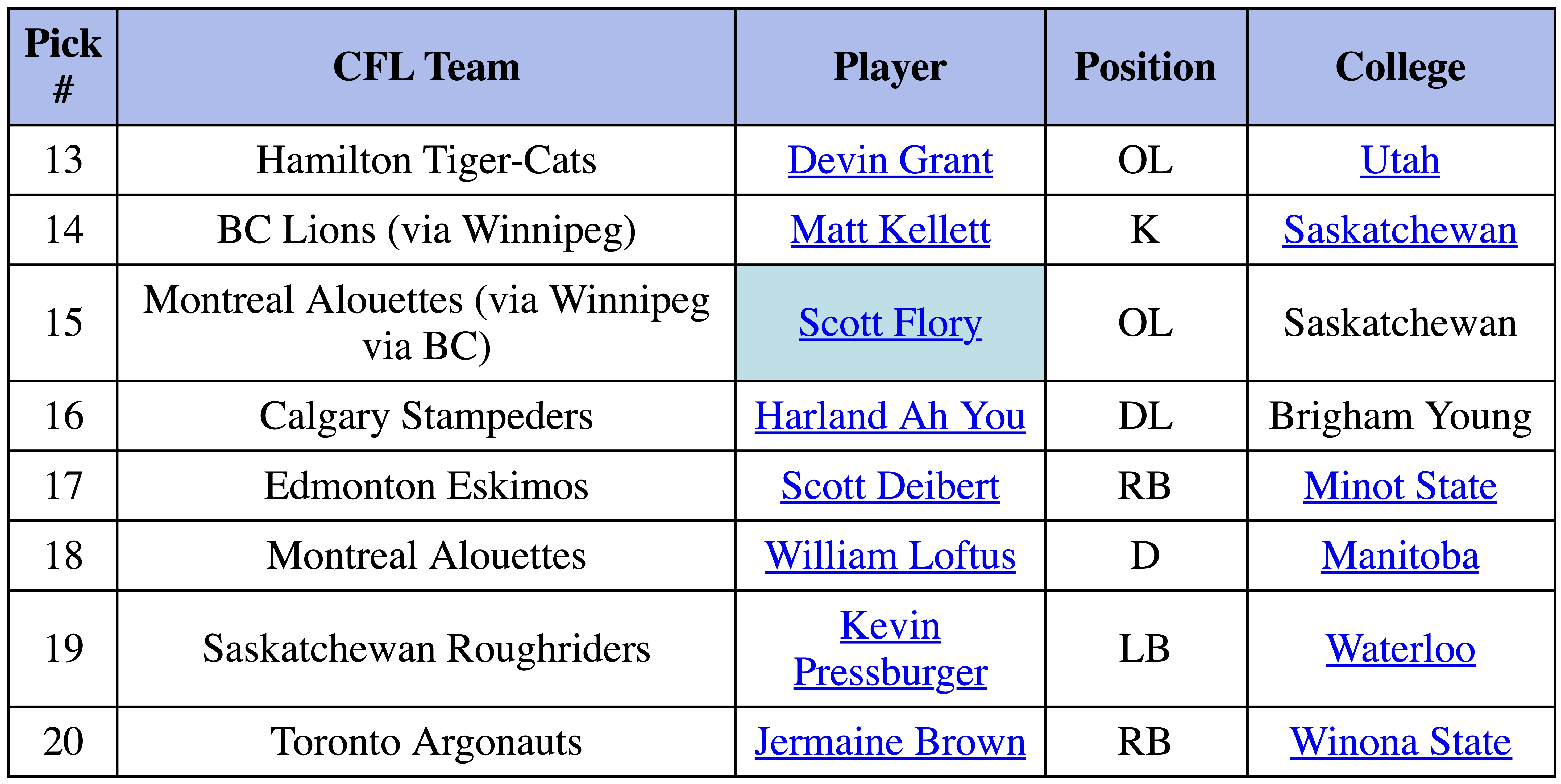}
\end{center}

\texttt{\textbf{Title:} 1998 cfl draft}

\vspace{0.5em}

\texttt{\textbf{Logical form:} and \{ only \{ filter\_eq \{ filter\_eq \{ all\_rows ; college ; saskatchewan \} ; position ; k \} \} ; eq \{ hop \{ filter\_eq \{ filter\_eq \{ all\_rows ; college ; saskatchewan \} ; position ; k \} ; player \} ; matt kellett \} \} = true}

\vspace{0.5em}

\texttt{\textbf{Statement:} the only kicker drafted by saskatchewan college in the 1998 cfl draft was matt kellett .}

\end{tcolorbox}
 \caption{An example from Logic2Text, illustrating a table, a logical form, and a corresponding gold statement.}
  \label{fig: data_ex_logic2text}
\end{figure*}

\twocolumn
\newpage
\clearpage

\section{Table formats collection}
\label{appendix: additiona_info_formats}

In what follows, we provide additional details on the collection process of the table formats. 

\textbf{XML and HTML}. As was mentioned in~\S\ref{table_formats}, XML and XML/HTML for the PubTables-1M subset of ComTQA and SciGen, respectively, are extracted from the source papers. For the former, the target tables are identified based on their titles and the highest cosine similarity with table content annotations available in PubTables-1M. For Scigen we use the fuzzy match score with a threshold of 0.8 to identify the relevant tables based on their captions. Note that not all instances have these formats (see Table~\ref{table: instances_per_format}) due to \LaTeX{ML} conversion errors, low fuzzy match score, discrepancies between captions in the gold data and \LaTeX{} files or a scholarly paper not being available on arXiv anymore. We also exclude cases with multiple tables sharing the same caption but annotated separately, as it is challenging to accurately link the corresponding HTML/XML code for each table. HTML in LogicNLG and Logic2Text are retrieved from the Wikipedia pages. However, due to the lack of metadata on the data collection timestamps, we choose a time interval close to the year of publication of these datasets for our search in the Wikipedia archive. To extract the relevant tables, we employ a cosine similarity comparison against the gold tables, using a threshold of 0.9. Since Wikipedia is constantly updated, we further manually check the results and filter out cases where the mismatch affects the ground truth, e.\,g., cell values being out of date or the removal/addition of both rows and columns. Note that for all subsets except SciGen, we follow the PMC table formatting rules\footnote{\url{https://www.ncbi.nlm.nih.gov/pmc/pmcdoc/tagging-guidelines/article/dobs.html\#dob-tables}} to obtain XML. Additionally, all generated HTML underwent automatic validation using the PyTidyLib\footnote{\url{https://countergram.github.io/pytidylib/}} package.

\textbf{\LaTeX{}}. Similar to HTML/XML, we obtain \LaTeX{} from the source scholarly papers in SciGen (see~\S\ref{table_formats}) and extract the target tables based on their captions using the fuzzy match. Some instances are excluded due to low similarity scores (below 0.8), parsing errors or lack of \LaTeX{} source code (tables from ACL papers). For numericNLG and PubTables-1M tables, \LaTeX{} is generated from HTML. This process involves preprocessing the HTML code to replace symbols, such as Greek letters and mathematical operators, with their \LaTeX{} equivalents. The resulting HTML is then converted to a dataframe and subsequently to \LaTeX{} using pandas.

\textbf{Dict}. The conventions of already available linearised tables in SciGen, numericNLG, LogicNLG, and Logic2Text are slightly diverse. In particular, the distinction between column and row heads exists only in numericNLG. Furthermore, compared to LogicNLG and Logic2Text, header hierarchy is preserved in numericNLG and SciGen by merging headers and subheaders into a single string. 

\onecolumn
\newpage
\clearpage

\section{Image aspect ratios}
\label{appendix: aspect_ratios}
\begin{figure*}[ht] 
    \center{\includegraphics[width=0.6\textwidth]{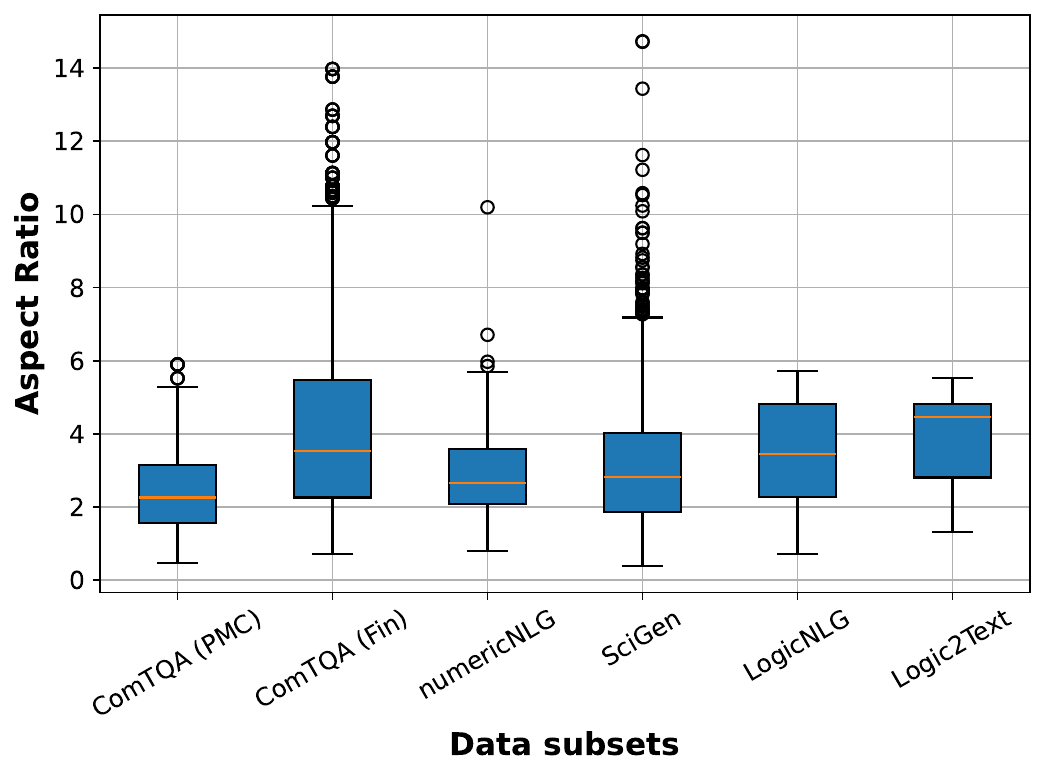}}
    \caption{Distribution of image aspect ratios (width/height) across subsets in the TableEval benchmark. Each box represents the interquartile range (IQR), with the central orange line indicating the median. Circles denote outliers, while whiskers (set to 1.5 × IQR by default) extend to the minimum and maximum non-outlier values. Here ``Fin'' stands for FinTabNet, while ``PMC'' denotes PubTables-1M.}
    \label{fig: aspect_ratio}
\end{figure*}

\newpage
\clearpage

\section{Prompts}
\label{appendix: prompts}

\begin{figure*}[ht]
\begin{tcolorbox}[colback=white, colframe=purple!50, title=~, width=\textwidth]
\raggedright
\textbf{ComTQA (FinTabNet):} \\
\vspace{0.5em}
\texttt{Refer to the provided table and answer the question. Question: \{question\}} \\
\vspace{0.5em}
\textbf{ComTQA (PubTables-1M):} \\
\vspace{0.5em}
\texttt{Refer to the provided table and answer the question. Question: \{question\}. Table caption: \{caption\}. Table footnote: \{footnote\}.} \\
\vspace{0.5em}
\textbf{SciGen:} \\
\vspace{0.5em}
\texttt{Describe the given table focusing on the most important findings reported by reasoning over its content. The summary must be factual, coherent, and well-written. Do not introduce new information or speculate. Table caption: \{caption\}} \\
\vspace{0.5em}
\textbf{numericNLG:} \\
\vspace{0.5em}
\texttt{Describe the given table focusing on the insights and trends revealed by the results. The summary must be factual, coherent, and well-written. Do not introduce new information or speculate. Table caption: \{caption\}}\\
\vspace{0.5em}
\textbf{Logic2Text:} \\
\vspace{0.5em}
\texttt{Generate a one sentence statement based on the table and logical form. Logical form: \{logical\_form\}. Table title: \{title\}}\\
\vspace{0.5em}
\textbf{LogicNLG:} \\
\vspace{0.5em}
\texttt{Based on a given table, fill in the entities masked by [ENT] in the following sentence: \{sentence\}. Output the sentence with filled in masked entities. Table title: \{title\}} \\
\end{tcolorbox}
 \caption{Prompts used for experiments based on images of tables.}
  \label{fig:promts_image}
\end{figure*}

\newpage

\begin{figure*}[ht]
\begin{tcolorbox}[colback=white, colframe=purple!50, title=~, width=\textwidth]
\raggedright
\vspace{0.5em}
\textbf{ComTQA (FinTabNet):} \\
\vspace{0.5em}
\texttt{Refer to the provided table and answer the question. Question: \{question\}. Table: \{table\}.} \\
\vspace{0.5em}
\textbf{ComTQA (PubTables-1M):} \\
\vspace{0.5em}
\texttt{Refer to the provided table and answer the question. Question: \{question\}. Table: \{table\}.} \\
\vspace{0.5em}
\textbf{SciGen:} \\
\vspace{0.5em}
\texttt{Describe the given table focusing on the most important findings reported by reasoning over its content. The summary must be factual, coherent, and well-written. Do not introduce new information or speculate. Table: \{table\}.} \\
\vspace{0.5em}
\textbf{numericNLG:} \\
\vspace{0.5em}
\texttt{Describe the given table focusing on the insights and trends revealed by the results. The summary must be factual, coherent, and well-written. Do not introduce new information or speculate. Table: \{table\}.} \\
\vspace{0.5em}
\textbf{Logic2Text:} \\
\vspace{0.5em}
\texttt{Generate a one sentence statement based on the table and logical form. Logical form: \{logical\_form\}. Table title: \{title\}. Table: \{table\}.} \\
\vspace{0.5em}
\textbf{LogicNLG:} \\
\vspace{0.5em}
\texttt{Based on a given table, fill in the entities masked by [ENT] in the following sentence: \{sentence\}. Output the sentence with filled in masked entities. Table title: \{title\}. Table: \{table\}.} \\
\end{tcolorbox}
 \caption{Prompts used for experiments based on textual representations of tables.}
  \label{fig:prompts_text}
\end{figure*}

\twocolumn
\newpage
\clearpage

\section{Experimental results}
\label{appendix: tables_results}

\begin{table}[ht]  
\centering
\resizebox{\columnwidth}{!}{
\begin{tabular}{lccccc}
\toprule
\textbf{Metric}&\textbf{Dict}&\textbf{HTML}&\textbf{Image}&\textbf{\LaTeX{}}&\textbf{XML} \\
\midrule
BertScore.F1	&0.85&	0.84&	\textbf{0.86}&	0.84&	0.85 \\
BLEU-1	& 0.16&	0.15&	\textbf{0.19}	&0.16&	0.16 \\
BLEU-2&0.09	&0.09&	\textbf{0.12}&	0.09&	0.09 \\
BLEU-3	&0.06	&0.06&	\textbf{0.09}&	0.06&	0.07 \\
BLEU-4	&0.04	&0.04&	\textbf{0.06}&	0.05&	0.05 \\
BLEURT &	$-$0.51	&$-$0.55&\textbf{$-$0.42}	&$-$0.54&	$-$0.53 \\
METEOR&	0.24&	0.24&	\textbf{0.25}&	0.24&	0.24\\
MoverScore	&0.54&	0.53&	\textbf{0.56}&	0.54&	0.54 \\ 
ROUGE-1.F1	&0.30&	0.29&	\textbf{0.38}&	0.29&	0.29\\ 
ROUGE-2.F1	&0.15&	0.14&\textbf{0.20}&	0.15&	0.15\\ 
ROUGE-3.F1	&0.09&	0.09&	\textbf{0.12}&	0.09&	0.09 \\
ROUGE-4.F1	&0.06&	0.06&	\textbf{0.08}&	0.07&	0.06 \\
ROUGE-L.F1&	0.24&	0.23	&\textbf{0.32}	&0.24&	0.24 \\
SacreBLEU	&0.04	&0.04&\textbf{0.08}&0.05&	0.05 \\
\bottomrule
\end{tabular}}
\caption{Values across evaluation metrics for table formats averaged over data subsets and models.}
\label{table: scores_per_format}
\end{table}

\begin{table}[ht]  
\centering
\resizebox{\columnwidth}{!}{
\begin{tabular}{lccccc}
\toprule
\textbf{Metric}&\textbf{Dict}&\textbf{HTML}&\textbf{Image}&\textbf{\LaTeX{}}&\textbf{XML} \\
\midrule
BertScore.F1 &	0.83      &      0.84      &      \textbf{0.86}      &      0.83      &      0.84       \\
BLEU-1	&0.02      &      0.02      &      \textbf{0.05}      &      0.02      &      0.02       \\
BLEU-2&0.01      &      0.01      &      \textbf{0.03}      &      0.01      &      0.01       \\
BLEU-3	&0.01      &      0.01      &      \textbf{0.02}      &      0.01      &      0.01       \\
BLEU-4	&0.01      &      0.01      &      \textbf{0.02}      &      0.01      &      0.01       \\
BLEURT &	$-$0.58      &      $-$0.55      &      \textbf{$-$0.39}      &      $-$0.59      &      $-$0.54      \\
METEOR&	0.06      &      0.07      &      \textbf{0.08}      &      0.06      &      0.07       \\
MoverScore	&0.50      &      0.50      &      \textbf{0.53}      &      0.49      &      0.50       \\
ROUGE-1.F1	&0.14      &      0.14      &      \textbf{0.27}      &      0.14      &      0.15       \\
ROUGE-2.F1	&0.08      &      0.08      &      \textbf{0.17}      &      0.08      &      0.09       \\
ROUGE-3.F1	&0.03      &      0.03      &      \textbf{0.05}      &      0.03      &      0.03       \\
ROUGE-4.F1	&0.01      &      0.01      &      \textbf{0.02}      &      0.01      &      0.01       \\
ROUGE-L.F1&	0.13      &      0.14      &      \textbf{0.27}      &      0.14      &      0.15       \\
SacreBLEU	&0.01      &      0.02      &      \textbf{0.04}      &      0.01      &      0.02       \\
\bottomrule
\end{tabular}}
\caption{Raw values of BertScore.F1, BLEU-N.F1, BLEURT, METEOR, MoverScore, ROUGE-N.F1, ROUGE-L.F1, and SacreBLEU for ComTQA (FinTabNet) subset for individual formats averaged over models.}
\label{table: raw_scores_per_format_comtqa_fin}
\end{table}

\begin{table}[ht]  
\centering
\resizebox{\columnwidth}{!}{
\begin{tabular}{lccccc}
\toprule
\textbf{Metric}&\textbf{Dict}&\textbf{HTML}&\textbf{Image}&\textbf{\LaTeX{}}&\textbf{XML} \\
\midrule
BertScore.F1 &0.82      &      0.82      &      \textbf{0.85}      &      0.82      &      0.82       \\
BLEU-1	&0.03      &      0.03      &      \textbf{0.05}      &      0.03      &      0.03       \\
BLEU-2&0.02      &      0.02      &      \textbf{0.03}      &      0.02      &      0.02       \\
BLEU-3	&0.01      &      \textbf{0.02}      &      \textbf{0.02}      &      0.01      &      \textbf{0.02}       \\
BLEU-4	&0.01      &      0.01      &      \textbf{0.02}      &      0.01      &      0.01       \\
BLEURT &	$-$0.73      &      $-$0.72      &      \textbf{$-$0.59}      &      $-$0.73      &      $-$0.72       \\
METEOR&	\textbf{0.09}      &      0.10      &      \textbf{0.09}      &      \textbf{0.09}      &      0.10       \\
MoverScore	&0.48      &      0.48      &      \textbf{0.51}      &      0.48      &      0.48       \\
ROUGE-1.F1	&0.12      &      0.12      &      \textbf{0.22}      &      0.12      &      0.12       \\
ROUGE-2.F1	&0.06      &      0.06      &      \textbf{0.11}      &      0.06      &      0.06       \\
ROUGE-3.F1	&0.03      &      0.03      &      \textbf{0.04}      &      0.03      &      0.03       \\
ROUGE-4.F1	&0.02      &      0.02      &      \textbf{0.03}      &      0.02      &      0.02       \\
ROUGE-L.F1&	0.12      &      0.12      &      \textbf{0.22}      &      0.11      &      0.12       \\
SacreBLEU	&0.01      &      0.01      &      \textbf{0.04}      &      0.01      &      0.01       \\
\bottomrule
\end{tabular}}
\caption{Raw values of BertScore.F1, BLEU-N.F1, BLEURT, METEOR, MoverScore, ROUGE-N.F1, ROUGE-L.F1, and SacreBLEU for ComTQA (PubTables-1M) subset for individual formats averaged over models.}
\label{table: raw_scores_per_format_comtqa_pmc}
\end{table}

\begin{table}[ht]  
\centering
\resizebox{\columnwidth}{!}{
\begin{tabular}{lccccc}
\toprule
\textbf{Metric}&\textbf{Dict}&\textbf{HTML}&\textbf{Image}&\textbf{\LaTeX{}}&\textbf{XML} \\
\midrule
BertScore.F1 &0.88      &      0.88      &      \textbf{0.89}      &      0.88      &      0.88       \\
BLEU-1	&\textbf{0.24}      &      \textbf{0.24}      &      0.22      &      \textbf{0.24}      &      \textbf{0.24}       \\
BLEU-2&\textbf{0.13}      &      \textbf{0.13}      &      0.12      &      \textbf{0.13}      &      \textbf{0.13}       \\
BLEU-3	&0.07      &      0.07      &      0.07      &      0.07      &      \textbf{0.08}       \\
BLEU-4	&0.04      &      0.04      &      0.04      &      0.04      &      \textbf{0.05}       \\
BLEURT &	$-$0.14      &      $-$0.11      &      $-$0.19      &      \textbf{$-$0.09}      &      \textbf{$-$0.09}       \\
METEOR&	0.35      &      0.37      &      0.33      &      0.37      &      \textbf{0.38}       \\
MoverScore	&0.59      &      \textbf{0.60}      &      \textbf{0.60}      &      \textbf{0.60}      &      \textbf{0.60}       \\
ROUGE-1.F1	&0.48      &      \textbf{0.49}      &      \textbf{0.49}      &      \textbf{0.49}      &      \textbf{0.49}       \\
ROUGE-2.F1	&0.23      &      0.24      &      0.24      &      \textbf{0.25}      &      0.24       \\
ROUGE-3.F1	&0.12      &      0.13      &      0.12      &      \textbf{0.14}      &      0.13       \\
ROUGE-4.F1	&0.06      &      0.07      &      0.07      &      \textbf{0.08}      &      0.07       \\
ROUGE-L.F1&	0.37      &      \textbf{0.39}      &      \textbf{0.39}      &      0.38      &      0.38       \\
SacreBLEU	&\textbf{0.05}      &      \textbf{0.05}      &      \textbf{0.05}      &      \textbf{0.05}      &      \textbf{0.05}       \\
\bottomrule
\end{tabular}}
\caption{Raw values of BertScore.F1, BLEU-N.F1, BLEURT, METEOR, MoverScore, ROUGE-N.F1, ROUGE-L.F1, and SacreBLEU for Logic2Text subset for individual formats averaged over models.}
\label{table: raw_scores_per_format_logic2text}
\end{table}

\begin{table}[ht]  
\centering
\resizebox{\columnwidth}{!}{
\begin{tabular}{lccccc}
\toprule
\textbf{Metric}&\textbf{Dict}&\textbf{HTML}&\textbf{Image}&\textbf{\LaTeX{}}&\textbf{XML} \\
\midrule
BertScore.F1 &0.87      &      0.88      &      \textbf{0.91}      &      0.89      &      0.88       \\
BLEU-1	&0.32      &      0.33      &      \textbf{0.51}      &      0.36      &      0.36       \\
BLEU-2&0.26      &      0.27      &      \textbf{0.43}      &      0.30      &      0.29       \\
BLEU-3	&0.21      &      0.23      &      \textbf{0.35}      &      0.25      &      0.24       \\
BLEU-4	&0.17      &      0.18      &      \textbf{0.28}      &      0.20      &      0.20       \\
BLEURT &	$-$0.46      &      $-$0.47      &      \textbf{$-$0.13}      &      $-$0.40      &      $-$0.41       \\
METEOR&	0.52      &      0.53      &      \textbf{0.63}      &      0.55      &      0.55       \\
MoverScore	&0.60      &      0.59      &      \textbf{0.64}      &      0.61      &      0.60       \\
ROUGE-1.F1	&0.48      &      0.48      &      \textbf{0.69}      &      0.52      &      0.51       \\
ROUGE-2.F1	&0.38      &      0.38      &      \textbf{0.55}      &      0.41      &      0.40       \\
ROUGE-3.F1	&0.31      &      0.30      &      \textbf{0.45}      &      0.34      &      0.33       \\
ROUGE-4.F1	&0.25      &      0.25      &      \textbf{0.37}      &      0.28      &      0.27       \\
ROUGE-L.F1&	0.46      &      0.47      &      \textbf{0.67}      &      0.51      &      0.49       \\
SacreBLEU	&0.13      &      0.15      &      \textbf{0.28}      &      0.16      &      0.16       \\
\bottomrule
\end{tabular}}
\caption{Raw values of BertScore.F1, BLEU-N.F1, BLEURT, METEOR, MoverScore, ROUGE-N.F1, ROUGE-L.F1, and SacreBLEU for LogicNLG subset for individual formats averaged over models.}
\label{table: raw_scores_per_format_logicnlg}
\end{table}

\begin{table}[ht]  
\centering
\resizebox{\columnwidth}{!}{
\begin{tabular}{lccccc}
\toprule
\textbf{Metric}&\textbf{Dict}&\textbf{HTML}&\textbf{Image}&\textbf{\LaTeX{}}&\textbf{XML} \\
\midrule
BertScore.F1 &0.83      &      \textbf{0.84}      &      0.83      &      \textbf{0.84}      &      \textbf{0.84}  \\
BLEU-1	&0.16      &      \textbf{0.18}      &      0.16      &      \textbf{0.18}      &      \textbf{0.18}      \\
BLEU-2&0.06      &      \textbf{0.07}      &      \textbf{0.07}      &      \textbf{0.07}      &      \textbf{0.07}       \\
BLEU-3	&\textbf{0.03}      &      \textbf{0.03}      &      \textbf{0.03}      &      \textbf{0.03}      &      \textbf{0.03}       \\
BLEU-4	&0.01      &      \textbf{0.02}      &      0.01      &      \textbf{0.02}      &      \textbf{0.02}       \\
BLEURT &	$-$0.58      &      $-$0.54      &      $-$0.60      &      $-$0.54      &      \textbf{$-$0.53}       \\
METEOR&0.19      &      \textbf{0.21}      &      0.19      &      \textbf{0.21}      &      \textbf{0.21}       \\
MoverScore	&0.52      &      \textbf{0.53}      &      \textbf{0.53}      &      \textbf{0.53}      &      \textbf{0.53}       \\
ROUGE-1.F1	&0.28      &      0.31      &      0.30      &      \textbf{0.32}      &      \textbf{0.32}       \\   
ROUGE-2.F1	&0.06      &      \textbf{0.08}      &      0.07      &      \textbf{0.08}      &      \textbf{0.08}       \\
ROUGE-3.F1	&0.02      &      0.02      &      0.02      &      0.02      &      \textbf{0.03}       \\
ROUGE-4.F1	&\textbf{0.01}      &      \textbf{0.01}      &      \textbf{0.01}      &      \textbf{0.01}      &      \textbf{0.01}       \\
ROUGE-L.F1&	0.16      &      \textbf{0.17}      &      \textbf{0.17}      &      \textbf{0.17}      &      \textbf{0.17} \\
SacreBLEU	&\textbf{0.03}      &      \textbf{0.03}      &      \textbf{0.03}      &      \textbf{0.03}      &      \textbf{0.03}       \\
\bottomrule
\end{tabular}}
\caption{Raw values of BertScore.F1, BLEU-N.F1, BLEURT, METEOR, MoverScore, ROUGE-N.F1, ROUGE-L.F1, and SacreBLEU for numericNLG subset for individual formats averaged over models.}
\label{table: raw_scores_per_format_numericnlg}
\end{table}

\begin{table}[ht]  
\centering
\resizebox{\columnwidth}{!}{
\begin{tabular}{lccccc}
\toprule
\textbf{Metric}&\textbf{Dict}&\textbf{HTML}&\textbf{Image}&\textbf{\LaTeX{}}&\textbf{XML} \\
\midrule
BertScore.F1 &\textbf{0.84}      &      0.81      &      \textbf{0.84}      &      0.81      &      0.81       \\
BLEU-1	&\textbf{0.16}      &      0.11      &      0.15      &      0.11      &      0.11       \\
BLEU-2&\textbf{0.07}      &      0.03      &      \textbf{0.07}      &      0.03      &      0.03       \\
BLEU-3	&\textbf{0.03}      &      0.01      &      \textbf{0.03}      &      0.01      &      0.01       \\
BLEU-4	&\textbf{0.02}      &      0.00      &      \textbf{0.02}      &      0.00      &      0.00       \\
BLEURT &	\textbf{$-$0.59}      &      $-$0.90      &      $-$0.64      &      $-$0.91      &      $-$0.90       \\
METEOR&\textbf{0.20}      &      0.13      &      0.19      &      0.13      &      0.13       \\
MoverScore	&\textbf{0.53}      &      0.50      &      \textbf{0.53}      &      0.50      &      0.50       \\
ROUGE-1.F1	&\textbf{0.30}      &      0.18      &      0.29      &      0.18      &      0.18       \\  
ROUGE-2.F1	&\textbf{0.07}      &      0.02      &      \textbf{0.07}      &      0.02      &      0.02       \\
ROUGE-3.F1	&0.02      &      0.00      &      \textbf{0.03}      &      0.00      &      0.00       \\
ROUGE-4.F1	&\textbf{0.01}      &      0.00      &      \textbf{0.01}      &      0.00      &      0.00       \\
ROUGE-L.F1&	\textbf{0.17}      &      0.11      &      \textbf{0.17}      &      0.11      &      0.11       \\
SacreBLEU	& \textbf{0.03}      &      0.01      &      \textbf{0.03}      &      0.01      &      0.01       \\
\bottomrule
\end{tabular}}
\caption{Raw values of BertScore.F1, BLEU-N.F1, BLEURT, METEOR, MoverScore, ROUGE-N.F1, ROUGE-L.F1, and SacreBLEU for SciGen subset for individual formats averaged over models.}
\label{table: raw_scores_per_format_scigen}
\end{table}

\begin{table}[ht]  
\centering
\resizebox{\columnwidth}{!}{
\begin{tabular}{lcccc}
\toprule
\textbf{Metric}&\textbf{Dict}&\textbf{HTML}&\textbf{\LaTeX{}}&\textbf{XML} \\
\midrule
BertScore.F1 &\textbf{0.83}      &      \textbf{0.83}      &      \textbf{0.83}      &      \textbf{0.83}       \\
BLEU-1	&\textbf{0.12}      &      \textbf{0.12}      &      0.11      &      0.11       \\
BLEU-2&\textbf{0.06}      &      \textbf{0.06}      &      \textbf{0.06}      &      \textbf{0.06}       \\
BLEU-3	&0.03      &      \textbf{0.04}      &      \textbf{0.04}      &      \textbf{0.04}       \\
BLEU-4	&\textbf{0.02}      &      \textbf{0.02}      &      \textbf{0.02}      &      \textbf{0.02}       \\
BLEURT &	\textbf{$-$0.64}      &      $-$0.67      &      $-$0.67      &      $-$0.66       \\
METEOR&0.20      &      \textbf{0.21}      &      0.20      &      \textbf{0.21}       \\
MoverScore	&\textbf{0.52}      &      \textbf{0.52}      &      \textbf{0.52}      &      \textbf{0.52}       \\
ROUGE-1.F1	&\textbf{0.23}      &      \textbf{0.23}      &      \textbf{0.23}      &      \textbf{0.23}       \\ 
ROUGE-2.F1	&0.09      &      \textbf{0.10}      &      \textbf{0.10}      &      \textbf{0.10}       \\
ROUGE-3.F1	&\textbf{0.05}      &      \textbf{0.05}      &      \textbf{0.05}      &      \textbf{0.05}       \\
ROUGE-4.F1	&\textbf{0.03}      &      \textbf{0.03}      &      \textbf{0.03}      &      \textbf{0.03}       \\
ROUGE-L.F1&	0.17      &      \textbf{0.18}      &      \textbf{0.18}      &      \textbf{0.18}       \\
SacreBLEU	&\textbf{0.02}      &      \textbf{0.02}      &      \textbf{0.02}      &      \textbf{0.02}       \\
\bottomrule
\end{tabular}}
\caption{Raw values of BertScore.F1, BLEU-N.F1, BLEURT, METEOR, MoverScore, ROUGE-N.F1, ROUGE-L.F1, and SacreBLEU for Llama-3.2-3B-Instruct and individual text formats averaged over data subsets.}
\label{table: raw_scores_per_llm_format_llama}
\end{table}

\begin{table}[ht]  
\centering
\resizebox{\columnwidth}{!}{
\begin{tabular}{lcccc}
\toprule
\textbf{Metric}&\textbf{Dict}&\textbf{HTML}&\textbf{\LaTeX{}}&\textbf{XML} \\
\midrule
BertScore.F1 &\textbf{0.85}      &      \textbf{0.85}      &      \textbf{0.85}      &      \textbf{0.85}       \\
BLEU-1	&0.17      &      0.15      &      \textbf{0.18}      &      0.17       \\
BLEU-2&0.10      &      0.09      &      \textbf{0.11}      &      0.10       \\
BLEU-3	&0.06      &      0.06      &      \textbf{0.07}      &      \textbf{0.07}       \\
BLEU-4	&0.04      &      0.04      &      \textbf{0.05}      &      \textbf{0.05}       \\
BLEURT &	\textbf{$-$0.48}      &      $-$0.54      &      \textbf{$-$0.48}      &      $-$0.49       \\
METEOR&\textbf{0.25}      &      0.24      &      \textbf{0.25}      &      \textbf{0.25}       \\
MoverScore	&\textbf{0.54}      &      \textbf{0.54}      &      \textbf{0.54}      &      \textbf{0.54}       \\
ROUGE-1.F1	&0.33      &      0.31      &      \textbf{0.34}      &      0.33       \\
ROUGE-2.F1	&0.17      &      0.16      &      \textbf{0.18}      &      \textbf{0.18}       \\
ROUGE-3.F1	&\textbf{0.11}      &      0.10      &      \textbf{0.11}      &      \textbf{0.11}       \\
ROUGE-4.F1	&0.07      &      0.07      &      \textbf{0.08}      &      \textbf{0.08}       \\
ROUGE-L.F1&0.27      &      0.26      &      \textbf{0.28}      &      \textbf{0.28}       \\
SacreBLEU	&0.04      &      0.04      &      \textbf{0.05}      &      \textbf{0.05}       \\
\bottomrule
\end{tabular}}
\caption{Raw values of BertScore.F1, BLEU-N.F1, BLEURT, METEOR, MoverScore, ROUGE-N.F1, ROUGE-L.F1, and SacreBLEU for Mistral-Nemo-Instruct-2407 and individual text formats averaged over data subsets.}
\label{table: raw_scores_per_llm_format_mistral}
\end{table}

\begin{table}[ht]  
\centering
\resizebox{\columnwidth}{!}{
\begin{tabular}{lcccc}
\toprule
\textbf{Metric}&\textbf{Dict}&\textbf{HTML}&\textbf{\LaTeX{}}&\textbf{XML} \\
\midrule
BertScore.F1 &\textbf{0.84}      &      \textbf{0.84}      &      \textbf{0.84}      &      \textbf{0.84}       \\
BLEU-1	&\textbf{0.13}      &      \textbf{0.13}      &      \textbf{0.13}      &      \textbf{0.13}       \\
BLEU-2&\textbf{0.07}      &      \textbf{0.07}      &      \textbf{0.07}      &      \textbf{0.07}       \\
BLEU-3	&0.04      &      \textbf{0.05}      &      \textbf{0.05}      &      \textbf{0.05}       \\
BLEU-4	&\textbf{0.03}      &      \textbf{0.03}      &      \textbf{0.03}      &      \textbf{0.03}       \\
BLEURT &	\textbf{$-$0.54}      &      $-$0.55      &      $-$0.57      &      $-$0.56       \\
METEOR&0.23      &      \textbf{0.24}      &      0.23      &      \textbf{0.24}       \\
MoverScore	&\textbf{0.53}      &      \textbf{0.53}      &      \textbf{0.53}      &      \textbf{0.53}       \\
ROUGE-1.F1	&\textbf{0.26}      &      \textbf{0.26}      &      \textbf{0.26}      &      \textbf{0.26}       \\
ROUGE-2.F1	&0.12      &      \textbf{0.13}      &    0.12      &      \textbf{0.13}       \\
ROUGE-3.F1	&\textbf{0.07}      &      \textbf{0.07}      &      \textbf{0.07}      &      \textbf{0.07}       \\
ROUGE-4.F1	&\textbf{0.05}      &      \textbf{0.05}      &      \textbf{0.05}      &      \textbf{0.05}       \\
ROUGE-L.F1&0.20      &      \textbf{0.21}      &      0.20      &      0.20       \\
SacreBLEU	&\textbf{0.03}      &      \textbf{0.03}      &      \textbf{0.03}      &      \textbf{0.03}       \\
\bottomrule
\end{tabular}}
\caption{Raw values of BertScore.F1, BLEU-N.F1, BLEURT, METEOR, MoverScore, ROUGE-N.F1, ROUGE-L.F1, and SacreBLEU for Qwen2.5-14B-Instruct and individual text formats averaged over data subsets.}
\label{table: raw_scores_per_llm_format_qwen14}
\end{table}

\begin{table}[ht]  
\centering
\resizebox{\columnwidth}{!}{
\begin{tabular}{lcccc}
\toprule
\textbf{Metric}&\textbf{Dict}&\textbf{HTML}&\textbf{\LaTeX{}}&\textbf{XML} \\
\midrule
BertScore.F1 &\textbf{0.84}      &      \textbf{0.84}      &      \textbf{0.84}      &      \textbf{0.84}       \\
BLEU-1	&\textbf{0.16}      &      0.15      &      \textbf{0.16}      &      0.15       \\
BLEU-2&\textbf{0.09}      &      0.08      &      \textbf{0.09}      &      \textbf{0.09}       \\
BLEU-3	&0.06      &      0.06      &      \textbf{0.07}      &      0.06       \\
BLEU-4	&0.04      &      0.04      &      \textbf{0.05}      &      \textbf{0.05}       \\
BLEURT &\textbf{$-$0.54}      &      $-$0.59      &      $-$0.57      &      $-$0.57       \\
METEOR&\textbf{0.24}      &      0.23      &      \textbf{0.24}      &      0.23       \\
MoverScore	&\textbf{0.53}      &      \textbf{0.53}      &      \textbf{0.53}      &      \textbf{0.53}       \\
ROUGE-1.F1	&\textbf{0.28}      &      0.27      &      \textbf{0.28}      &      \textbf{0.28}       \\
ROUGE-2.F1	&0.13      &      0.13      &      \textbf{0.14}      &      0.13       \\
ROUGE-3.F1	&0.08      &      0.08      &      \textbf{0.09}      &      0.08       \\
ROUGE-4.F1	&\textbf{0.06}      &      0.05      &      \textbf{0.06}      &      \textbf{0.06}       \\
ROUGE-L.F1&0.22      &      0.21      &      \textbf{0.23}      &      0.22       \\
SacreBLEU	&0.03      &      0.03      &      \textbf{0.04}      &      0.03       \\
\bottomrule
\end{tabular}}
\caption{Raw values of BertScore.F1, BLEU-N.F1, BLEURT, METEOR, MoverScore, ROUGE-N.F1, ROUGE-L.F1, and SacreBLEU for Qwen2.5-3B-Instruct and individual text formats averaged over data subsets.}
\label{table: raw_scores_per_llm_format_qwen3}
\end{table}

\begin{table}[ht]  
\centering
\resizebox{\columnwidth}{!}{
\begin{tabular}{lcccc}
\toprule
\textbf{Metric}&\textbf{Dict}&\textbf{HTML}&\textbf{\LaTeX{}}&\textbf{XML} \\
\midrule
BertScore.F1 &\textbf{0.86}      &      \textbf{0.86}      &      \textbf{0.86}      &      \textbf{0.86}       \\
BLEU-1	&0.21      &      \textbf{0.22}      &      0.21      &      \textbf{0.22}       \\
BLEU-2&0.13      &      0.14    &    0.14      &      \textbf{0.15}       \\
BLEU-3	&0.10      &      \textbf{0.11}      &     0.10      &      \textbf{0.11}       \\
BLEU-4	&0.08      &      \textbf{0.09}      &      0.08      &      \textbf{0.09}       \\
BLEURT &\textbf{$-$0.37}      &      $-$0.39      &      $-$0.41      &      $-$0.38       \\
METEOR&0.26      &      \textbf{0.27}      &      0.26      &      \textbf{0.27}       \\
MoverScore	&\textbf{0.56}      &      \textbf{0.56}      &      0.55      &      \textbf{0.56}       \\
ROUGE-1.F1	&\textbf{0.38}      &      0.37      &      0.36      &      0.37       \\
ROUGE-2.F1	&\textbf{0.21}      &      \textbf{0.21}      &      0.20      &      \textbf{0.21}       \\
ROUGE-3.F1	&0.13      &      \textbf{0.14}      &      0.13      &      \textbf{0.14}       \\
ROUGE-4.F1	&\textbf{0.10}      &      \textbf{0.10}      &      \textbf{0.10}      &      \textbf{0.10}       \\
ROUGE-L.F1&\textbf{0.32}      &      0.31      &      0.30      &      0.31       \\
SacreBLEU	&0.09      &      0.10      &      0.10      &      \textbf{0.11}       \\
\bottomrule
\end{tabular}}
\caption{Raw values of BertScore.F1, BLEU-N.F1, BLEURT, METEOR, MoverScore, ROUGE-N.F1, ROUGE-L.F1, and SacreBLEU for Gemini-2.0-Flash and individual text formats averaged over data subsets.}
\label{table: raw_scores_per_llm_format_gemini}
\end{table}

\begin{table}[ht]  
\centering
\begin{tabular}{lccc}
\toprule
\textbf{Metric}&\textbf{Non-Scientific}&\textbf{Scientific} \\
\midrule
BertScore.F1	&\textbf{0.87}	&0.83\\
BLEU-1&	\textbf{0.21}&	0.11\\
BLEU-2&	\textbf{0.15}&	0.04\\
BLEU-3&\textbf{0.11}&	0.02\\
BLEU-4	&\textbf{0.09}	&0.01\\
BLEURT	& \textbf{$-$0.34}&	$-$0.68\\
METEOR&	\textbf{0.33}	&0.15\\
MoverScore&\textbf{0.57}	&0.51\\
ROUGE-1.F1 &\textbf{0.40}	&0.22\\
ROUGE-2.F1&	\textbf{0.25}	&0.06\\
ROUGE-3.F1&\textbf{0.17}	&0.02\\
ROUGE-4.F1&\textbf{0.12}	&0.01\\
ROUGE-L.F1&\textbf{0.36}&	0.15\\
SacreBLEU&	\textbf{0.08}	&0.02 \\
\bottomrule
\end{tabular}
\caption{Values across evaluation metrics for scientific and non-scientific domains averaged over data subsets, models, and table formats.}
\label{table: scores_per_domain}
\end{table}

\begin{table*}[ht]  
\centering
\resizebox{\textwidth}{!}{
\begin{tabular}{lcccccc}
\toprule
\textbf{Metric}& \thead{\textbf{ComTQA} \\ \textbf{(FinTabNet)}} & \thead{\textbf{ComTQA} \\ \textbf{(PubTables-1M)}}	&\textbf{Logic2Text} &	\textbf{LogicNLG} &	\textbf{numericNLG}& \textbf{SciGen} \\
\midrule
BertScore.F1&	0.84   &0.83	&0.88&	\textbf{0.89}&	0.83	&0.82	\\
BLEU-1	  &     0.03	&0.04	&0.23	&\textbf{0.38}&	0.17	&0.13	\\
BLEU-2	  &     0.02&	0.02&	0.13	&\textbf{0.31}&	0.07	&0.04\\	
BLEU-3	  &     0.01&	0.02	&0.07&	\textbf{0.26}&	0.03	&0.02	\\
BLEU-4	  &     0.01&	0.01&	0.04&	\textbf{0.20}&	0.01	&0.01\\	
BLEURT	  &    $-$0.53&	$-$0.70&	\textbf{$-$0.13}&	$-$0.37&	$-$0.56&	$-$0.79	\\
METEOR	  &     0.07&	0.09&	0.36&	\textbf{0.56}&	0.20	&0.16	\\
MoverScore	&   0.50	&0.49	&0.60&	\textbf{0.61}&	0.53	&0.51	\\
ROUGE-1.F1	&   0.17&	0.14&	0.49&	\textbf{0.54}&	0.31&	0.23	\\
ROUGE-2.F1	&   0.10	&0.07&	0.24&	\textbf{0.42}&	0.07&	0.04\\
ROUGE-3.F1	&   0.03	&0.03&	0.13	&\textbf{0.34}&	0.02&	0.01	\\
ROUGE-4.F1	&   0.01&	0.02&	0.07&	\textbf{0.28}&	0.01&	0.00	\\
ROUGE-L.F1	 &  0.17	&0.14&	0.38&	\textbf{0.52}&	0.17	&0.13\\
SacreBLEU	 &  0.02&	0.02	&0.05&	\textbf{0.18}	&0.03&	0.02	\\
\bottomrule
\end{tabular}}
\caption{Values across evaluation metrics for each data subset averaged over models and table formats.}
\label{table: scores_per_subset}
\end{table*}

\begin{table*}[ht]
\centering
\resizebox{\textwidth}{!}{
\begin{tabular}{lcccccccccccccc}
\toprule
\textbf{Model}& \thead{\textbf{Bert-} \\ \textbf{Score.F1}} & \thead{\textbf{BLEU-} \\ \textbf{1}}	&\thead{\textbf{BLEU-} \\ \textbf{2}} &	\thead{\textbf{BLEU-} \\ \textbf{3}} &	\thead{\textbf{BLEU-} \\ \textbf{4}}& \textbf{BLEURT} & \textbf{METEOR} & \thead{\textbf{Mover-} \\ \textbf{Score}} & \thead{\textbf{ROUGE-} \\ \textbf{1.F1}} & \thead{\textbf{ROUGE-} \\ \textbf{2.F1}} & \thead{\textbf{ROUGE-} \\ \textbf{3.F1}} & \thead{\textbf{ROUGE-} \\ \textbf{4.F1}} & \thead{\textbf{ROUGE-} \\ \textbf{L.F1}} & \thead{\textbf{Sacre-} \\ \textbf{BLEU}}\\
\midrule
\multicolumn{15}{c}{\emph{Baseline}} \\
\midrule
Gemini-2.0-Flash\_mm &	\textbf{0.87}	& \textbf{0.22}&	\textbf{0.14}&	\textbf{0.11}&	\textbf{0.08}	& \textbf{$-$0.35}&	\textbf{0.27}&	\textbf{0.56}&	\textbf{0.40}&	\textbf{0.22}&	\textbf{0.14}&	\textbf{0.10}&	\textbf{0.33}	&\textbf{0.11}\\
Gemini-2.0-Flash\_llm & 0.86	& 0.21& \textbf{0.14}&	\textbf{0.11}	&\textbf{0.08}&	$-$0.39	&0.26&	\textbf{0.56}&	0.37&0.20&	\textbf{0.14}&	\textbf{0.10}&	0.31&	0.10 \\
\midrule
\multicolumn{15}{c}{\emph{MLLMs}} \\
\midrule
Idefics3-8B-Llama3	&\textbf{0.88}	&\textbf{0.19}	&\textbf{0.12}	&\textbf{0.09}	&\textbf{0.07}	&\textbf{$-$0.36}	&0.23	&\textbf{0.59}&	\textbf{0.47}	&\textbf{0.27}	&\textbf{0.13}	&\textbf{0.09}&	\textbf{0.42}&	\textbf{0.11}\\
Qwen2.5-VL-3B-Instruct	&0.85	&0.18	&\textbf{0.12}&	\textbf{0.09}	&\textbf{0.07}&	$-$0.51	&0.25&	0.55	&0.34&	0.18&	0.11	&0.08&	0.28	&0.07   \\
Qwen2.5-VL-7B-Instruct	&0.86&	\textbf{0.19}&	\textbf{0.12}& 0.08	&0.06&	$-$0.39&	\textbf{0.27}&	0.55	&0.36	&0.19&	0.12&	\textbf{0.09}	&0.30&	0.07\\
llama3-llava-next-8b-hf&	0.85&	0.16	&0.10&	0.06	&0.04	&$-$0.50	&0.24	&0.54	&0.31&	0.15&	0.09&	0.06&	0.25&	0.04\\
\midrule
\multicolumn{15}{c}{\emph{LLMs}} \\
\midrule
Mistral-Nemo-Instruct-2407&	\textbf{0.85}&	\textbf{0.17}&	\textbf{0.10}	&\textbf{0.07}&	\textbf{0.05}	&\textbf{$-$0.50}	&\textbf{0.25}&	\textbf{0.54}&	\textbf{0.33}	&\textbf{0.17}	&\textbf{0.11}&	\textbf{0.07}	&\textbf{0.27}&	\textbf{0.04}\\
Qwen2.5-3B-Instruct&	0.84&	0.15	&0.09&	0.06&	0.04	&$-$0.57	&0.24&	0.53	&0.28&	0.13	&0.08	&0.06&	0.22	&0.03\\
Qwen2.5-14B-Instruct&	0.84&	0.13&	0.07	&0.05	&0.03&	$-$0.56&	0.24&	0.53	&0.26&	0.12&	0.07&	0.05&	0.20	&0.03\\
Llama-3.2-3B-Instruct&	0.83&	0.12&	0.06	&0.04	&0.02&	$-$0.66	&0.20&	0.52&	0.23&	0.10	&0.05	&0.03&	0.18	&0.02\\
\bottomrule
\end{tabular}}
\caption{Values across evaluation metrics for individual models averaged over data subsets and table formats.}
\label{table: scores_per_model}
\end{table*}

\newpage
\onecolumn
\begin{figure*}[ht] 
    \center{\includegraphics[width=0.99\textwidth]{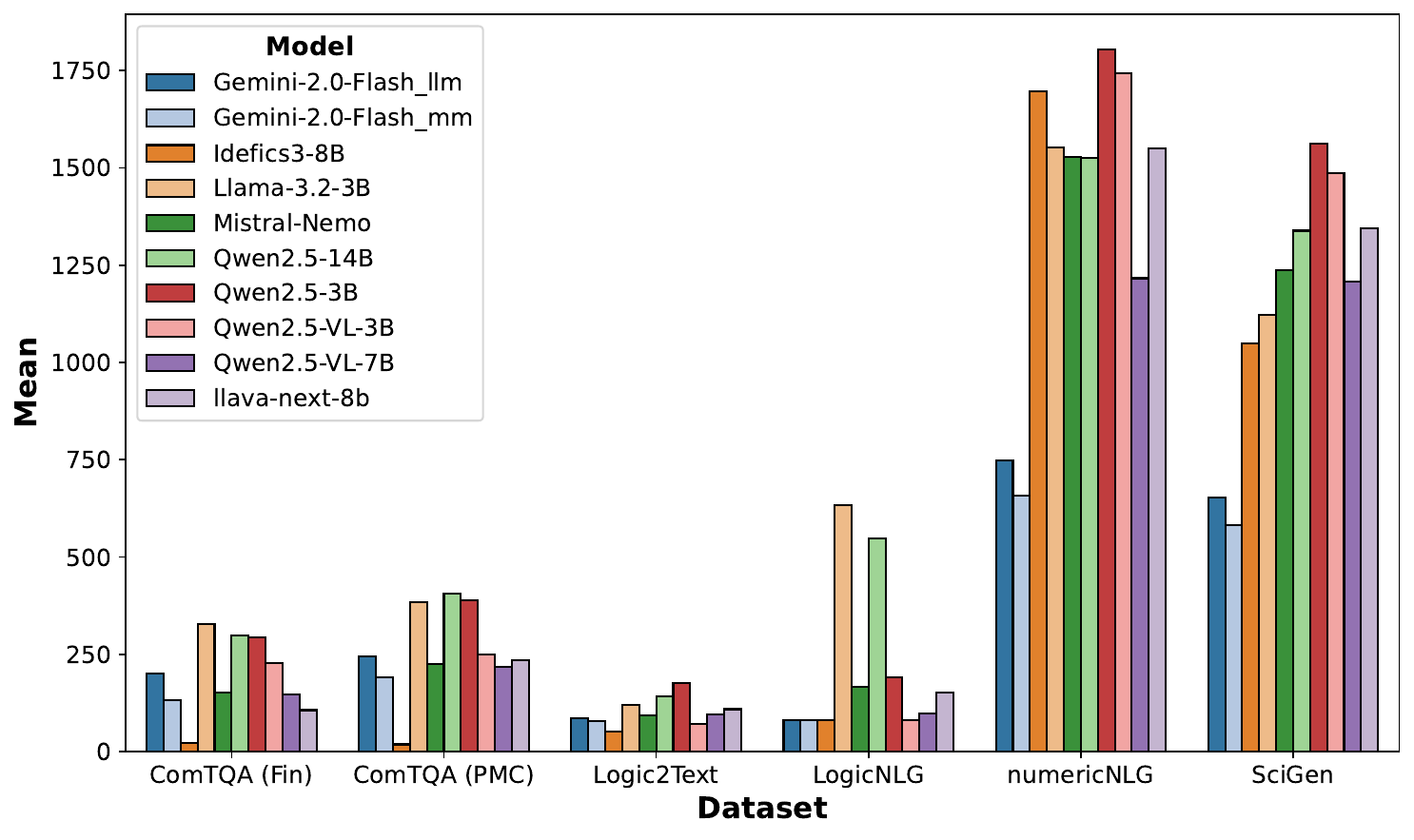}}
    \caption{Mean prediction lengths (in characters) for each model and data subset. Here ``\_llm'' and ``\_mm'' are used to distinguish between text and image input for Gemini-2.0-Flash, respectively.}
    \label{fig: pred_lenth_for_each_dataset}
\end{figure*}

\newpage
\clearpage

\section{Case Study}
\label{appendix: case_study}

\begin{figure*}[ht] 
    \center{\includegraphics[width=0.99\textwidth]{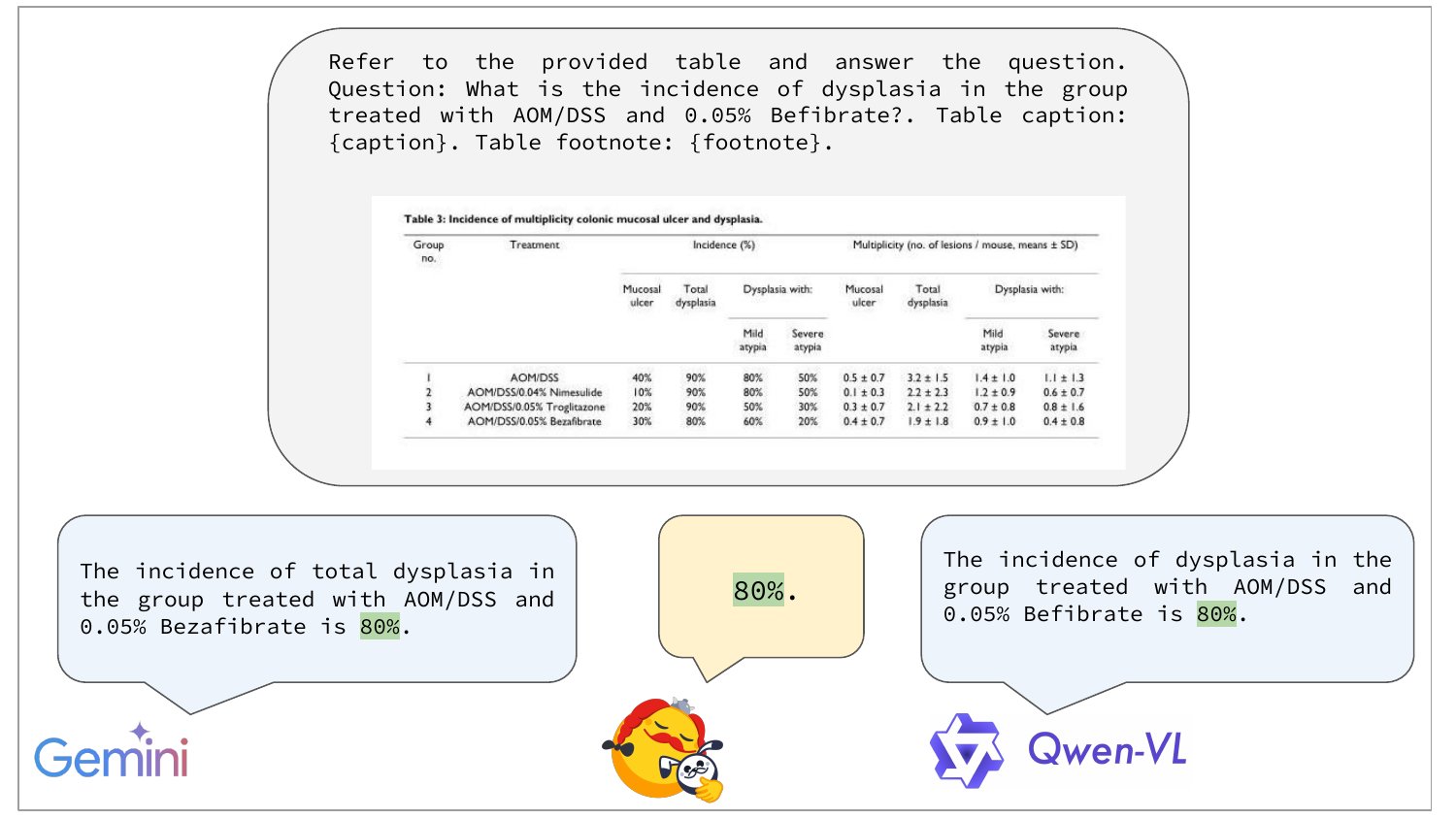}}
    \caption{An example illustrating differences in prediction length across Idefics3, Gemini-2.0-Flash, and Qwen2.0-VL (7B) models on a sample from the ComTQA (PubTables-1M) subset.}
    \label{fig: pred_lenght_ex}
\end{figure*}

\twocolumn
\newpage
\clearpage

\section{Additional interpretability analyses}
\label{appendix: interp}

\paragraph{Mistral-Nemo vs.\ Llama3.} The following figures show further examples of feature attribution and log-probability analysis comparing MistralNemo with Llama3.

\begin{figure*}[ht]
    \centering
    \begin{subfigure}[t]{0.48\textwidth}
        \centering
        \textbf{\small Mistral-Nemo-Instruct-2407} \\
        \includegraphics[width=\columnwidth]{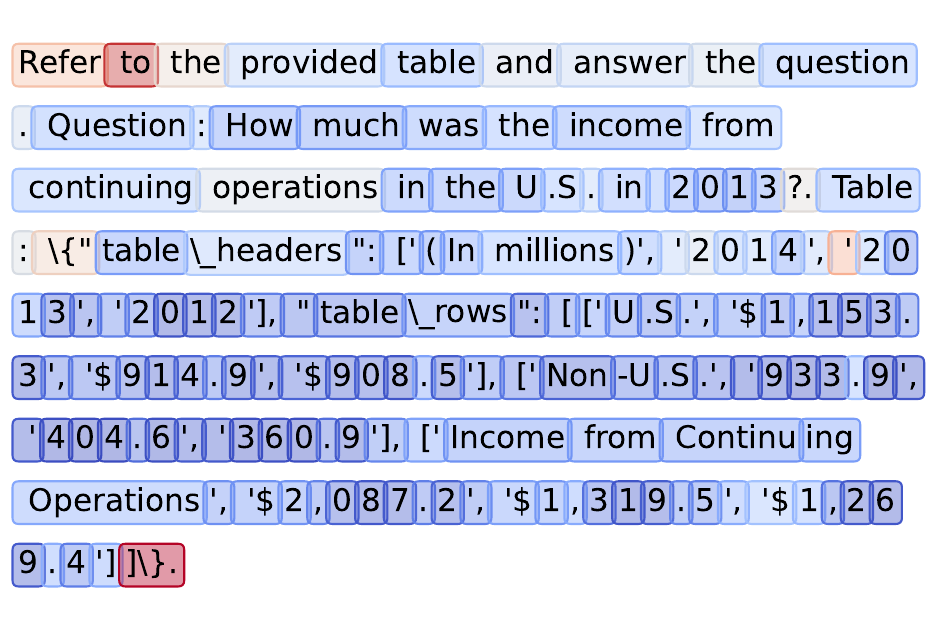}
    \end{subfigure}\hfill
    \begin{subfigure}[t]{0.48\textwidth}
        \centering
        \textbf{\small Llama-3.2-3B-Instruct} \\
        \includegraphics[width=\columnwidth]{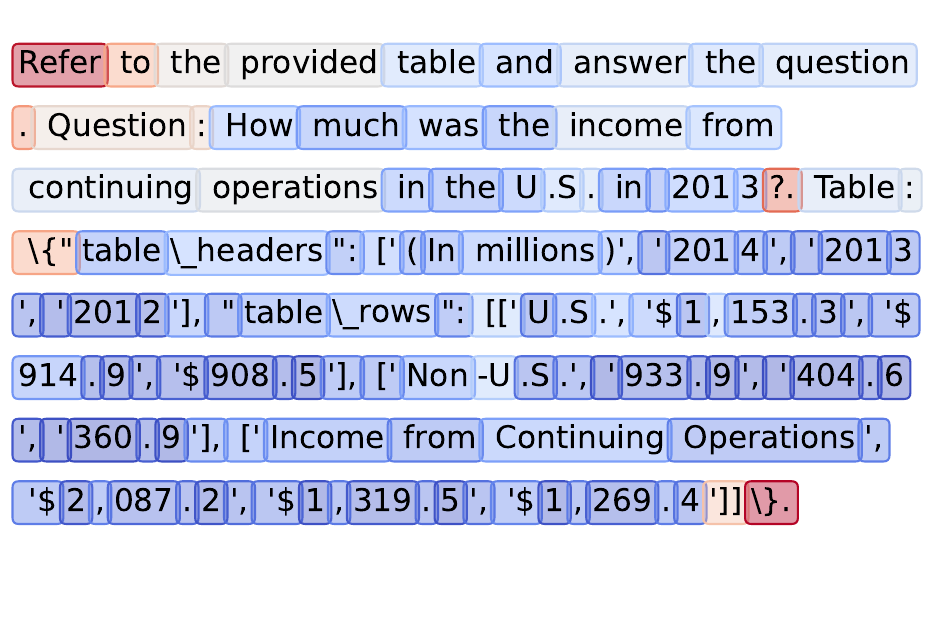}
    \end{subfigure}

    \vspace{0.125cm}
    \hrule

    \begin{subfigure}[t]{0.48\textwidth}
        \centering
        \includegraphics[width=\columnwidth]{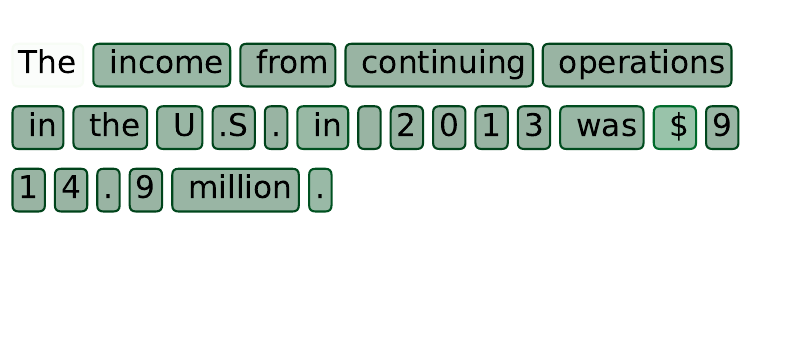}
    \end{subfigure}\hfill
    \begin{subfigure}[t]{0.48\textwidth}
        \centering
        \includegraphics[width=\columnwidth]{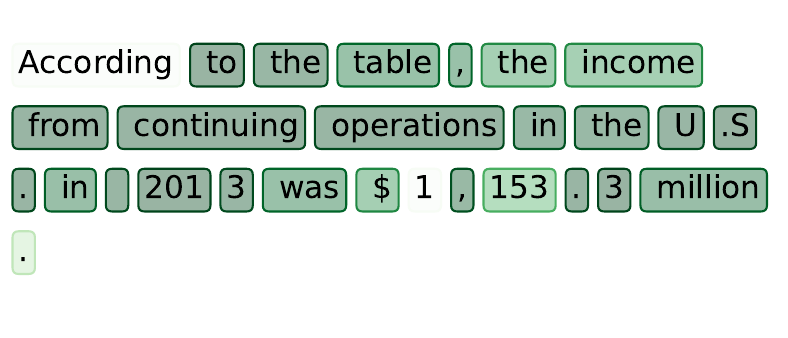}
    \end{subfigure}
    
    \caption{Interpretability analysis for the ComTQA (FinTabNet) instance with a table represented in a Dict format. The ground truth is \emph{``\$914.9 million''}. The visualisation follows the same procedure as Figure~\ref{fig:interp_results}.}
    \label{fig:interp_fin_2}
\end{figure*}

In Figure~\ref{fig:interp_fin_2} (ComTQA FinTabNet), Mistral-Nemo correctly predicts the answer, while Llama3 fails. We find a key difference in the attribution pattern around the columns \emph{``2014''} and \emph{``2013''}, where Mistral-Nemo assigns a slightly higher score (lighter blue) than Llama3. In the log-probability analysis, we see high uncertainty in Llama3 generating the final answer starting with \emph{``1''}. On the contrary, Mistral-Nemo shows a high level of confidence in the predicted value. 

\begin{figure*}[ht]
    \centering
    \begin{subfigure}[t]{0.48\textwidth}
        \centering
        \textbf{\small Mistral-Nemo-Instruct-2407} \\
        \includegraphics[width=\columnwidth]{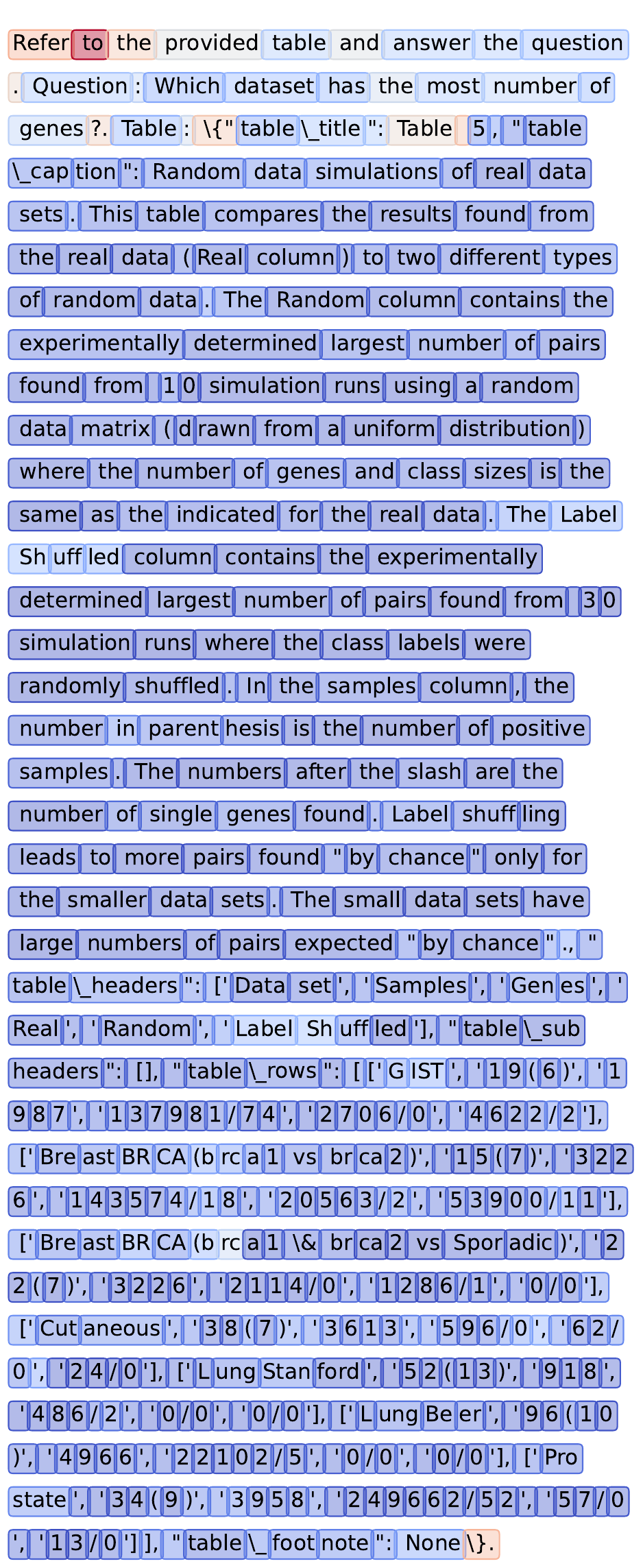}
    \end{subfigure}\hfill
    \begin{subfigure}[t]{0.48\textwidth}
        \centering
        \textbf{\small Llama-3.2-3B-Instruct} \\
        \includegraphics[width=\columnwidth]{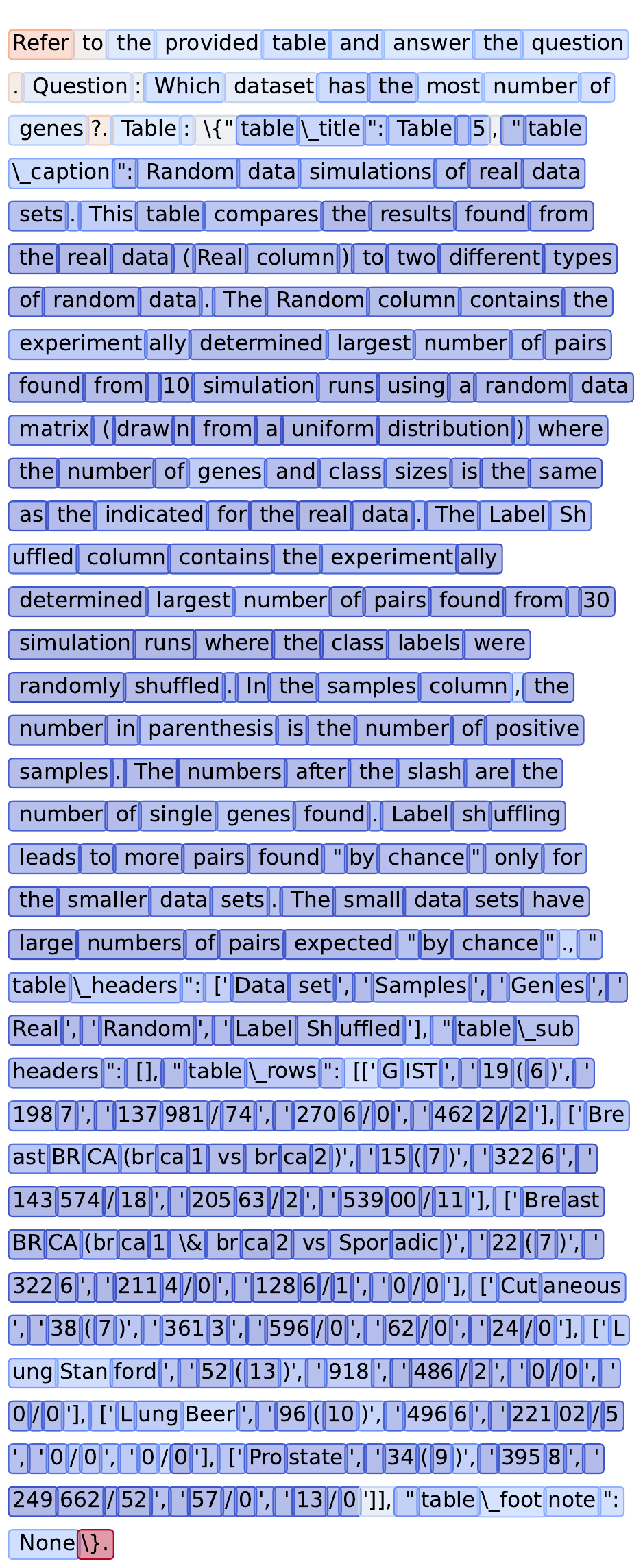}
    \end{subfigure}

    \vspace{0.125cm}
    \hrule

    \begin{subfigure}[t]{0.48\textwidth}
        \centering
        \includegraphics[width=\columnwidth]{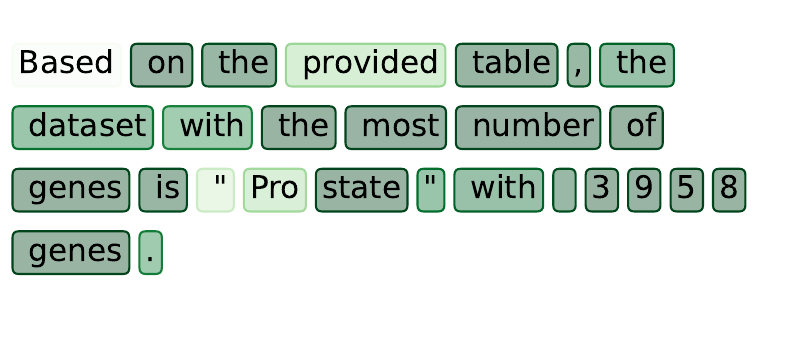}
    \end{subfigure}\hfill
    \begin{subfigure}[t]{0.48\textwidth}
        \centering
        \includegraphics[width=\columnwidth]{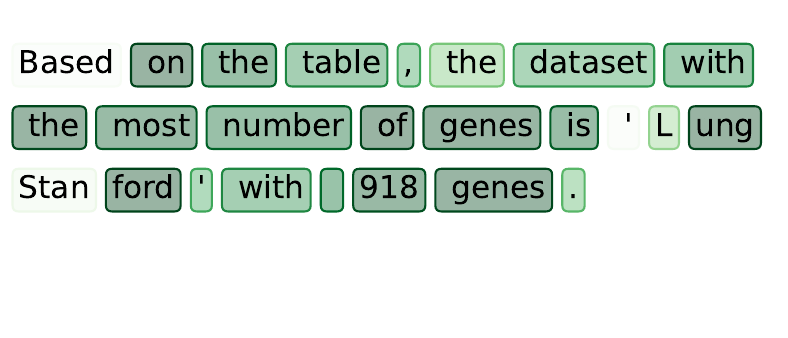}
    \end{subfigure}
    
    \caption{Interpretability analysis for the ComTQA (PubTables-1M) instance with a table represented in a Dict format. The ground truth is \emph{``LungBeer''}. The visualisation follows the same procedure as Figure~\ref{fig:interp_results}.}
    \label{fig:interp_pmc_1}
\end{figure*}

In Figure~\ref{fig:interp_pmc_1} (ComTQA PubTables-1M), both models generate incorrect answers. For Mistral-Nemo, one can barely see any attribution in the decisive row of the table. For Llama3, there is a slightly higher attribution for \emph{``Beer''} in \emph{``LungBeer''}. We also observe that the tokeniser splits the number into \emph{``496''} and \emph{``6''}. A plausible explanation for the failure is that when it processes \emph{``Lung Stanford''} with 918 genes, it likely finds it to be higher than 496 (ignoring the fourth digit \emph{``6''}). Regarding the log-probabilities, the decision of which feature to name after \emph{``the most number of genes is''} is controversial for both models, judged by the low confidence in the following token. 

\begin{figure*}[ht]
    \centering
    \begin{subfigure}[t]{0.48\textwidth}
        \centering
        \textbf{\small Mistral-Nemo-Instruct-2407} \\
        \includegraphics[width=\columnwidth]{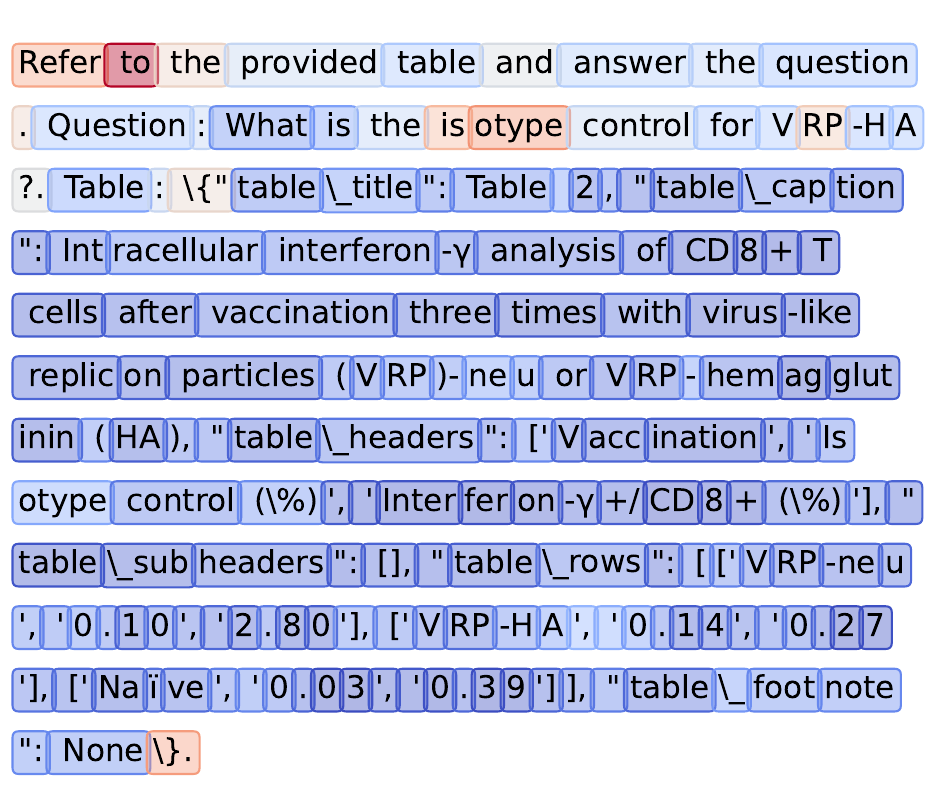}
    \end{subfigure}\hfill
    \begin{subfigure}[t]{0.48\textwidth}
        \centering
        \textbf{\small Llama-3.2-3B-Instruct} \\
        \includegraphics[width=\columnwidth]{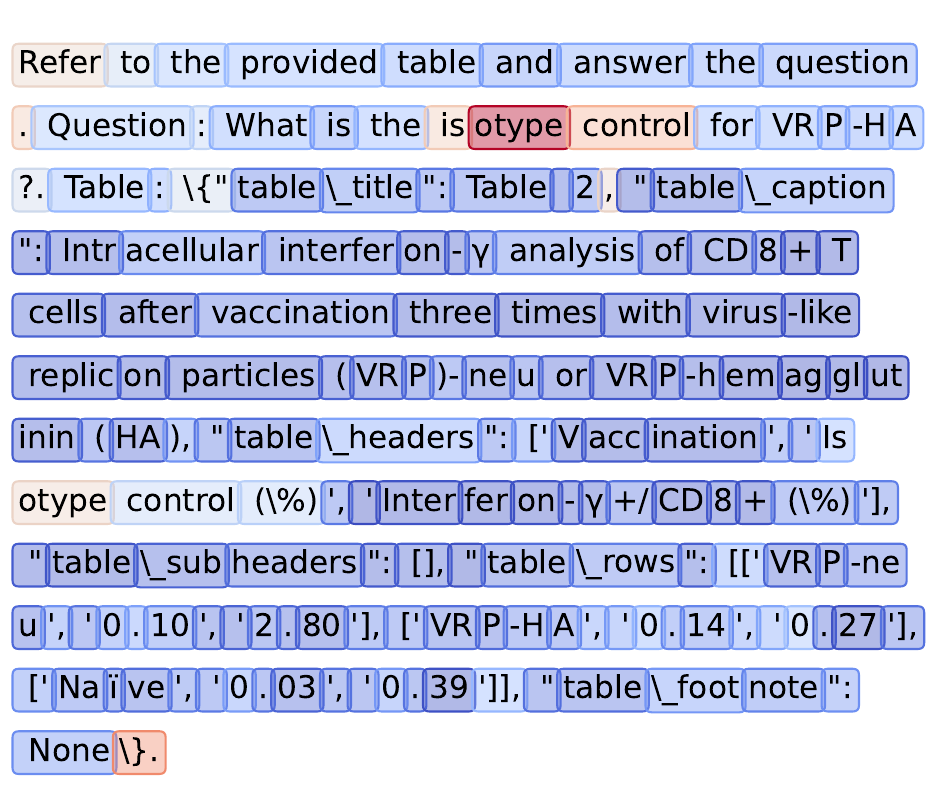}
    \end{subfigure}

    \vspace{0.125cm}
    \hrule

    \begin{subfigure}[t]{0.48\textwidth}
        \centering
        \includegraphics[width=\columnwidth]{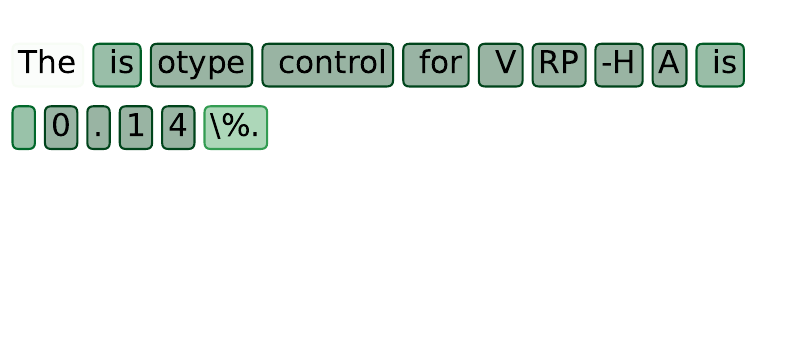}
    \end{subfigure}\hfill
    \begin{subfigure}[t]{0.48\textwidth}
        \centering
        \includegraphics[width=\columnwidth]{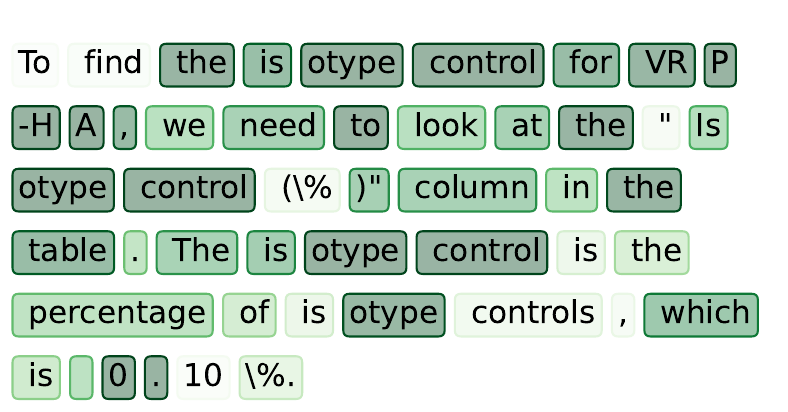}
    \end{subfigure}
    
    \caption{Interpretability analysis for  ComTQA (PubTables-1M) instance with the Dict format. The ground truth is \emph{``0.14\%''}. The visualisation follows the same procedure as Figure~\ref{fig:interp_results}.
    }
    \label{fig:interp_pmc_2}
\end{figure*}

In Figure~\ref{fig:interp_pmc_2} (ComTQA PubTables-1M), Mistral-Nemo solves the task correctly, whereas Llama3 fails to distinguish \emph{``VRP-HA''} from \emph{``VRP-neu''} and is not confident in the predicted value (\emph{10}). Mistral-Nemo focuses on the \emph{``VRP-HA''} row in the table more than the similar alternative \emph{``VRP-neu''} and generally finds the relevant feature name in the question to be more important, judging by the attribution patterns. When we compare this to the log-probabilities, the model is very confident about its decision (\emph{``VRP-HA''}) throughout the generation.

\paragraph{Dict vs.\ \LaTeX{} input format.} The following figures show examples of feature attribution and log-probability analysis. We compare predictions across Dict vs.\ \LaTeX{} representations of tables for Mistral-Nemo and Llama3 based on instances from the LogicNLG subset.

\begin{figure*}[ht]
    \centering
    \begin{subfigure}[t]{0.48\textwidth}
        \centering
        \textbf{\small Mistral-Nemo-Instruct-2407 (Dict)} \\
        \includegraphics[width=\columnwidth]{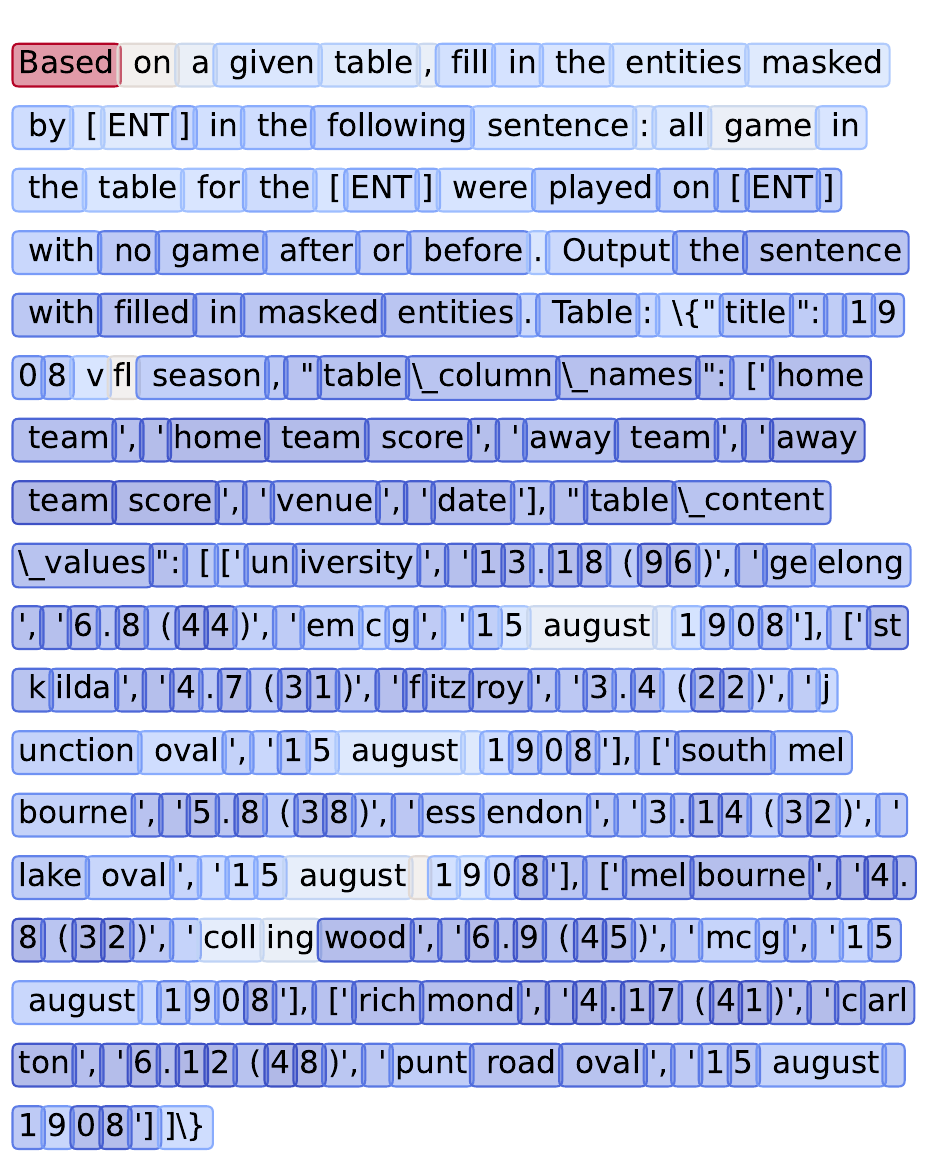}
    \end{subfigure}\hfill
    \begin{subfigure}[t]{0.48\textwidth}
        \centering
        \textbf{\small Mistral-Nemo-Instruct-2407 (\LaTeX{})} \\
        \includegraphics[width=\columnwidth]{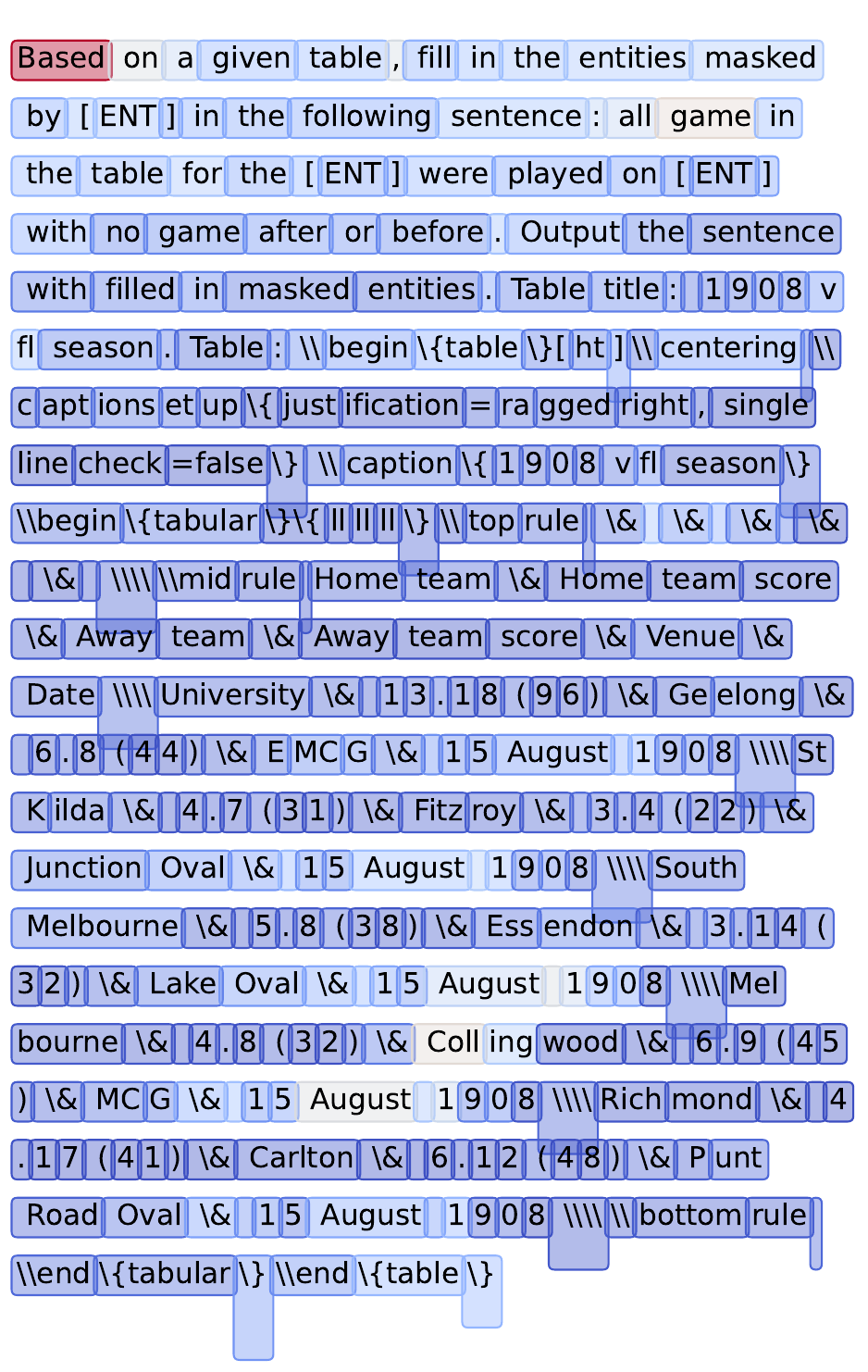}
    \end{subfigure}

    \vspace{0.125cm}
    \hrule

    \begin{subfigure}[t]{0.48\textwidth}
        \centering
        \includegraphics[width=\columnwidth]{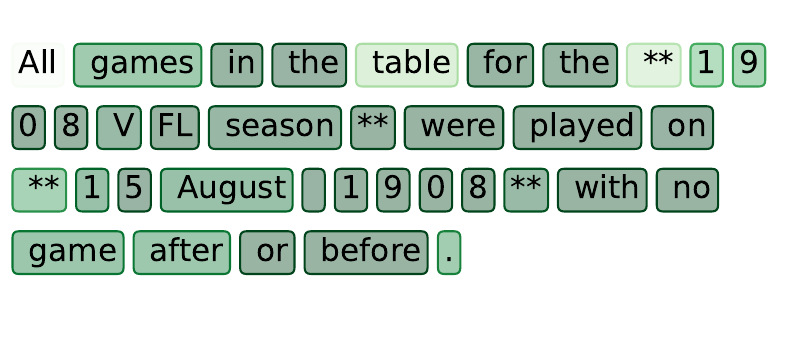}
    \end{subfigure}\hfill
    \begin{subfigure}[t]{0.48\textwidth}
        \centering
        \includegraphics[width=\columnwidth]{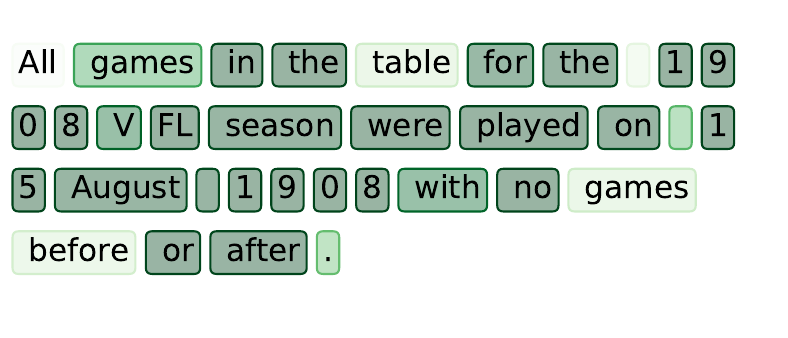}
    \end{subfigure}
    
    \caption{Interpretability analysis the LogicNLG instance comparing the Dict (left) with the \LaTeX{} (right) input format of the table. The ground truth is \emph{``all game in the table for the 1908 Vfl Season were played on 15 August 1908 with no game after or before''}. The visualisation follows the same procedure as Figure~\ref{fig:interp_results}.}
    \label{fig:interp_logicnlg_format}
\end{figure*}

In Figure~\ref{fig:interp_logicnlg_format}, Mistral-Nemo correctly predicts the missing entities with a high level of confidence. We notice high similarity between the input attribution patterns across two formats. In both cases, one of the most relevant tokens (month \emph{``August''}) is correctly identified to produce the right answer according to the ground truth and hence receives high attribution. The model focuses on the tokens relevant to the task and does not pay much attention to \LaTeX{} formatting tags, since the respective tokens generally remain barely considered throughout the generation. However, we can see some decreases in model confidence at the end of the generation (\emph{``games before''}).

\begin{figure*}[ht]
    \centering
    \begin{subfigure}[t]{0.48\textwidth}
        \centering
        \textbf{\small Llama-3.2-3B-Instruct (Dict)} \\
        \includegraphics[width=\columnwidth]{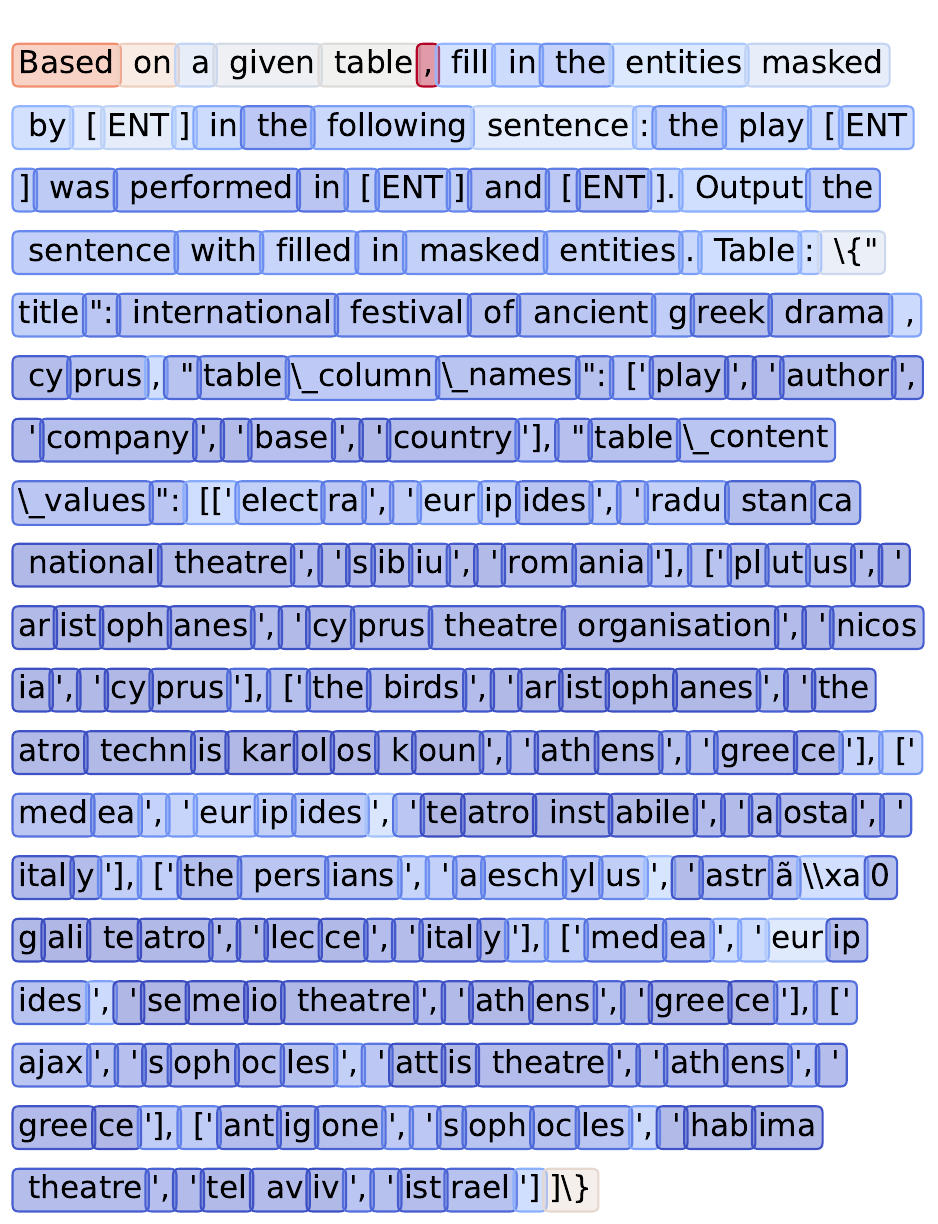}
    \end{subfigure}\hfill
    \begin{subfigure}[t]{0.48\textwidth}
        \centering
        \textbf{\small Llama-3.2-3B-Instruct (\LaTeX{})} \\
        \includegraphics[width=\columnwidth]{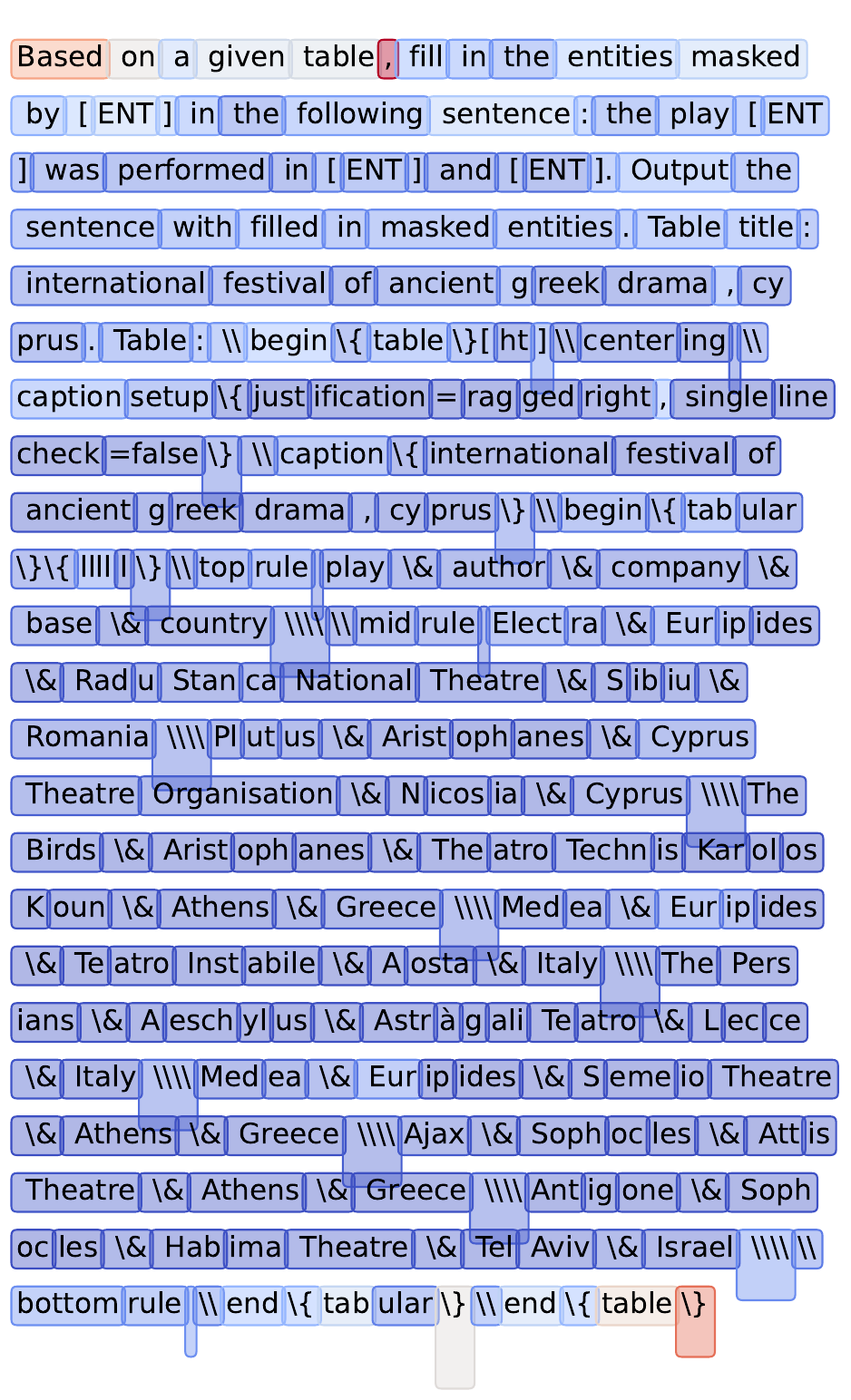}
    \end{subfigure}

    \vspace{0.125cm}
    \hrule

    \begin{subfigure}[t]{0.48\textwidth}
        \centering
        \includegraphics[width=\columnwidth]{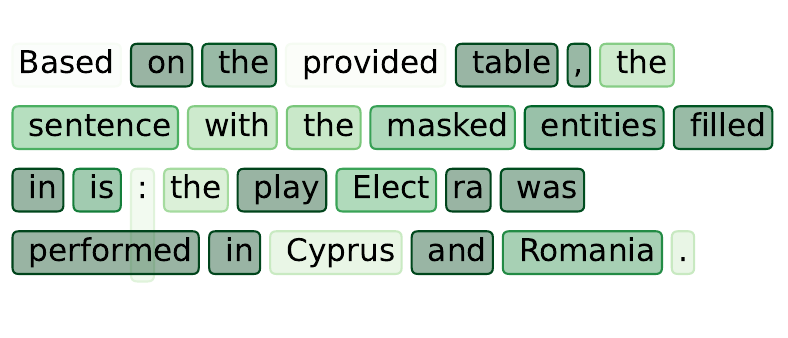}
    \end{subfigure}\hfill
    \begin{subfigure}[t]{0.48\textwidth}
        \centering
        \includegraphics[width=\columnwidth]{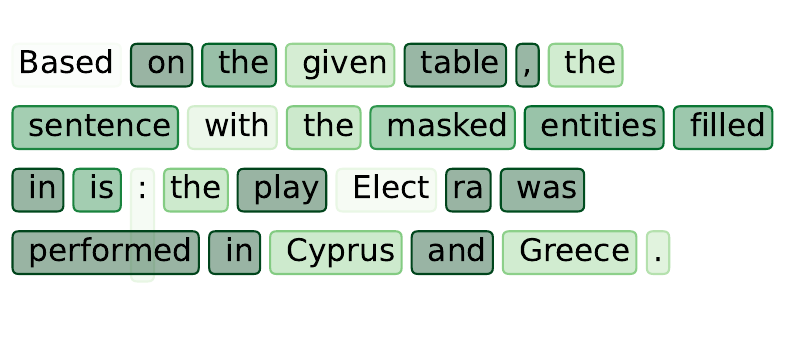}
    \end{subfigure}
    
    \caption{Interpretability analysis for the LogicNLG instance comparing the Dict (left) with the \LaTeX{} (right) input format of the table. The ground truth is \emph{``the play Medea was performed in Greece and Italy''}. The visualisation follows the same procedure as Figure~\ref{fig:interp_results}.}
    \label{fig:interp_logicnlg_format_llama}
\end{figure*}

In Figure~\ref{fig:interp_logicnlg_format_llama}, Llama3 generates the wrong responses in both cases. However, the Dict variant also makes the model focus on bracketing, separators, and punctuation quite often. Only for \LaTeX{}, there is a noticeably lower confidence about generating \emph{``Electra''} as the play of choice. For both representations of the table, however, Llama3 is not certain about the last two entities (\emph{``Cyprus and Romania''}, \emph{``Cyprus and Greece''}), which are either fully or partially incorrect according to the ground truth (\emph{``Greece and Italy''}).

\begin{figure*}[ht] 
    \center{\includegraphics[width=0.99\textwidth]{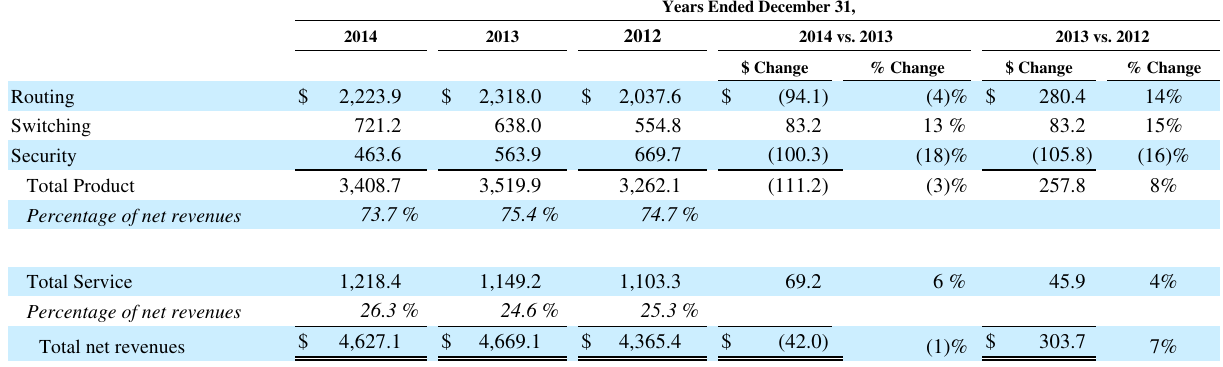}}
    \caption{Table image corresponding to the ComTQA (FinTabNet) example in Figure~\ref{fig:interp_results}. }
    \label{fig: interp_fin-id-11-image.png}
\end{figure*}


\begin{thebibliography}{75}
\providecommand{\natexlab}[1]{#1}

\bibitem[{Bai et~al.(2025)Bai, Chen, Liu, Wang, Ge, Song, Dang, Wang, Wang, Tang, Zhong, Zhu, Yang, Li, Wan, Wang, Ding, Fu, Xu, Ye, Zhang, Xie, Cheng, Zhang, Yang, Xu, and Lin}]{bai2025qwen25vltechnicalreport}
Shuai Bai, Keqin Chen, Xuejing Liu, Jialin Wang, Wenbin Ge, Sibo Song, Kai Dang, Peng Wang, Shijie Wang, Jun Tang, Humen Zhong, Yuanzhi Zhu, Mingkun Yang, Zhaohai Li, Jianqiang Wan, Pengfei Wang, Wei Ding, Zheren Fu, Yiheng Xu, Jiabo Ye, Xi~Zhang, Tianbao Xie, Zesen Cheng, Hang Zhang, Zhibo Yang, Haiyang Xu, and Junyang Lin. 2025.
\newblock \href {https://arxiv.org/abs/2502.13923} {Qwen2.5-{VL} technical report}.
\newblock \emph{Preprint}, arXiv:2502.13923.

\bibitem[{Banerjee and Lavie(2005)}]{banerjee-lavie-2005-meteor}
Satanjeev Banerjee and Alon Lavie. 2005.
\newblock \href {https://aclanthology.org/W05-0909/} {{METEOR}: An automatic metric for {MT} evaluation with improved correlation with human judgments}.
\newblock In \emph{Proceedings of the {ACL} Workshop on Intrinsic and Extrinsic Evaluation Measures for Machine Translation and/or Summarization}, pages 65--72, Ann Arbor, Michigan. Association for Computational Linguistics.

\bibitem[{Bhagavatula et~al.(2015)Bhagavatula, Noraset, and Downey}]{bhagavatulawikitables2015}
Chandra~Sekhar Bhagavatula, Thanapon Noraset, and Doug Downey. 2015.
\newblock Ta{bEL}: Entity linking in web tables.
\newblock In \emph{The Semantic Web - ISWC 2015}, pages 425--441, Cham. Springer International Publishing.

\bibitem[{Borisov et~al.(2022)Borisov, Leemann, Sessler, Haug, Pawelczyk, and Kasneci}]{Borisov_2022}
Vadim Borisov, Tobias Leemann, Kathrin Sessler, Johannes Haug, Martin Pawelczyk, and Gjergji Kasneci. 2022.
\newblock \href {https://doi.org/10.1109/tnnls.2022.3229161} {Deep neural networks and tabular data: A survey}.
\newblock \emph{{IEEE} Transactions on Neural Networks and Learning Systems}, pages 1--21.

\bibitem[{Bornmann et~al.(2021)Bornmann, Haunschild, and Mutz}]{bornmann2021growth}
Lutz Bornmann, Robin Haunschild, and R{\"u}diger Mutz. 2021.
\newblock Growth rates of modern science: A latent piecewise growth curve approach to model publication numbers from established and new literature databases.
\newblock \emph{Humanities and Social Sciences Communications}, 8(1):1--15.

\bibitem[{Caffagni et~al.(2024)Caffagni, Cocchi, Barsellotti, Moratelli, Sarto, Baraldi, Baraldi, Cornia, and Cucchiara}]{caffagni-etal-2024-revolution}
Davide Caffagni, Federico Cocchi, Luca Barsellotti, Nicholas Moratelli, Sara Sarto, Lorenzo Baraldi, Lorenzo Baraldi, Marcella Cornia, and Rita Cucchiara. 2024.
\newblock \href {https://doi.org/10.18653/v1/2024.findings-acl.807} {The revolution of multimodal large language models: A survey}.
\newblock In \emph{Findings of the Association for Computational Linguistics: ACL 2024}, pages 13590--13618, Bangkok, Thailand. Association for Computational Linguistics.

\bibitem[{Chang et~al.(2024)Chang, Wang, Wang, Wu, Yang, Zhu, Chen, Yi, Wang, Wang, Ye, Zhang, Chang, Yu, Yang, and Xie}]{chang2024evaluationofllms}
Yupeng Chang, Xu~Wang, Jindong Wang, Yuan Wu, Linyi Yang, Kaijie Zhu, Hao Chen, Xiaoyuan Yi, Cunxiang Wang, Yidong Wang, Wei Ye, Yue Zhang, Yi~Chang, Philip~S. Yu, Qiang Yang, and Xing Xie. 2024.
\newblock \href {https://doi.org/10.1145/3641289} {A survey on evaluation of large language models}.
\newblock \emph{ACM Trans. Intell. Syst. Technol.}, 15(3).

\bibitem[{Chen(2023)}]{chen-2023-large}
Wenhu Chen. 2023.
\newblock \href {https://doi.org/10.18653/v1/2023.findings-eacl.83} {Large language models are few(1)-shot table reasoners}.
\newblock In \emph{Findings of the Association for Computational Linguistics: EACL 2023}, pages 1120--1130, Dubrovnik, Croatia. Association for Computational Linguistics.

\bibitem[{Chen et~al.(2021)Chen, Chang, Schlinger, Wang, and Cohen}]{chen2021openquestionansweringtables}
Wenhu Chen, Ming-Wei Chang, Eva Schlinger, William Wang, and William~W. Cohen. 2021.
\newblock \href {https://arxiv.org/abs/2010.10439} {Open question answering over tables and text}.
\newblock \emph{Preprint}, arXiv:2010.10439.

\bibitem[{Chen et~al.(2020{\natexlab{a}})Chen, Chen, Su, Chen, and Wang}]{chen-etal-2020-logical}
Wenhu Chen, Jianshu Chen, Yu~Su, Zhiyu Chen, and William~Yang Wang. 2020{\natexlab{a}}.
\newblock \href {https://doi.org/10.18653/v1/2020.acl-main.708} {Logical natural language generation from open-domain tables}.
\newblock In \emph{Proceedings of the 58th Annual Meeting of the Association for Computational Linguistics}, pages 7929--7942, Online. Association for Computational Linguistics.

\bibitem[{Chen et~al.(2020{\natexlab{b}})Chen, Wang, Chen, Zhang, Wang, Li, Zhou, and Wang}]{chentabfact2020}
Wenhu Chen, Hongmin Wang, Jianshu Chen, Yunkai Zhang, Hong Wang, Shiyang Li, Xiyou Zhou, and William~Yang Wang. 2020{\natexlab{b}}.
\newblock Tab{F}act: A large-scale dataset for table-based fact verification.
\newblock In \emph{International Conference on Learning Representations (ICLR)}, Addis Ababa, Ethiopia.

\bibitem[{Chen et~al.(2020{\natexlab{c}})Chen, Chen, Zha, Zhou, Zhang, Sundaresan, and Wang}]{chen-etal-2020-logic2text}
Zhiyu Chen, Wenhu Chen, Hanwen Zha, Xiyou Zhou, Yunkai Zhang, Sairam Sundaresan, and William~Yang Wang. 2020{\natexlab{c}}.
\newblock \href {https://doi.org/10.18653/v1/2020.findings-emnlp.190} {{L}ogic2{T}ext: High-fidelity natural language generation from logical forms}.
\newblock In \emph{Findings of the Association for Computational Linguistics: EMNLP 2020}, pages 2096--2111, Online. Association for Computational Linguistics.

\bibitem[{Cheng et~al.(2022)Cheng, Dong, Wang, Jia, Guo, Gao, Han, Lou, and Zhang}]{cheng-etal-2022}
Zhoujun Cheng, Haoyu Dong, Zhiruo Wang, Ran Jia, Jiaqi Guo, Yan Gao, Shi Han, Jian-Guang Lou, and Dongmei Zhang. 2022.
\newblock \href {https://doi.org/10.18653/v1/2022.acl-long.78} {{H}i{T}ab: A hierarchical table dataset for question answering and natural language generation}.
\newblock In \emph{Proceedings of the 60th Annual Meeting of the Association for Computational Linguistics (Volume 1: Long Papers)}, pages 1094--1110, Dublin, Ireland. Association for Computational Linguistics.

\bibitem[{Clark and Divvala(2016)}]{PDFFigures}
Christopher Clark and Santosh Divvala. 2016.
\newblock \href {https://doi.org/10.1145/2910896.2910904} {{PDFF}igures 2.0: Mining figures from research papers}.
\newblock In \emph{Proceedings of the 16th ACM/IEEE-CS on Joint Conference on Digital Libraries}, JCDL '16, page 143–152, New York, NY, USA. Association for Computing Machinery.

\bibitem[{Deng et~al.(2024)Deng, Sun, He, Sikka, Chen, Ma, Zhang, and Mihalcea}]{deng-2024-tables-as-texts-or-images}
Naihao Deng, Zhenjie Sun, Ruiqi He, Aman Sikka, Yulong Chen, Lin Ma, Yue Zhang, and Rada Mihalcea. 2024.
\newblock \href {https://doi.org/10.18653/v1/2024.findings-acl.23} {Tables as texts or images: Evaluating the table reasoning ability of {LLM}s and {MLLM}s}.
\newblock In \emph{Findings of the Association for Computational Linguistics: ACL 2024}, pages 407--426, Bangkok, Thailand. Association for Computational Linguistics.

\bibitem[{Enouen et~al.(2024)Enouen, Nakhost, Ebrahimi, Arik, Liu, and Pfister}]{enouen-2024-textgenshap}
James Enouen, Hootan Nakhost, Sayna Ebrahimi, Sercan Arik, Yan Liu, and Tomas Pfister. 2024.
\newblock \href {https://doi.org/10.18653/v1/2024.findings-acl.832} {{T}ext{G}en{SHAP}: Scalable post-hoc explanations in text generation with long documents}.
\newblock In \emph{Findings of the Association for Computational Linguistics: ACL 2024}, pages 13984--14011, Bangkok, Thailand. Association for Computational Linguistics.

\bibitem[{Fang et~al.(2024)Fang, Xu, Tan, Zhang, Hu, Qi, Nickleach, Socolinsky, Sengamedu, and Faloutsos}]{fang2024largelanguagemodelsllmstabular}
Xi~Fang, Weijie Xu, Fiona~Anting Tan, Jiani Zhang, Ziqing Hu, Yanjun Qi, Scott Nickleach, Diego Socolinsky, Srinivasan Sengamedu, and Christos Faloutsos. 2024.
\newblock \href {https://arxiv.org/abs/2402.17944} {Large language models ({LLMs}) on tabular data: Prediction, generation, and understanding -- a survey}.
\newblock \emph{Preprint}, arXiv:2402.17944.

\bibitem[{Ferrando et~al.(2024)Ferrando, Sarti, Bisazza, and Costa-jussà}]{ferrando-2024-primer-inner-workings}
Javier Ferrando, Gabriele Sarti, Arianna Bisazza, and Marta~R. Costa-jussà. 2024.
\newblock \href {https://arxiv.org/abs/2405.00208} {A primer on the inner workings of transformer-based language models}.
\newblock \emph{arXiv}, abs/2405.00208.

\bibitem[{Fortunato et~al.(2018)Fortunato, Bergstrom, B{\"o}rner, Evans, Helbing, Milojevi{\'c}, Petersen, Radicchi, Sinatra, Uzzi et~al.}]{fortunato2018science}
Santo Fortunato, Carl~T Bergstrom, Katy B{\"o}rner, James~A Evans, Dirk Helbing, Sta{\v{s}}a Milojevi{\'c}, Alexander~M Petersen, Filippo Radicchi, Roberta Sinatra, Brian Uzzi, et~al. 2018.
\newblock Science of science.
\newblock \emph{Science}, 359(6379):eaao0185.

\bibitem[{Gehrmann et~al.(2023)Gehrmann, Clark, and Sellam}]{gehrmann2023}
Sebastian Gehrmann, Elizabeth Clark, and Thibault Sellam. 2023.
\newblock \href {https://doi.org/10.1613/jair.1.13715} {Repairing the cracked foundation: A survey of obstacles in evaluation practices for generated text}.
\newblock \emph{J. Artif. Int. Res.}, 77.

\bibitem[{Gong et~al.(2020)Gong, Sun, Feng, Qin, Bi, Liu, and Liu}]{gong-etal-2020-tablegpt}
Heng Gong, Yawei Sun, Xiaocheng Feng, Bing Qin, Wei Bi, Xiaojiang Liu, and Ting Liu. 2020.
\newblock \href {https://doi.org/10.18653/v1/2020.coling-main.179} {{T}able{GPT}: Few-shot table-to-text generation with table structure reconstruction and content matching}.
\newblock In \emph{Proceedings of the 28th International Conference on Computational Linguistics}, pages 1978--1988, Barcelona, Spain (Online). International Committee on Computational Linguistics.

\bibitem[{Gorishniy et~al.(2021)Gorishniy, Rubachev, Khrulkov, and Babenko}]{gorishniy2021}
Yury Gorishniy, Ivan Rubachev, Valentin Khrulkov, and Artem Babenko. 2021.
\newblock Revisiting deep learning models for tabular data.
\newblock In \emph{Advances in Neural Information Processing Systems}, volume~34, pages 18932--18943. Curran Associates, Inc.

\bibitem[{Grattafiori et~al.(2024)Grattafiori, Dubey, and et. al}]{grattafiori2024llama3herdmodels}
Aaron Grattafiori, Abhimanyu Dubey, and Abhinav~Jauhri et. al. 2024.
\newblock \href {https://arxiv.org/abs/2407.21783} {The {L}lama 3 herd of models}.
\newblock \emph{Preprint}, arXiv:2407.21783.

\bibitem[{Herzig et~al.(2020)Herzig, Nowak, M{\"u}ller, Piccinno, and Eisenschlos}]{herzig-etal-2020-tapas}
Jonathan Herzig, Pawel~Krzysztof Nowak, Thomas M{\"u}ller, Francesco Piccinno, and Julian Eisenschlos. 2020.
\newblock \href {https://doi.org/10.18653/v1/2020.acl-main.398} {{T}a{P}as: Weakly supervised table parsing via pre-training}.
\newblock In \emph{Proceedings of the 58th Annual Meeting of the Association for Computational Linguistics}, pages 4320--4333, Online. Association for Computational Linguistics.

\bibitem[{Ho et~al.(2024)Ho, Nguyen, Dao, Jiang, Chida, Sugimoto, To, Boudin, and Aizawa}]{ho2024surveypretrainedlanguagemodels}
Xanh Ho, Anh Khoa~Duong Nguyen, An~Tuan Dao, Junfeng Jiang, Yuki Chida, Kaito Sugimoto, Huy~Quoc To, Florian Boudin, and Akiko Aizawa. 2024.
\newblock \href {https://arxiv.org/abs/2401.17824} {A survey of pre-trained language models for processing scientific text}.
\newblock \emph{Preprint}, arXiv:2401.17824.

\bibitem[{Hong et~al.(2021)Hong, Ward, Chard, Blaiszik, and Foster}]{Hong2021}
Zhi Hong, Logan Ward, Kyle Chard, Ben Blaiszik, and Ian Foster. 2021.
\newblock \href {https://doi.org/10.1007/s11837-021-04902-9} {Challenges and advances in information extraction from scientific literature: a review}.
\newblock \emph{JOM}, 73:1543--1851.

\bibitem[{Iida et~al.(2021)Iida, Thai, Manjunatha, and Iyyer}]{iida-etal-2021-tabbie}
Hiroshi Iida, Dung Thai, Varun Manjunatha, and Mohit Iyyer. 2021.
\newblock \href {https://doi.org/10.18653/v1/2021.naacl-main.270} {{TABBIE}: Pretrained representations of tabular data}.
\newblock In \emph{Proceedings of the 2021 Conference of the North American Chapter of the Association for Computational Linguistics: Human Language Technologies}, pages 3446--3456, Online. Association for Computational Linguistics.

\bibitem[{Laurençon et~al.(2024)Laurençon, Marafioti, Sanh, and Tronchon}]{laurençon2024buildingbetterunderstandingvisionlanguage}
Hugo Laurençon, Andrés Marafioti, Victor Sanh, and Léo Tronchon. 2024.
\newblock \href {https://arxiv.org/abs/2408.12637} {Building and better understanding vision-language models: insights and future directions}.
\newblock \emph{Preprint}, arXiv:2408.12637.

\bibitem[{Li et~al.(2024)Li, Zhang, Zhang, Guo, Zhang, Li, Zhang, Liu, and Li}]{li2024llavanext-strong}
Bo~Li, Kaichen Zhang, Hao Zhang, Dong Guo, Renrui Zhang, Feng Li, Yuanhan Zhang, Ziwei Liu, and Chunyuan Li. 2024.
\newblock \href {https://llava-vl.github.io/blog/2024-05-10-llava-next-stronger-llms/} {{LLaVA-NeXT}: Stronger {LLM}s supercharge multimodal capabilities in the wild}.

\bibitem[{Lin(2004)}]{lin-2004-rouge}
Chin-Yew Lin. 2004.
\newblock \href {https://aclanthology.org/W04-1013/} {{ROUGE}: A package for automatic evaluation of summaries}.
\newblock In \emph{Text Summarization Branches Out}, pages 74--81, Barcelona, Spain. Association for Computational Linguistics.

\bibitem[{Lu et~al.(2024)Lu, Pan, Ma, Nakov, and Kan}]{lu-2024-tart}
Xinyuan Lu, Liangming Pan, Yubo Ma, Preslav Nakov, and Min-Yen Kan. 2024.
\newblock \href {https://openreview.net/forum?id=UF8RE1rkfU} {{TART}: An open-source tool-augmented framework for explainable table-based reasoning}.
\newblock In \emph{NeurIPS 2024 Third Table Representation Learning Workshop}.

\bibitem[{Marzocchi et~al.(2022)Marzocchi, Cremaschi, Pozzi, Avogadro, and Palmonari}]{mammotab2022}
Mattia Marzocchi, Marco Cremaschi, Riccardo Pozzi, Roberto Avogadro, and Matteo Palmonari. 2022.
\newblock \href {https://ceur-ws.org/Vol-3320/paper3.pdf} {Mammo{T}ab: A giant and comprehensive dataset for semantic table interpretation}.
\newblock In \emph{Proceedings of the Semantic Web Challenge on Tabular Data to Knowledge Graph Matching (SemTab2022)}.

\bibitem[{Moosavi et~al.(2021)Moosavi, R{\"u}ckl{\'e}, Roth, and Gurevych}]{moosavi2021learningreasontextgeneration}
Nafise~Sadat Moosavi, Andreas R{\"u}ckl{\'e}, Dan Roth, and Iryna Gurevych. 2021.
\newblock \href {https://openreview.net/forum?id=Jul-uX7EV_I} {Sci{G}en: a dataset for reasoning-aware text generation from scientific tables}.
\newblock In \emph{Thirty-fifth Conference on Neural Information Processing Systems Datasets and Benchmarks Track (Round 2)}.

\bibitem[{Nan et~al.(2022)Nan, Hsieh, Mao, Lin, Verma, Zhang, Kry{\'s}ci{\'n}ski, Schoelkopf, Kong, Tang, Mutuma, Rosand, Trindade, Bandaru, Cunningham, Xiong, Radev, and Radev}]{nan-etal-2022}
Linyong Nan, Chiachun Hsieh, Ziming Mao, Xi~Victoria Lin, Neha Verma, Rui Zhang, Wojciech Kry{\'s}ci{\'n}ski, Hailey Schoelkopf, Riley Kong, Xiangru Tang, Mutethia Mutuma, Ben Rosand, Isabel Trindade, Renusree Bandaru, Jacob Cunningham, Caiming Xiong, Dragomir Radev, and Dragomir Radev. 2022.
\newblock \href {https://doi.org/10.1162/tacl_a_00446} {{F}e{T}a{QA}: Free-form table question answering}.
\newblock \emph{Transactions of the Association for Computational Linguistics}, 10:35--49.

\bibitem[{Nguyen et~al.(2025)Nguyen, Brugere, Sharma, Kariyappa, Nguyen, and Lecue}]{nguyen-2025-interpretable-llm-based-table-qa}
Giang Nguyen, Ivan Brugere, Shubham Sharma, Sanjay Kariyappa, Anh~Totti Nguyen, and Freddy Lecue. 2025.
\newblock \href {https://arxiv.org/abs/2412.12386} {Interpretable {LLM}-based table question answering}.
\newblock \emph{arXiv}, abs/2412.12386.

\bibitem[{OpenAI et~al.(2024)OpenAI, Achiam, and et. al}]{openai2024gpt4technicalreport}
OpenAI, Josh Achiam, and Steven~Adler et. al. 2024.
\newblock \href {https://arxiv.org/abs/2303.08774} {{GPT}-4 technical report}.
\newblock \emph{Preprint}, arXiv:2303.08774.

\bibitem[{Os{\'e}s~Grijalba et~al.(2024)Os{\'e}s~Grijalba, Ure{\~n}a-L{\'o}pez, Mart{\'i}nez~C{\'a}mara, and Camacho-Collados}]{oses-grijalba-etal-2024-question}
Jorge Os{\'e}s~Grijalba, L.~Alfonso Ure{\~n}a-L{\'o}pez, Eugenio Mart{\'i}nez~C{\'a}mara, and Jose Camacho-Collados. 2024.
\newblock \href {https://aclanthology.org/2024.lrec-main.1179/} {Question answering over tabular data with {D}ata{B}ench: A large-scale empirical evaluation of {LLM}s}.
\newblock In \emph{Proceedings of the 2024 Joint International Conference on Computational Linguistics, Language Resources and Evaluation (LREC-COLING 2024)}, pages 13471--13488, Torino, Italia. ELRA and ICCL.

\bibitem[{Pang et~al.(2024)Pang, Cao, Yang, and Luo}]{pang-2024-tabis}
Chaoxu Pang, Yixuan Cao, Chunhao Yang, and Ping Luo. 2024.
\newblock \href {https://doi.org/10.18653/v1/2024.findings-acl.82} {Uncovering limitations of large language models in information seeking from tables}.
\newblock In \emph{Findings of the Association for Computational Linguistics: ACL 2024}, pages 1388--1409, Bangkok, Thailand. Association for Computational Linguistics.

\bibitem[{Papineni et~al.(2002)Papineni, Roukos, Ward, and Zhu}]{papineni-etal-2002-bleu}
Kishore Papineni, Salim Roukos, Todd Ward, and Wei-Jing Zhu. 2002.
\newblock \href {https://doi.org/10.3115/1073083.1073135} {{B}leu: a method for automatic evaluation of machine translation}.
\newblock In \emph{Proceedings of the 40th Annual Meeting of the Association for Computational Linguistics}, pages 311--318, Philadelphia, Pennsylvania, USA. Association for Computational Linguistics.

\bibitem[{Parcalabescu and Frank(2025)}]{parcalabescu-frank-2025-cc-shap-vlm}
Letitia Parcalabescu and Anette Frank. 2025.
\newblock \href {https://openreview.net/forum?id=lCasyP21Bf} {Do vision \& language decoders use images and text equally? {H}ow self-consistent are their explanations?}
\newblock In \emph{The Thirteenth International Conference on Learning Representations}.

\bibitem[{Parikh et~al.(2020)Parikh, Wang, Gehrmann, Faruqui, Dhingra, Yang, and Das}]{parikh-etal-2020-totto}
Ankur Parikh, Xuezhi Wang, Sebastian Gehrmann, Manaal Faruqui, Bhuwan Dhingra, Diyi Yang, and Dipanjan Das. 2020.
\newblock \href {https://doi.org/10.18653/v1/2020.emnlp-main.89} {{ToTTo}: A controlled table-to-text generation dataset}.
\newblock In \emph{Proceedings of the 2020 Conference on Empirical Methods in Natural Language Processing (EMNLP)}, pages 1173--1186, Online. Association for Computational Linguistics.

\bibitem[{Post(2018)}]{post-2018-call}
Matt Post. 2018.
\newblock \href {https://doi.org/10.18653/v1/W18-6319} {A call for clarity in reporting {BLEU} scores}.
\newblock In \emph{Proceedings of the Third Conference on Machine Translation: Research Papers}, pages 186--191, Brussels, Belgium. Association for Computational Linguistics.

\bibitem[{Qi et~al.(2024)Qi, Sarti, Fern{\'a}ndez, and Bisazza}]{qi-2024-mirage}
Jirui Qi, Gabriele Sarti, Raquel Fern{\'a}ndez, and Arianna Bisazza. 2024.
\newblock \href {https://doi.org/10.18653/v1/2024.emnlp-main.347} {Model internals-based answer attribution for trustworthy retrieval-augmented generation}.
\newblock In \emph{Proceedings of the 2024 Conference on Empirical Methods in Natural Language Processing}, pages 6037--6053, Miami, Florida, USA. Association for Computational Linguistics.

\bibitem[{Qwen et~al.(2025)Qwen, :, Yang, Yang, Zhang, Hui, Zheng, Yu, Li, Liu, Huang, Wei, Lin, Yang, Tu, Zhang, Yang, Yang, Zhou, Lin, Dang, Lu, Bao, Yang, Yu, Li, Xue, Zhang, Zhu, Men, Lin, Li, Tang, Xia, Ren, Ren, Fan, Su, Zhang, Wan, Liu, Cui, Zhang, and Qiu}]{qwen2025qwen25technicalreport}
Qwen, :, An~Yang, Baosong Yang, Beichen Zhang, Binyuan Hui, Bo~Zheng, Bowen Yu, Chengyuan Li, Dayiheng Liu, Fei Huang, Haoran Wei, Huan Lin, Jian Yang, Jianhong Tu, Jianwei Zhang, Jianxin Yang, Jiaxi Yang, Jingren Zhou, Junyang Lin, Kai Dang, Keming Lu, Keqin Bao, Kexin Yang, Le~Yu, Mei Li, Mingfeng Xue, Pei Zhang, Qin Zhu, Rui Men, Runji Lin, Tianhao Li, Tianyi Tang, Tingyu Xia, Xingzhang Ren, Xuancheng Ren, Yang Fan, Yang Su, Yichang Zhang, Yu~Wan, Yuqiong Liu, Zeyu Cui, Zhenru Zhang, and Zihan Qiu. 2025.
\newblock \href {https://arxiv.org/abs/2412.15115} {Qwen2.5 technical report}.
\newblock \emph{Preprint}, arXiv:2412.15115.

\bibitem[{Raiaan et~al.(2024)Raiaan, Mukta, Fatema, Fahad, Sakib, Mim, Ahmad, Ali, and Azam}]{raiaan2024reviewonlargelanguagemodels}
Mohaimenul Azam~Khan Raiaan, Md. Saddam~Hossain Mukta, Kaniz Fatema, Nur~Mohammad Fahad, Sadman Sakib, Most Marufatul~Jannat Mim, Jubaer Ahmad, Mohammed~Eunus Ali, and Sami Azam. 2024.
\newblock \href {https://doi.org/10.1109/ACCESS.2024.3365742} {A review on large language models: Architectures, applications, taxonomies, open issues and challenges}.
\newblock \emph{IEEE Access}, 12:26839--26874.

\bibitem[{R{\"o}nnqvist et~al.(2022)R{\"o}nnqvist, Kyr{\"o}l{\"a}inen, Myntti, Ginter, and Laippala}]{ronnqvist-2022-explaining-classes}
Samuel R{\"o}nnqvist, Aki-Juhani Kyr{\"o}l{\"a}inen, Amanda Myntti, Filip Ginter, and Veronika Laippala. 2022.
\newblock \href {https://doi.org/10.18653/v1/2022.findings-acl.85} {Explaining classes through stable word attributions}.
\newblock In \emph{Findings of the Association for Computational Linguistics: ACL 2022}, pages 1063--1074, Dublin, Ireland. Association for Computational Linguistics.

\bibitem[{Sahakyan et~al.(2021)Sahakyan, Aung, and Rahwan}]{sahakyan2021}
Maria Sahakyan, Zeyar Aung, and Talal Rahwan. 2021.
\newblock \href {https://doi.org/10.1109/ACCESS.2021.3116481} {Explainable artificial intelligence for tabular data: A survey}.
\newblock \emph{IEEE Access}, 9:135392--135422.

\bibitem[{Sarti et~al.(2023)Sarti, Feldhus, Sickert, and van~der Wal}]{sarti-2023-inseq}
Gabriele Sarti, Nils Feldhus, Ludwig Sickert, and Oskar van~der Wal. 2023.
\newblock \href {https://aclanthology.org/2023.acl-demo.40} {Inseq: An interpretability toolkit for sequence generation models}.
\newblock In \emph{Proceedings of the 61st Annual Meeting of the Association for Computational Linguistics (Volume 3: System Demonstrations)}, pages 421--435, Toronto, Canada. Association for Computational Linguistics.

\bibitem[{Schmidtova et~al.(2024)Schmidtova, Mahamood, Balloccu, Dusek, Gatt, Gkatzia, Howcroft, Platek, and Sivaprasad}]{schmidtova-etal-2024-automatic-metrics}
Patricia Schmidtova, Saad Mahamood, Simone Balloccu, Ondrej Dusek, Albert Gatt, Dimitra Gkatzia, David~M. Howcroft, Ondrej Platek, and Adarsa Sivaprasad. 2024.
\newblock \href {https://aclanthology.org/2024.inlg-main.44/} {Automatic metrics in natural language generation: A survey of current evaluation practices}.
\newblock In \emph{Proceedings of the 17th International Natural Language Generation Conference}, pages 557--583, Tokyo, Japan. Association for Computational Linguistics.

\bibitem[{Sellam et~al.(2020)Sellam, Das, and Parikh}]{sellam-etal-2020-bleurt}
Thibault Sellam, Dipanjan Das, and Ankur Parikh. 2020.
\newblock \href {https://doi.org/10.18653/v1/2020.acl-main.704} {{BLEURT}: Learning robust metrics for text generation}.
\newblock In \emph{Proceedings of the 58th Annual Meeting of the Association for Computational Linguistics}, pages 7881--7892, Online. Association for Computational Linguistics.

\bibitem[{Shrikumar et~al.(2017)Shrikumar, Greenside, and Kundaje}]{shrikumar-2017-deeplift}
Avanti Shrikumar, Peyton Greenside, and Anshul Kundaje. 2017.
\newblock Learning important features through propagating activation differences.
\newblock In \emph{Proceedings of the 34th International Conference on Machine Learning - Volume 70}, ICML'17, page 3145–3153. JMLR.org.

\bibitem[{Simonyan et~al.(2014)Simonyan, Vedaldi, and Zisserman}]{simonyan-2014-deep-inside}
Karen Simonyan, Andrea Vedaldi, and Andrew Zisserman. 2014.
\newblock \href {https://arxiv.org/abs/1312.6034} {Deep inside convolutional networks: Visualising image classification models and saliency maps}.
\newblock In \emph{Workshop at International Conference on Learning Representations}.

\bibitem[{Singha et~al.(2023)Singha, Cambronero, Gulwani, Le, and Parnin}]{singha2023tabularrepresentationnoisyoperators}
Ananya Singha, José Cambronero, Sumit Gulwani, Vu~Le, and Chris Parnin. 2023.
\newblock \href {https://arxiv.org/abs/2310.10358} {Tabular representation, noisy operators, and impacts on table structure understanding tasks in {LLM}s}.
\newblock \emph{Preprint}, arXiv:2310.10358.

\bibitem[{Smock et~al.(2022)Smock, Pesala, and Abraham}]{pubtables1m}
Brandon Smock, Rohith Pesala, and Robin Abraham. 2022.
\newblock \href {https://doi.org/10.1109/CVPR52688.2022.00459} {Pub{T}ables-1{M}: Towards comprehensive table extraction from unstructured documents}.
\newblock In \emph{2022 IEEE/CVF Conference on Computer Vision and Pattern Recognition (CVPR)}, pages 4624--4632.

\bibitem[{Suadaa et~al.(2021)Suadaa, Kamigaito, Funakoshi, Okumura, and Takamura}]{suadaa-etal-2021-towards}
Lya~Hulliyyatus Suadaa, Hidetaka Kamigaito, Kotaro Funakoshi, Manabu Okumura, and Hiroya Takamura. 2021.
\newblock \href {https://doi.org/10.18653/v1/2021.acl-long.115} {Towards table-to-text generation with numerical reasoning}.
\newblock In \emph{Proceedings of the 59th Annual Meeting of the Association for Computational Linguistics and the 11th International Joint Conference on Natural Language Processing (Volume 1: Long Papers)}, pages 1451--1465, Online. Association for Computational Linguistics.

\bibitem[{Sui et~al.(2024)Sui, Zhou, Zhou, Han, and Zhang}]{sui-2024-table-meets-llm}
Yuan Sui, Mengyu Zhou, Mingjie Zhou, Shi Han, and Dongmei Zhang. 2024.
\newblock \href {https://doi.org/10.1145/3616855.3635752} {Table meets {LLM}: Can large language models understand structured table data? {A} benchmark and empirical study}.
\newblock In \emph{Proceedings of the 17th ACM International Conference on Web Search and Data Mining}, WSDM '24, page 645–654, New York, NY, USA. Association for Computing Machinery.

\bibitem[{Team et~al.(2024)Team, Anil, and et. al}]{geminiteam2024geminifamilyhighlycapable}
Gemini Team, Rohan Anil, and Sebastian~Borgeaud et. al. 2024.
\newblock \href {https://arxiv.org/abs/2312.11805} {Gemini: A family of highly capable multimodal models}.
\newblock \emph{Preprint}, arXiv:2312.11805.

\bibitem[{Tenney et~al.(2024)Tenney, Mullins, Du, Pandya, Kahng, and Dixon}]{tenney-2024-sequence-salience}
Ian Tenney, Ryan Mullins, Bin Du, Shree Pandya, Minsuk Kahng, and Lucas Dixon. 2024.
\newblock \href {https://arxiv.org/abs/2404.07498} {Interactive prompt debugging with sequence salience}.
\newblock \emph{arXiv}, abs/2404.07498.

\bibitem[{Vaswani et~al.(2017)Vaswani, Shazeer, Parmar, Uszkoreit, Jones, Gomez, Kaiser, and Polosukhin}]{vaswani2017}
Ashish Vaswani, Noam Shazeer, Niki Parmar, Jakob Uszkoreit, Llion Jones, Aidan~N. Gomez, \L{}ukasz Kaiser, and Illia Polosukhin. 2017.
\newblock Attention is all you need.
\newblock In \emph{Proceedings of the 31st International Conference on Neural Information Processing Systems}, NIPS'17, page 6000–6010, Red Hook, NY, USA. Curran Associates Inc.

\bibitem[{Wang et~al.(2024)Wang, Xu, Zhao, Ouyang, Wu, Zhao, Xu, Liu, Qu, Shang, Zhang, Wei, Sui, Li, Shi, Qiao, Lin, and He}]{wang2024mineruopensourcesolutionprecise}
Bin Wang, Chao Xu, Xiaomeng Zhao, Linke Ouyang, Fan Wu, Zhiyuan Zhao, Rui Xu, Kaiwen Liu, Yuan Qu, Fukai Shang, Bo~Zhang, Liqun Wei, Zhihao Sui, Wei Li, Botian Shi, Yu~Qiao, Dahua Lin, and Conghui He. 2024.
\newblock \href {https://arxiv.org/abs/2409.18839} {Miner{U}: An open-source solution for precise document content extraction}.
\newblock \emph{Preprint}, arXiv:2409.18839.

\bibitem[{Wu et~al.(2024{\natexlab{a}})Wu, Li, Zhu, Zhang, Liang, Ma, Xiao, Zhang, Yang, Chen, Huang, Al~Moubayed, Fu, and Lin}]{wu-2024-scimmir}
Siwei Wu, Yizhi Li, Kang Zhu, Ge~Zhang, Yiming Liang, Kaijing Ma, Chenghao Xiao, Haoran Zhang, Bohao Yang, Wenhu Chen, Wenhao Huang, Noura Al~Moubayed, Jie Fu, and Chenghua Lin. 2024{\natexlab{a}}.
\newblock \href {https://doi.org/10.18653/v1/2024.findings-acl.746} {{S}ci{MMIR}: Benchmarking scientific multi-modal information retrieval}.
\newblock In \emph{Findings of the Association for Computational Linguistics: ACL 2024}, pages 12560--12574, Bangkok, Thailand. Association for Computational Linguistics.

\bibitem[{Wu et~al.(2024{\natexlab{b}})Wu, Yang, Chai, Zhang, Liu, Du, Liang, Shu, Cheng, Sun, Niu, Li, and Li}]{wu2024tablebenchcomprehensivecomplexbenchmark}
Xianjie Wu, Jian Yang, Linzheng Chai, Ge~Zhang, Jiaheng Liu, Xinrun Du, Di~Liang, Daixin Shu, Xianfu Cheng, Tianzhen Sun, Guanglin Niu, Tongliang Li, and Zhoujun Li. 2024{\natexlab{b}}.
\newblock \href {https://arxiv.org/abs/2408.09174} {Table{B}ench: A comprehensive and complex benchmark for table question answering}.
\newblock \emph{Preprint}, arXiv:2408.09174.

\bibitem[{Yang et~al.(2025)Yang, Zhang, Liu, Freitas, and Lin}]{yang-2025-does-table-source-matter}
Bohao Yang, Yingji Zhang, Dong Liu, André Freitas, and Chenghua Lin. 2025.
\newblock \href {https://arxiv.org/abs/2501.13042} {Does table source matter? {B}enchmarking and improving multimodal scientific table understanding and reasoning}.
\newblock \emph{arXiv}, abs/2501.13042.

\bibitem[{Yin et~al.(2020)Yin, Neubig, Yih, and Riedel}]{yin-etal-2020-tabert}
Pengcheng Yin, Graham Neubig, Wen-tau Yih, and Sebastian Riedel. 2020.
\newblock \href {https://doi.org/10.18653/v1/2020.acl-main.745} {{T}a{BERT}: Pretraining for joint understanding of textual and tabular data}.
\newblock In \emph{Proceedings of the 58th Annual Meeting of the Association for Computational Linguistics}, pages 8413--8426, Online. Association for Computational Linguistics.

\bibitem[{Zhang et~al.(2024{\natexlab{a}})Zhang, Yu, Dong, Li, Su, Chu, and Yu}]{zhang-etal-2024-mm}
Duzhen Zhang, Yahan Yu, Jiahua Dong, Chenxing Li, Dan Su, Chenhui Chu, and Dong Yu. 2024{\natexlab{a}}.
\newblock \href {https://doi.org/10.18653/v1/2024.findings-acl.738} {{MM}-{LLM}s: Recent advances in {M}ulti{M}odal large language models}.
\newblock In \emph{Findings of the Association for Computational Linguistics: ACL 2024}, pages 12401--12430, Bangkok, Thailand. Association for Computational Linguistics.

\bibitem[{Zhang and Balog(2020)}]{zhang2020}
Shuo Zhang and Krisztian Balog. 2020.
\newblock \href {https://doi.org/10.1145/3372117} {Web table extraction, retrieval, and augmentation: A survey}.
\newblock \emph{ACM Trans. Intell. Syst. Technol.}, 11(2).

\bibitem[{Zhang et~al.(2024{\natexlab{b}})Zhang, Yue, Li, and Sun}]{zhang-etal-2024-tablellama}
Tianshu Zhang, Xiang Yue, Yifei Li, and Huan Sun. 2024{\natexlab{b}}.
\newblock \href {https://doi.org/10.18653/v1/2024.naacl-long.335} {{T}able{L}lama: Towards open large generalist models for tables}.
\newblock In \emph{Proceedings of the 2024 Conference of the North American Chapter of the Association for Computational Linguistics: Human Language Technologies (Volume 1: Long Papers)}, pages 6024--6044, Mexico City, Mexico. Association for Computational Linguistics.

\bibitem[{Zhang* et~al.(2020)Zhang*, Kishore*, Wu*, Weinberger, and Artzi}]{zhang2020BERTScore}
Tianyi Zhang*, Varsha Kishore*, Felix Wu*, Kilian~Q. Weinberger, and Yoav Artzi. 2020.
\newblock \href {https://openreview.net/forum?id=SkeHuCVFDr} {{BERTS}core: Evaluating text generation with {BERT}}.
\newblock In \emph{International Conference on Learning Representations}.

\bibitem[{Zhang et~al.(2024{\natexlab{c}})Zhang, Zhang, Ma, Li, Zhang, Li, Yao, Xu, Zhou, Zhang-Li, Yu, Zhao, Li, and Tang}]{zhang2024tablellmenablingtabulardata}
Xiaokang Zhang, Jing Zhang, Zeyao Ma, Yang Li, Bohan Zhang, Guanlin Li, Zijun Yao, Kangli Xu, Jinchang Zhou, Daniel Zhang-Li, Jifan Yu, Shu Zhao, Juanzi Li, and Jie Tang. 2024{\natexlab{c}}.
\newblock \href {https://arxiv.org/abs/2403.19318} {Table{LLM}: Enabling tabular data manipulation by {LLM}s in real office usage scenarios}.
\newblock \emph{Preprint}, arXiv:2403.19318.

\bibitem[{Zhang et~al.(2024{\natexlab{d}})Zhang, Wang, Dou, Wang, Wu, Zhu, and Che}]{zhang-2024-flextaf}
Xuanliang Zhang, Dingzirui Wang, Longxu Dou, Baoxin Wang, Dayong Wu, Qingfu Zhu, and Wanxiang Che. 2024{\natexlab{d}}.
\newblock \href {https://arxiv.org/abs/2408.08841} {{FLEXTAF}: Enhancing table reasoning with flexible tabular formats}.
\newblock \emph{arXiv}, abs/2408.08841.

\bibitem[{Zhao et~al.(2023)Zhao, Ji, Zhang, He, Wang, Wang, Feng, and Zhang}]{zhao-etal-2023-large}
Bowen Zhao, Changkai Ji, Yuejie Zhang, Wen He, Yingwen Wang, Qing Wang, Rui Feng, and Xiaobo Zhang. 2023.
\newblock \href {https://doi.org/10.18653/v1/2023.emnlp-main.914} {Large language models are complex table parsers}.
\newblock In \emph{Proceedings of the 2023 Conference on Empirical Methods in Natural Language Processing}, pages 14786--14802, Singapore. Association for Computational Linguistics.

\bibitem[{Zhao et~al.(2019)Zhao, Peyrard, Liu, Gao, Meyer, and Eger}]{zhao-etal-2019-moverscore}
Wei Zhao, Maxime Peyrard, Fei Liu, Yang Gao, Christian~M. Meyer, and Steffen Eger. 2019.
\newblock \href {https://doi.org/10.18653/v1/D19-1053} {{M}over{S}core: Text generation evaluating with contextualized embeddings and earth mover distance}.
\newblock In \emph{Proceedings of the 2019 Conference on Empirical Methods in Natural Language Processing and the 9th International Joint Conference on Natural Language Processing (EMNLP-IJCNLP)}, pages 563--578, Hong Kong, China. Association for Computational Linguistics.

\bibitem[{Zhao et~al.(2024)Zhao, Feng, Liu, Tang, Wei, Wu, Liao, Ye, Liu, Zhou, Li, and Huang}]{zhao2024tabpedia}
Weichao Zhao, Hao Feng, Qi~Liu, Jingqun Tang, Shu Wei, Binghong Wu, Lei Liao, Yongjie Ye, Hao Liu, Wengang Zhou, Houqiang Li, and Can Huang. 2024.
\newblock \href {https://arxiv.org/abs/2406.01326} {Tab{P}edia: Towards comprehensive visual table understanding with concept synergy}.
\newblock \emph{Preprint}, arXiv:2406.01326.

\bibitem[{Zheng et~al.(2024)Zheng, Feng, Si, She, Lin, Jiang, and Wang}]{zheng-etal-2024-multimodal}
Mingyu Zheng, Xinwei Feng, Qingyi Si, Qiaoqiao She, Zheng Lin, Wenbin Jiang, and Weiping Wang. 2024.
\newblock \href {https://doi.org/10.18653/v1/2024.acl-long.493} {Multimodal table understanding}.
\newblock In \emph{Proceedings of the 62nd Annual Meeting of the Association for Computational Linguistics (Volume 1: Long Papers)}, pages 9102--9124, Bangkok, Thailand. Association for Computational Linguistics.

\bibitem[{Zheng et~al.(2020)Zheng, Burdick, Popa, Zhong, and Wang}]{zheng2020globaltableextractorgte}
Xinyi Zheng, Doug Burdick, Lucian Popa, Xu~Zhong, and Nancy Xin~Ru Wang. 2020.
\newblock \href {https://arxiv.org/abs/2005.00589} {Global table extractor ({GTE}): A framework for joint table identification and cell structure recognition using visual context}.
\newblock \emph{Preprint}, arXiv:2005.00589.

\end{thebibliography}
\end{document}